\documentclass[]{style}

\usepackage{titletoc}
\usepackage[toc,page,header]{appendix}

\usepackage{minitoc}
\usepackage{natbib}
\usepackage{CJKutf8}

\usepackage{xargs}

\usepackage{todonotes}

\usepackage{multirow}

\usepackage{cleveref}

\usepackage{amsmath}
\usepackage{dsfont}

\usepackage{svg}

\usepackage{mathrsfs}
\usepackage{adjustbox}
\usepackage{multirow}
\usepackage{multicol}
\usepackage{tcolorbox}
\usepackage{changepage}
\usepackage{enumitem}
\usepackage{graphicx}
\usepackage{amssymb}
\usepackage{xcolor}
\usepackage{float}
\usepackage{multirow}
\usepackage{threeparttable}
\usepackage{graphicx}
\usepackage{subcaption}
\usepackage{algorithm}
\usepackage{algpseudocode}
\usepackage{wrapfig}
\usepackage[table]{xcolor}  % loads xcolor with table support
\usepackage{colortbl}       % gives \columncolor, \rowcolor
\usepackage{tabularx} % Add to your preamble
\usepackage{makecell} % Add to your preamble
\usepackage[dvipsnames]{xcolor}

\newcolumntype{g}{>{\columncolor{gray!10}}c} % gray background

\definecolor{catgray}{gray}{0.9}
\definecolor{skyblue}{rgb}{0.53,0.81,0.92} % sky blue

\colorlet{skyblue!30}{skyblue!30!white} % 30% skyblue, 70% white

\definecolor{customblue}{RGB}{70,130,180}  % This is equivalent to rgb(70,130,180)

\newtcolorbox{evolbox}[2][]{%
  enhanced,
  colframe=customblue,
  colback=white,
  coltitle=white,
  rounded corners,
  boxrule=1pt,
  titlerule=0pt,
  toptitle=1mm,
  bottomtitle=1mm,
  fonttitle=\bfseries,
  width=#2\textwidth, % This takes the second parameter as the width fraction
  #1
}
\usepackage{url}

\PassOptionsToPackage{table,xcdraw}{xcolor}
\usepackage{placeins}
\usepackage{pifont}

\definecolor{RowBlue}{HTML}{E9F2FB}
\definecolor{RowRed}{HTML}{F9EAEA}
\definecolor{Top1}{HTML}{50DB4B} % 深绿
\definecolor{Top2}{HTML}{A5FFA2} % 中绿
\definecolor{Top3}{HTML}{D9FFD9} % 浅绿
\definecolor{Sub1}{HTML}{C7DBF2}
\definecolor{Sub2}{HTML}{E4E4E4}

\newcommand{\myparagraph}[1]{\textbf{#1}\hspace{1.8ex}}

\renewcommand{\emph}[1]{\textit{#1}}

\usepackage{fixmath,mathtools,nicefrac,resources/mmstyle}
\usepackage{anyfontsize}
\usepackage[T1]{fontenc}
\usepackage{ulem}

\definecolor{my_green}{RGB}{51,102,0}
\definecolor{my_red}{RGB}{204, 0, 0}
\definecolor{myblue}{RGB}{218,232,252}
\definecolor{mygray}{RGB}{220,220,220}
\definecolor{mypink}{RGB}{251,49,153}
\newcommand{\ours}{Helios\xspace}

\title{Helios: Real Real-Time Long Video Generation Model}
\author[1,2, \circ]{Shenghai Yuan}
\author[2,4, \circ]{Yuanyang Yin}
\author[1]{Zongjian Li}
\author[2]{Xinwei Huang}
\author{~ \hspace{0.9\textwidth} ~}

\makeatletter
\renewcommand\author[2][]{%
  \addtolist[#1]{#2}{\authorlist}{\authorformat}{}
  \renewcommand\author[2][]{% 
    \addtolist[##1]{##2}{\authorlist}{\authorformat}{, }%
  }%
}
\makeatother

\author[3, \S]{Xiao Yang}
\author[1, \dagger]{Li Yuan}
\affiliation[1]{Peking University}
\affiliation[2]{ByteDance China}
\affiliation[3]{Canva}
\affiliation[4]{Chengdu Anu Intelligence}
\contribution[\circ]{Work done during internship at ByteDance}
\contribution[\S]{Project Leader}
\contribution[\dagger]{Corresponding Author}

\abstract{
  \textit{\uline{We introduce \textbf{\ours}, the first 14B video generation model that runs at \textbf{19.5 FPS on a single} NVIDIA H100 GPU and supports minute-scale generation while matching the quality of a strong baseline.}} We make breakthroughs along three key dimensions: \textbf{(1)} robustness to long-video drifting without commonly used anti-drifting heuristics such as self-forcing, error-banks, or keyframe sampling; \textbf{(2)} real-time generation without standard acceleration techniques such as KV-cache, sparse/linear attention, or quantization; and \textbf{(3)} training without parallelism or sharding frameworks, enabling image-diffusion-scale batch sizes while fitting up to four 14B models within 80 GB of GPU memory. Specifically, \ours is a \textbf{14B autoregressive diffusion model} with a unified input representation that natively supports T2V, I2V, and V2V tasks. To mitigate drifting in long-video generation, we characterize typical failure modes and propose simple yet effective training strategies that explicitly simulate drifting during training, while eliminating repetitive motion at its source. For efficiency, we heavily compress the historical and noisy context and reduce the number of sampling steps, yielding computational costs comparable to---or lower than---those of 1.3B video generative models. Moreover, we introduce infrastructure-level optimizations that accelerate both inference and training while reducing memory consumption. Extensive experiments demonstrate that \ours consistently outperforms prior methods on both short- and long-video generation.
  We plan to release the code, base model, and distilled model to support further development by the community.
}

\checkdata[
\raisebox{-0.2em}{\includegraphics[width=0.025\linewidth]{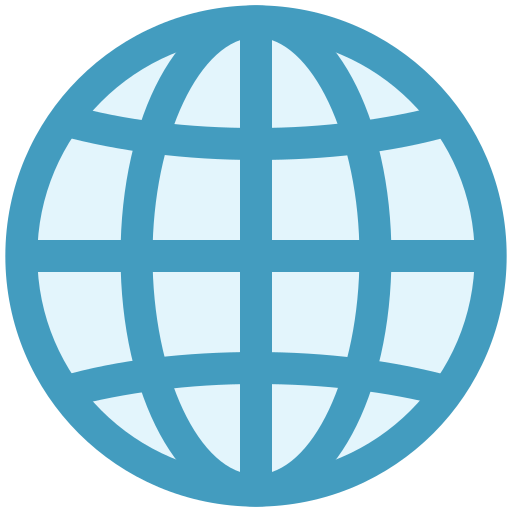}}~~Project Page]{\href{https://pku-yuangroup.github.io/Helios-Page}{\texttt{https://pku-yuangroup.github.io/Helios-Page}}
\\[-1.5ex]}

\checkdata[
\raisebox{-0.3em}{\includegraphics[width=0.026\linewidth]{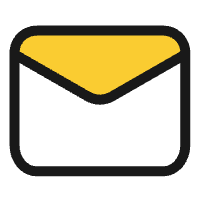}}~~Email Address]{\href{yuanshenghai@stu.pku.edu.cn}{\texttt{yuanshenghai@stu.pku.edu.cn}}}

\begin{document}
\maketitle

\section{Introduction}\label{sec:intro}

\begin{quote}
\centering
\textit{14B Real-Time Long Video Generation Model can be Cheaper, Faster but Keep Stronger than 1.3B} \\
\textsc{--- Helios Team}
\end{quote}

\begin{figure*}[!t]
  \centering
  \includegraphics[width=1\linewidth]{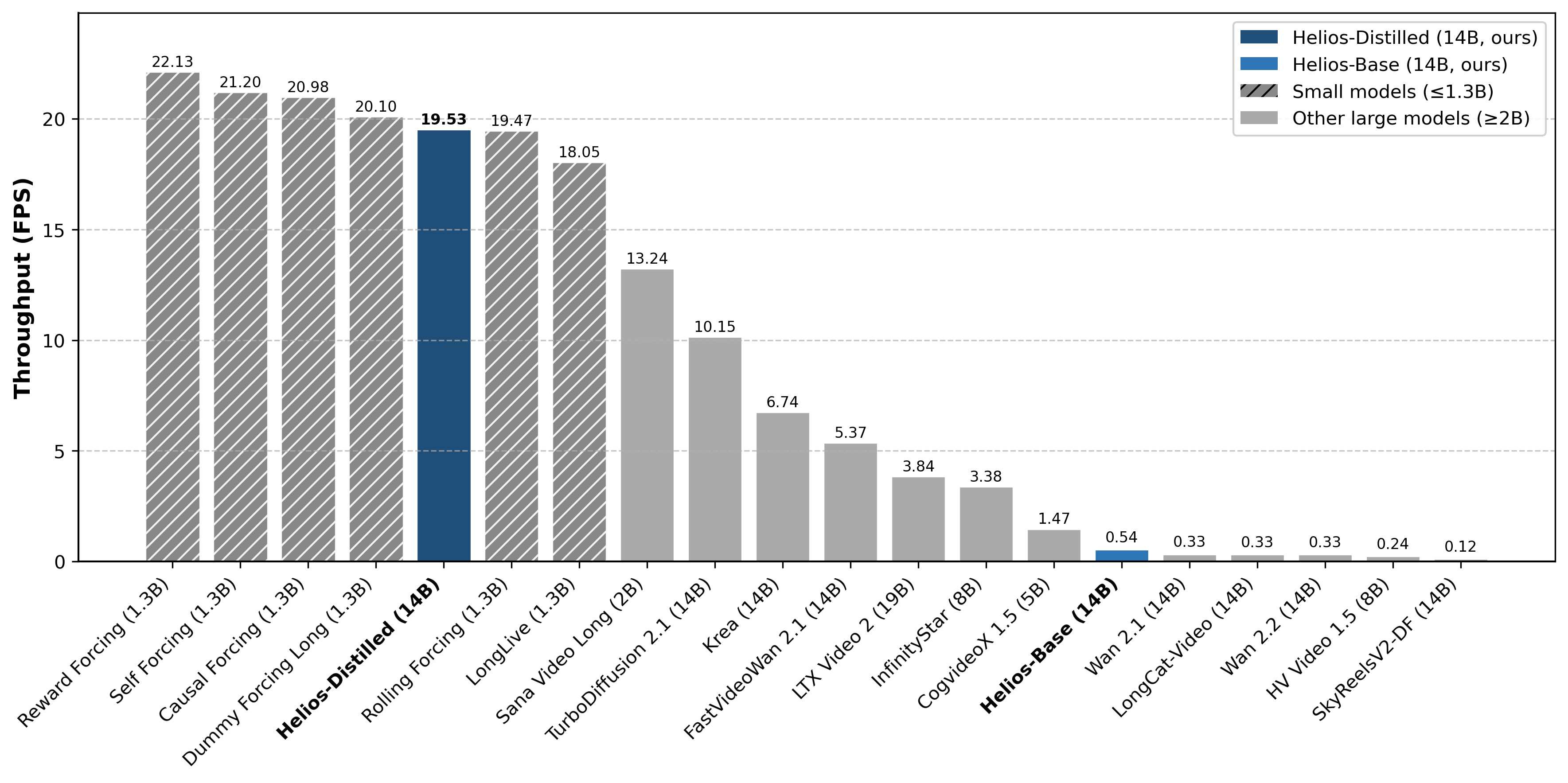}
  \caption{\textbf{End-to-end throughput (FPS) of various video generation models on a single H100.} The results are obtained at the same resolution with all official acceleration techniques, including FlashAttention, torch compile, and KV-cache. Helios is substantially faster than models at the same scale and matches the speed of smaller distilled ones.}
  \label{figure_highlight_speed}
\end{figure*}

\begin{figure*}[!t]
  \centering
  \includegraphics[width=1\linewidth]{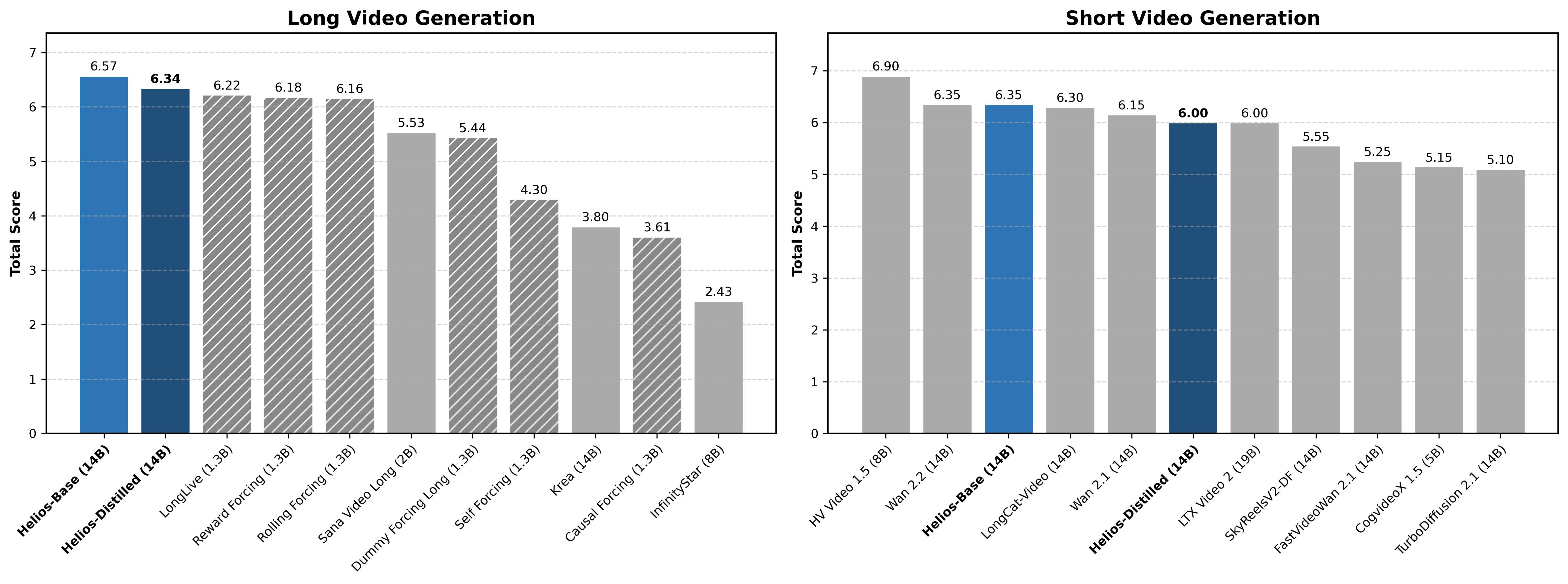}
  \caption{\textbf{Benchmark performance of Helios and its counterparts.} For both short- and long-video generation, \ours consistently outperforms existing distilled models while achieving performance comparable to that of base models.}
  \label{figure_highlight_performance}
\end{figure*}

\begin{figure*}[!t]
  \centering
  \includegraphics[width=1\linewidth]{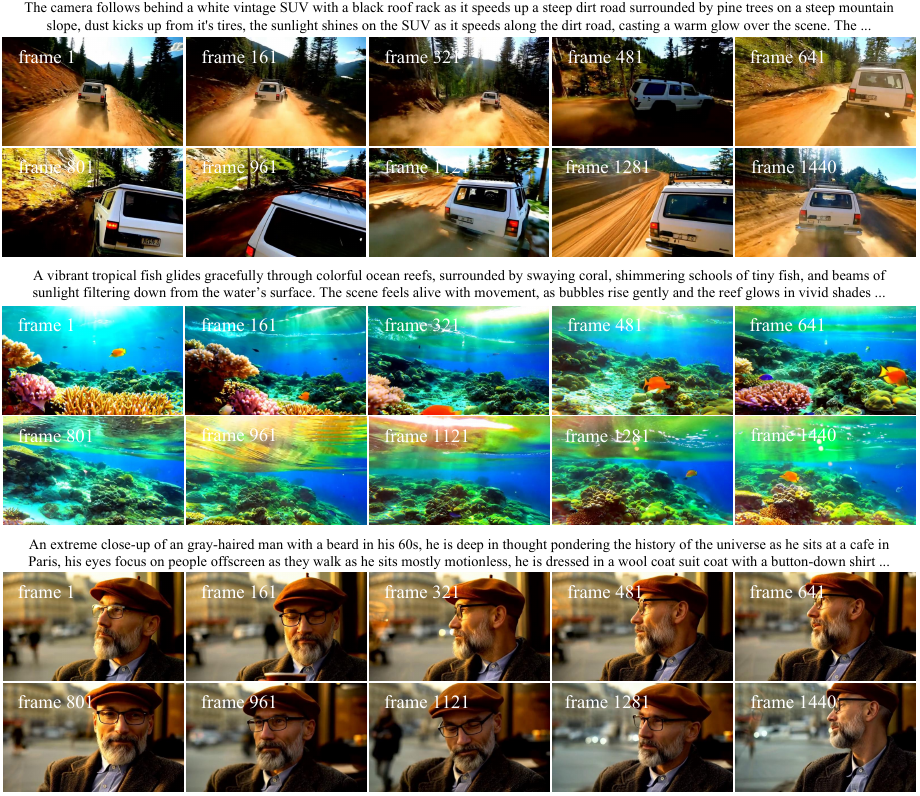}
  \caption{\textbf{Showcases of infinite videos generated by Helios.} Despite overhead comparable to that of the 1.3B models \cite{wan, longlive, rollingforcing, rewardforcing, causalforcing}, Helios still excels in visual quality, text alignment, and motion dynamics.}
  \label{figure_highlight_visual}
\end{figure*}

Over the past year, Diffusion Transformers have substantially advanced video generation \cite{opensoraplan, opensora, hunyuanvideo, hunyuanvideo15, wan, ltx, ltx2, skyreelsv2, skyreelsv3, stepvideo, kandinsky5, mochi} and shown potential as world models \cite{sora, genie, genie3, stableworld, yume, yume15, hv_world15, lingbot_world, magictime, rethinking_video_generation}. As video quality improves, demand for real-time generation has increased across applications, together with more requirements on video duration---especially for game engines \cite{mineworld, mineworldv2, hv_world, matrixgame, astra, hunyuan_Gamecraft} and interactive generation \cite{motionstream, consisid, omnihuman, magref, alphaface}. However, mainstream models remain far from real time and infinite: they typically generate only 5--10 seconds, and even these short clips can require tens of minutes to synthesize.

\textit{Real-Time Infinity Video Generation} aims to generate temporally coherent, high-quality long videos at interactive speeds, but this goal remains largely unsolved. Several community methods claim real-time infinite generation; however, these approaches typically rely on 1.3B models \cite{selfforcing++, longlive, rollingforcing, rewardforcing, dummyforcing}. The limited capacity of these models makes it difficult to represent complex motion and often leads to blurred high-frequency details. Krea-RealTime-14B \cite{krea_realtime_14b} increases the model scale, but it largely follows the same paradigm and reaches only 6.7 FPS on a single H100 GPU. In addition, these methods often rely on train-as-infer rollouts (Self-Forcing \cite{selfforcing}) to mitigate drifting, which substantially increases training cost and motivates step distillation \cite{sdxl-lightning, LCM, LCMlora, DMD, DMD2}. More critically, robustness to drifting is tightly coupled to the rollout length used during training: when training is restricted to 5-second clips, severe drifting often emerges beyond the 5-second horizon at inference. Finally, these long-video generation methods based on causal masking \cite{causvid} fundamentally change the inference regime of bidirectional pre-trained models and may limit the achievable quality.

To address these challenges, we propose \textbf{\ours}, a 14B recipe for real-time long-video generation that runs at up to 19.5 FPS on a single H100 GPU---even faster than some 1.3B models. Specifically, \textbf{(1) For Infinity Generation}, we cast long-video generation as infinite video continuation via Unified History Injection, and introduce Representation Control and Guidance Attention to efficiently inject historical context into the noisy context. This design avoids the limitations of causal masking \cite{causvid} while preserving bidirectional inference, and it unifies T2V, I2V, and V2V within a single architecture. \textbf{(2) For High-Quality Generation}, we identify three canonical manifestations of drifting: position shift, color shift, and restoration shift. Based on this analysis, we propose simple yet effective strategies that explicitly simulate drifting during training, enabling long-video generation without drifting---without self-forcing \cite{selfforcing} or error-banks \cite{SVI}. In addition, we resolve the conflict between the periodic structure of rotary position embeddings (RoPE) \cite{RoPE} and multi-head attention \cite{attention}, eliminating repetitive motion at its source \cite{LoL}. \textbf{(3) For Real-time Generation}, to remove redundancy in both historical and noisy contexts, we propose Multi-Term Memory Patchification and Pyramid Unified Predictor Corrector that substantially reduce the number of tokens fed into the DiT. We further reformulate flow matching from a ``full-resolution noise to full-resolution data'' trajectory to multiple ``low-resolution noise to multi-resolution data'' trajectories, reducing compute to a level comparable to---or even lower than---that of image diffusion models \cite{sd35, uniworld, uniworldv2, z_image, flux_2, qwen_image}. We introduce Adversarial Hierarchical Distillation, a purely teacher-forced approach that uses only the autoregressive model as the teacher, reducing the number of sampling steps from 50 to 3. Together with infrastructure-level optimizations for memory efficiency and throughput, these advances push the system toward real-time video generation. To the best of our knowledge, \ours is the first 14B video generation model to reach 19.5 FPS on a single H100 GPU, delivering a 128$\times$ speedup while maintaining comparable quality. Finally, to address the lack of a comprehensive open-source benchmark for real-time long-video generation, we construct \textbf{HeliosBench}, which comprises 240 prompts spanning four duration regimes: very short (81 frames), short (240 frames), medium (720 frames), and long (1440 frames).
Showcases, along with some benchmark results, are presented in Figures~\ref{figure_highlight_speed}, \ref{figure_highlight_performance}, and~\ref{figure_highlight_visual}.

Our contributions can be summarized as follows:

\begin{itemize}
  \item \textit{\uline{Without commonly used anti-drifting strategies}} (\eg, self-forcing, error-banks, keyframe sampling, or inverted sampling), \ours generates minute-scale videos with high quality and strong coherence.

  \item \textit{\uline{Without standard acceleration techniques}} (\eg, KV-cache, causal masking, sparse/linear attention, TinyVAE, progressive noise schedules, hidden-state caching, or quantization), \ours achieves 19.5 FPS in end-to-end inference for a 14B video generation model on a single H100 GPU.

  \item \textit{\uline{We introduce optimizations that improve both training and inference throughput while reducing memory consumption.}} These changes enable training a 14B video generation model without parallelism or sharding infrastructure, with batch sizes comparable to image models.

  \item \textit{\uline{To address the lack of standardized benchmarks for real-time long-video generation}}, we release HeliosBench. Extensive experiments demonstrate that \ours significantly outperforms existing methods in quality while achieving inference speeds that surpass some 1.3B distilled models.
\end{itemize}

\section{Related Work}\label{sec:related_work}

\subsection{Long Video Generation}
Most video generation models remain limited to short clips (typically 5--10 seconds), and scaling to longer durations without drifting remains challenging. Early methods such as FreeNoise \cite{freenoise} and FIFO-Diffusion \cite{fifo-diffusion} use training-free noise rescheduling. Subsequent approaches, including Diffusion Forcing \cite{diffusionforcing} and Rolling Diffusion \cite{rollingdiffusionmodel}, inject frame-wise independent noise over the full sequence during training to mimic inference-time context corruption, enabling long-video synthesis via autoregressive diffusion \cite{ar_diffusion}. Later work \cite{longcat_video, skyreelsv2, magi} extends this paradigm to larger models. FramePack \cite{framepack} trains a next-frame prediction model and introduces inverted sampling to reduce drifting. Self-Forcing \cite{selfforcing} adopts causal attention \cite{causvid} and proposes a train-as-infer rollout strategy to improve quality. Recent advances further explore error-bank mechanisms \cite{SVI, self_resampling, bagger}, GPT-like architectures \cite{infinitystar, nova, emu35}, keyframe sampling \cite{captain_cinema, storydiffusion, ic_lora}, test-time training \cite{ttt, ttrl}, and multi-shot generation \cite{LCT, cai2025mixture, moga}. Despite this progress, they often exhibit pronounced drifting beyond their training horizon or rely on costly long-video fine-tuning, which limits their practicality for long-video generation.

% 现有视频生成模型通常只能生成5~10秒短片段,扩展时长并避免drifring仍是难题。早期的FreeNoise和FIFO-Diffusion采用无需训练的噪声重调度方案。后续的Diffusion Forcing、Rolling Diffusion等方法通过训练时对全序列添加帧级独立噪声来模拟推理时的上下文污染,并通过AR-Diffusion实现长视频生成。而后的LongCat-Video、SkyReels-V2、MAGI-1则将该策略应用到了更大规模的模型上。FramePack则训练了一个next-frame prediction model并且提出inverted sampling缓解drifring。Self-Forcing则采用因果注意力机制，并且提出Train-as-infer rollout策略，以此来生成高质量视频。此外，还有Error-Bank (SVI)、类GPT架构（InfinityStar）、视频自回归建模（NOVA）、测试时训练（TTT）以及多镜头生成（LCT）等最新进展。然而，这些方法要么在训练范围之外表现出严重的颜色漂移，要么需要复杂的长视频微调，这对长视频生成提出了挑战。

\subsection{Real-Time Video Generation}
Long-video generation demands efficient architectures and inference pipelines. For instance, using Wan2.1 14B \cite{wan}, producing a 5-second video can take roughly 50 minutes on a single NVIDIA A100 GPU to reach acceptable quality. Common acceleration directions include parallelism, distillation \cite{sdxl-lightning, LCM, DMD2}, linear \cite{yang2023gated, team2025kimi, sana_video} or sparse attention \cite{yan2025flashsparseattentionalternative, zhang2025spargeattn, li2025radial}, hidden-state caching \cite{teacache, magcache, omnicache}, and quantization \cite{sageattention, sparse_videogen, turbodiffusion}. Existing real-time long-video systems are mostly distillation-based; \eg, \cite{selfforcing++, longlive, rollingforcing, rewardforcing, dummyforcing} follow CausVid \cite{causvid} and use DMD \cite{DMD} to reduce sampling steps from 50 to 4, together with Self-Forcing-style rollouts \cite{selfforcing} to narrow the train--inference gap. However, these methods are typically built on relatively small backbones (\eg, Wan2.1 1.3B \cite{wan}), which limits their ability to model complex motion and preserve high-frequency details. Moreover, although Krea \cite{krea_realtime_14b} reports 11 FPS on a single NVIDIA B200 GPU, its speed drops to 6.7 FPS on an H100 GPU, and the results suffer from severe drifting, which remains problematic for real-time interactive generation. Additionally, some works claim to be real-time but actually require 8 GPUs \cite{lingbot_world, hv_world15, streamdiffusionv2}.

% 生成更长的视频需要高效的架构支持。以Wan2.1 14B为例,即使只生成5秒的视频,在单个NVIDIA A100上也需要花费大约50分钟才能合成令人满意的结果。加速策略可以笼统分为以下几类，parallel、distillation、linear attention、sparse attention、hidden state caching、quantization。现有的实时长视频生成方案基本都限制在distillation，例如MotionStream、LongLive、Rolling-Forcing等都遵循CausVid使用DMD将推理步数从50步压缩至4步，并且采用Self-Forcing的rollout procedures减少train-infer的gap。然而，这些方案大多基于较小的模型，例如Wan 2.1 1.3B。这些模型难以模拟复杂的运动，并且高频细节表现欠佳，在复杂任务中显得力不从心。此外，Krea虽然声称能在单个NVIDIA B200上实现11 FPS的实时生成，但在NVIDIA H100上只有6.7FPS，并且其生成的视频具有严重的drifring，这对实时交互提出了挑战。

\section{Helios}\label{sec:method}
\textbf{(1)} \textbf{For Infinity Generation}, we introduce \textit{Unified History Injection} to convert a bidirectional pre-trained model \cite{wan} into an autoregressive generator, enabling text-to-video (T2V), image-to-video (I2V), and video-to-video (V2V) within a unified framework.
\textbf{(2)} \textbf{For High-Quality Generation}, we propose \textit{Easy Anti-Drifting} to mitigate drifting, enabling high-quality minute-scale video generation without inefficient self-forcing \cite{selfforcing} or error-banks \cite{SVI}.
\textbf{(3)} \textbf{For Real-Time Generation}, we further propose \textit{Deep Compression Flow} to reduce both the number of visual tokens and sampling steps, enabling real-time generation on a single GPU with a 14B model.

% The overview is illustrated in Figure XXX. 具体而言，(1) 首先For Infinity Generation，我们使用Unified History Injection将bidirectional预训练model转成autoregressive model，能够在单一框架内无缝支持文本转视频 (T2V)、图像转视频 (I2V) 和视频转视频 (V2V) 任务。(2) 然后For High-Quality Generation, 我们提出Easy Anti-Drifting策略避免drifting，无需低效的train-as-infer procedures即可实现高质量的分钟规模的视频生成。(3) 最后For Real-Time Generation, 我们进一步提出Deep Compression Flow降低所需输入给模型的token数以及推理步数，最终在14B模型上实现实时视频生成。

\begin{figure*}[!t]
  \centering
  \includegraphics[width=0.93\linewidth]{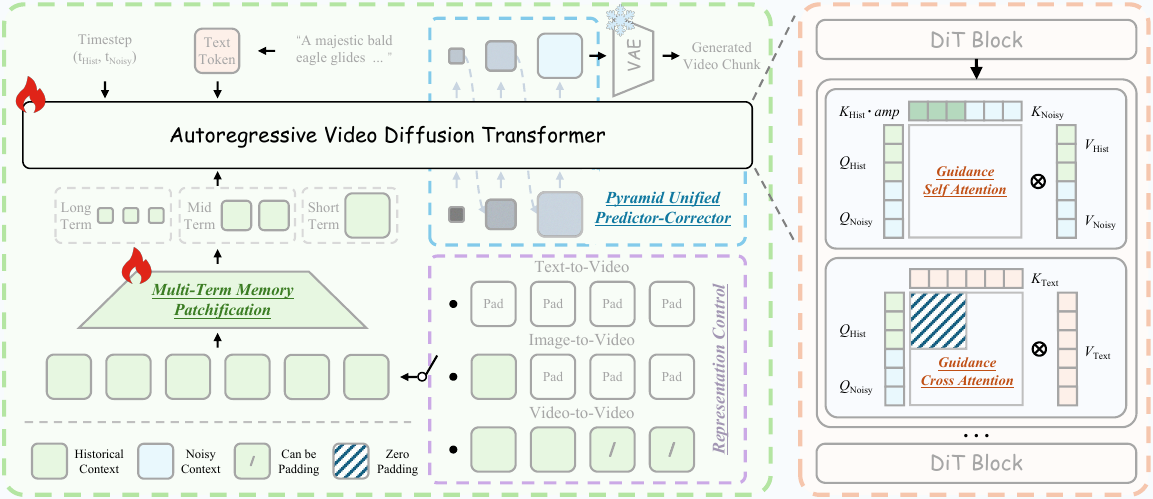}
  \caption{\textbf{Architecture of Helios.} Helios is an autoregressive video diffusion transformer built with Guidance Attention blocks. It reduces overhead by compressing historical and noisy context through Multi-Term Memory Patchification and Pyramid Unified Predictor Corrector, while unifying T2V, I2V, and V2V tasks via Representation Control.}
  \label{figure_pipeline}
\end{figure*}

\subsection{Unified History Injection}
% Infinity Generation
In this section, we describe how to extend a bidirectional model—originally limited to fixed-length generation—to synthesize videos of unbounded duration. The overall architecture is illustrated in Figure~\ref{figure_pipeline}.

% In this section，我们介绍如何让只能生成固定时长视频的双向模型生成无限时长视频。

\subsubsection{Representation Control}

Prior work typically turns a bidirectional model into an autoregressive generator by combining diffusion forcing \cite{diffusionforcing, diffusionforcingv2} with causal masking \cite{causvid}. However, the resulting frame-wise noise space is extremely large, which slows down optimization and often necessitates step distillation \cite{rcm, DMD, DMD2}. This approach is undesirable for two reasons: (i) the inference procedure deviates substantially from the pre-trained model, limiting the achievable performance; (ii) distilled models hinder further development within the community.

We address these issues with Representation Control, which formulates long-video generation as video continuation. As shown in Figure~\ref{figure_pipeline}, the input is the concatenation of a historical context $X_{\text{Hist}} \in \mathbb{R}^{B \times C \times T_{\text{Hist}} \times H \times W}$ and a noisy context $X_{\text{Noisy}} \in \mathbb{R}^{B \times C \times T_{\text{Noisy}} \times H \times W}$, where $B$, $C$, $T$, $H$, and $W$ denote the batch size, number of channels, number of frames, height, and width, respectively. We keep $T_{\text{Hist}}$ and $T_{\text{Noisy}}$ fixed during both training and inference, with $T_{\text{Hist}} \gg T_{\text{Noisy}}$. The model denoises $X_{\text{Noisy}}$ conditioned on $X_{\text{Hist}}$ to generate a temporally coherent continuation, enabling the generation of arbitrarily long videos. Representation Control enables automatic task switching via the representation of $X_{\text{Hist}}$: if $X_{\text{Hist}}$ is all zeros, the model performs T2V; if only the last frame is nonzero, it performs I2V; otherwise, it performs V2V.

% 现有方法通过结合diffusion forcing和causal mask将双向模型转为自回归模型，从而生成长视频。然而,这种方法面临一个关键挑战:其使用的frame-wise noise的搜索空间过于庞大,导致收敛较慢，因此现有方法通常需要结合step distillation来使用。但这种做法存在两个明显的局限性:一方面推理范式和预训练模型差异太大限制了性能上限,另一方面蒸馏后的模型也不利于社区进行二次开发。为了突破这些限制,我们提出了Representation Control,将长视频生成定义为视频续写任务。如Figure XXX所示，模型的输入由两部分拼接而成:historical context $X_{\text{Hist}} \in \mathbb{R}^{B \times C \times T_{\text{Hist}} \times H \times W}$和noisy context $X_{\text{Noisy}} \in \mathbb{R}^{B \times C \times T_{\text{Noisy}} \times H \times W}$,其中 $B, C, T, H, W$分别表示batch size、通道数、帧数、高度和宽度。在训练和推理时，$T_{\text{Noisy}}$和$T_{\text{Hist}}$的长度是固定的,且满足 $T_{\text{Hist}} > T_{\text{Noisy}}$。模型的目标是通过对 $X_{\text{Noisy}}$进行去噪处理,生成与$X_{\text{Hist}}$在内容上保持连贯的视频序列,从而实现视频的无限延续生成。此外，该策略还能通过$X_{\text{Hist}}$的representation自动切换任务模式:$X_{\text{Hist}}$全零时为text-to-video,最后一帧非零时切换为image-to-video,其他情况则执行video-to-video任务。

\subsubsection{Guidance Attention}

The historical and noisy contexts exhibit different statistics and should therefore be treated differently. The historical context contains clean content that is already aligned with the text prompt; it should not be denoised and should remain insensitive to $X_{\text{Noisy}}$. Instead, its role is to guide the denoising of $X_{\text{Noisy}}$. We explicitly enforce this separation in two ways. First, we fix the timestep of $X_{\text{Hist}}$ to 0 throughout the denoising process, indicating that it remains clean and noise-free. Second, inspired by \cite{concatid, standin}, we introduce Guidance Attention to strengthen the influence of the historical context on the generation of future frames:

In the self-attention layer, we compute the query, key, and value tensors for the noisy and historical contexts, denoted by $Q_{\text{Noisy}}, K_{\text{Noisy}}, V_{\text{Noisy}}$ and $Q_{\text{Hist}}, K_{\text{Hist}}, V_{\text{Hist}}$, respectively. To retain informative history while suppressing redundant or harmful signals, we introduce head-wise amplification tokens $amp$ to modulate the historical keys. This design selectively amplifies or attenuates historical information per attention head, encouraging the model to focus on the most discriminative components:
\begin{equation}
  X_{\text{Self}} = \text{Attention}([Q_{\text{Noisy}}, Q_{\text{Hist}}], [K_{\text{Noisy}}, K_{\text{Hist}} \cdot amp], [V_{\text{Noisy}}, V_{\text{Hist}}])
\end{equation}
where $[,]$ denotes concatenation, $\cdot$ means multiplication. In cross-attention, we inject semantic information from the text prompt into the model. Since $X_{\text{Hist}}$ has already incorporated the semantics from previous steps, re-injecting the same semantics is redundant. We therefore apply cross-attention only to $X_{\text{Noisy}}$:
\begin{equation}
  X_{\text{Cross}} = \text{Attention}(Q_{\text{Noisy}}, K_{\text{Text}}, V_{\text{Text}})
\end{equation}
where $K_{\text{Text}}$ and $V_{\text{Text}}$ are the key and value tensors of the encoded text prompt.

\begin{figure*}[!t]
  \centering
  \includegraphics[width=0.93\linewidth]{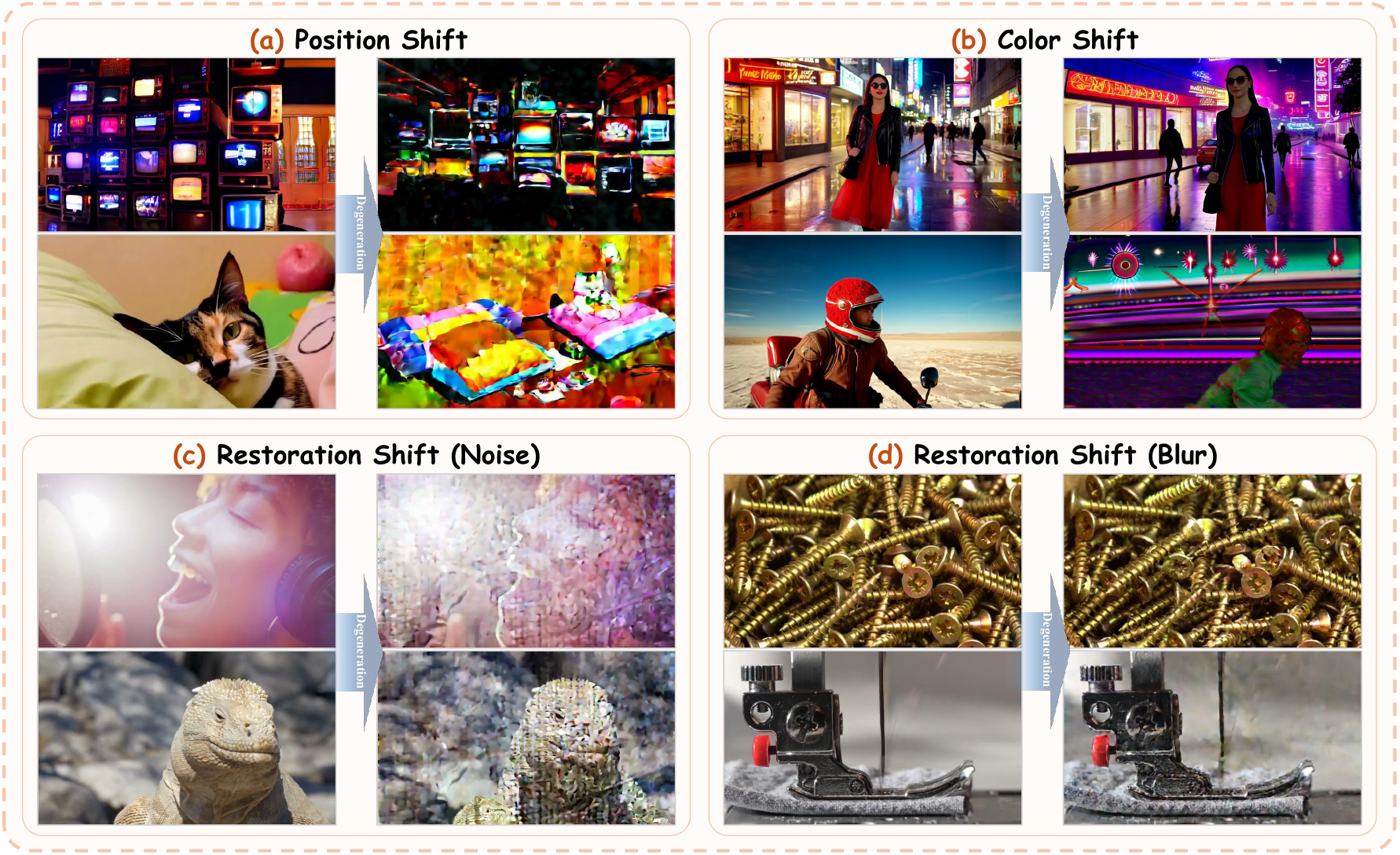}
  \caption{\textbf{Visualization of three representative drifting patterns in long-video generation.}}
  \label{figure_degeneration_visual}
\end{figure*}

\subsection{Easy Anti-Drifting}\label{sec:easy_anti_drifting}
% High-Quality Generation
In this section, we summarize three common manifestations of drifting, as shown in Figure~\ref{figure_degeneration_visual}, and present simple yet effective techniques to mitigate both drifting and repetitive motion in long-video generation, without relying on self-forcing \cite{selfforcing}, error-banks \cite{SVI}, or other commonly used anti-drifting strategies.

% In this section，我们通过实验观察揭示了曝光偏差的三种典型表现形式（如图 XXX 所示），并提出了一种在不使用 train-as-infer procedures和error-bank的情况下，有效缓解长视频退化问题的简洁方法。

\subsubsection{Relative RoPE}
A major source of drifting is positional encoding, which we term \textit{\uline{Position Shift}}. In practice, diffusion models often perform best when the inference horizon matches the training horizon; changing the video length exposes the model to unseen temporal positions and can substantially degrade quality. Existing long-video methods typically use absolute RoPE along the time dimension. For instance, generating a 1440-frame video uses indices $0\!:\!1399$, whereas training is often limited to short clips (\eg, 5 seconds), making drifting beyond the training horizon likely even with sophisticated mitigation. Training on longer videos is a direct but costly remedy \cite{longlive, rollingforcing, selfforcing++}. Moreover, absolute temporal indices may cause the generation to repeatedly snap back to early positions, leading to abrupt scene resets and cyclic patterns, which we refer to as \textit{repetitive motion} \cite{LoL}. To address these issues, we propose Relative RoPE. Regardless of the target video length, we constrain the temporal index range of $X_{\text{Hist}}$ to $0\!:\!T_{\text{Hist}}$ and assign $X_{\text{Noisy}}$ to $T_{\text{Hist}}\!:\!T_{\text{Hist}}+T_{\text{Noisy}}$. This relative indexing enables stable generation at arbitrary lengths while alleviating the interaction between RoPE periodicity and multi-head attention, thereby reducing repetitive motion at its source.

% Drifting的一个重要来源是位置编码，我们称之为position shift。通常来说,模型在推理时只有使用与训练时相同的时长才能获得最优结果。如果选择不同的时长,生成质量往往会大幅下降,这是因为模型在训练阶段未曾见过对应的位置编码。以长视频生成为例,现有方法通常对时间维度采用绝对RoPE编码。比如生成1440帧视频时,其位置索引范围为0~1399。然而这些方法在训练时通常只使用5秒的片段,这导致即使采用了复杂的抗退化策略,在5秒之后仍大概率出现明显的drifting。一种直接的解决方案是用更长的视频训练模型,但是这会急剧增加训练开销。并且该类方法会导致生成的内容反复回到第一帧，导致场景突然重置和循环运动，which we call repetitive motion。为此,我们提出了Relative RoPE: 无论生成视频有多长,$X_{\text{Hist}}$的位置索引范围始终保持在0~$T_{\text{Hist}}$,而$X_{\text{Noisy}}$则对应$T_{\text{Hist}}$~$T_{\text{Hist}}+T_{\text{Noisy}}$。通过Relative RoPE,模型即使在短clips上训练，也能实现任意长度的稳定生成，并且能够从源头杜绝旋转位置嵌入（RoPE）的周期性结构与多头注意力机制之间的固有冲突从而避免repetitive motion。

\subsubsection{First-Frame Anchor}
\begin{figure*}[!t]
  \centering
  \includegraphics[width=1\linewidth]{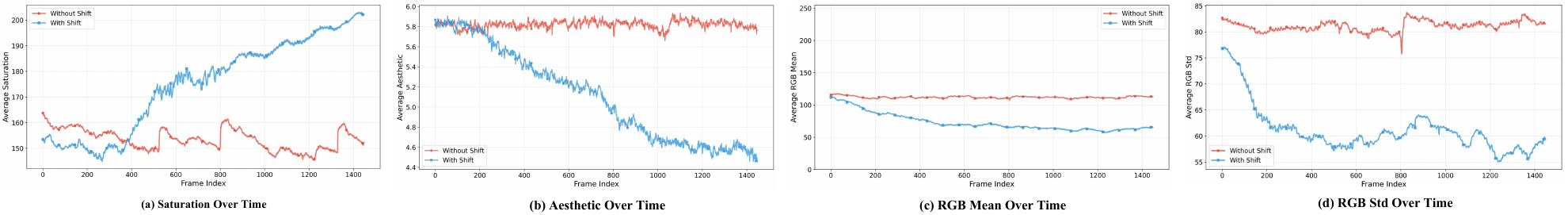}
  \caption{\textbf{Temporal trends of saturation, aesthetic, and RGB statistics for normal videos versus drifting videos.} Normal videos are stable, while drifting videos initially follow a similar trajectory but shift abruptly and remain unstable.}
  \label{figure_degeneration_score}
\end{figure*}

Drifting often appears as \textit{\uline{Color Shift}}, which becomes more severe as the generated video grows longer. To characterize this phenomenon, we analyze normal and drifting videos by tracking saturation, aesthetic scores \cite{aesthecit_predictor}, and RGB statistics (mean and variance) over time. As shown in Figure~\ref{figure_degeneration_score}, normal videos exhibit relatively stable statistics, whereas drifting videos initially follow a similar trajectory but undergo a sharp shift after a certain point and remain unstable thereafter. Notably, drifting rarely occurs at the beginning of generation. Motivated by this observation, we always retain the first frame in $X_{\text{Hist}}$ during both training and inference. Serving as a global visual anchor, this frame constrains distribution shifts in later segments, stabilizes statistics over time, and effectively mitigates color shift under autoregressive extrapolation.

% drifting的一个典型表现是：随着视频时长的增加，画面会出现color shift。为更直观地分析这一现象，我们收集了多组正常视频以及存在 drifting 的视频，并对其饱和度、美学评分、RGB 均值和 RGB 方差随时间的变化趋势进行了可视化分析。如图所示，正常视频的各项质量指标始终维持在相对稳定的区间内；相比之下，存在 drifting 的视频在生成初期表现与正常视频接近，但在某一时间点之后，其指标会突然上升或下降，并在后续时间段内持续处于不稳定状态。进一步观察可以发现，drifting 通常不会在视频生成的首段出现。基于这一观察，我们在训练和推理阶段始终保留first-frame在$X_{\text{Hist}}$中，期望其作为稳定的全局参考，对后续生成过程中的分布偏移起到约束和校正作用。在自回归生成框架下，该设计能够有效抑制长视频外推过程中的color shift问题。

\subsubsection{Frame-Aware Corrupt}
Drifting is not limited to color shifts; it can also appear as image-restoration artifacts, such as blur and noise \cite{SVI}. We refer to this phenomenon as \textit{\uline{Restoration Shift}}. This shift arises because the model is trained on clean videos but, at inference time, conditions on its own imperfect outputs as history; consequently, small errors can accumulate and amplify over time. To improve robustness to imperfect history, we propose \textit{Frame-Aware Corrupt}, inspired by \cite{diffusionforcing, diffusionforcingv2}, which simulates realistic history drift during training. Concretely, for each historical frame, we independently sample one of the following perturbations: (i) with probability $p_c$, adjust the frame exposure by a magnitude uniformly sampled from $[a_{\min}, a_{\max}]$; (ii) with probability $p_a$, add noise with a level uniformly sampled from $[b_{\min}, b_{\max}]$; (iii) with probability $p_b$, downsample and then upsample using a downsampling factor uniformly sampled from $[c_{\min}, c_{\max}]$; or (iv) with probability $p_d$, keep the latent clean, where $p_a + p_b + p_c + p_d = 1$. Perturbations are sampled independently per frame, so a history of $T_{\text{Hist}}$ frames yields $T_{\text{Hist}}$ independent corruption decisions, which is crucial for long-video stability.

% Here, we set $p_a=0.8$, $p_b=0.1$, $p_c=0.1$, $a_{\min}=0$, $a_{\max}=0.33$, $b_{\min}=0$, and $b_{\max}=0.1$.

% Drifting 的表现不仅限于color shift。在更普遍的情况下，它还会以image restoration领域常见的退化形式出现，例如模糊或噪声增强等，我们称之为restoration shift。其根本原因在于：训练阶段模型仅接触到真实且干净的视频数据，因此在推理阶段一旦生成结果出现轻微退化，模型便可能将这些错误模式作为新的历史上下文持续累积，从而使后续帧的退化不断放大。

% 为缓解上述问题，我们提出\textbf{Frame-Aware Corrupt}策略，在训练阶段显式模拟推理时可能出现的历史帧质量退化，从而提升模型对错误上下文的鲁棒性。具体而言，对于historical context，我们以帧为单位随机施加如下扰动：
% (1) 以概率 $p_a$ 向该帧添加扰动噪声，其强度从区间 $[a_{\min}, a_{\max}]$ 中均匀采样；
% (2) 以概率 $p_b$ 对该帧进行下采样，并再上采样回原始分辨率，其下采样倍率从区间 $[b_{\min}, b_{\max}]$ 中均匀采样；
% (3) 以概率 $p_c$ 保持该帧 latent 干净，不施加任何扰动，
% 其中 $p_a + p_b + p_c = 1$。

% 需要强调的是，上述操作是逐帧独立执行的：若$X_{\text{Hist}}$包含$T_{\text{Hist}}$帧，则每一帧均对应一次独立的扰动采样过程（即共有$T_{\text{Hist}}$次概率选择）。这一设计通过在训练阶段模拟推理时可能出现的质量退化，对于缓解长视频生成中的质量退化至关重要。

\subsection{Deep Compression Flow - From Token View}
% Real-Time Generation
In this section, we present a token-centric view of Deep Compression Flow. Our goal is to reduce the token-level computation of a 14B video generation model to a level comparable to that of a 1.3B model.

% In this section，我们介绍如何from token view，将14B的视频生成模型的计算开销降至与1.3B模型相当的水平，并在single-gpu上实现实时生成。

\subsubsection{Multi-Term Memory Patchification}
To enable real-time generation, we reduce redundancy in the historical context $X_{\text{Hist}}$ via Multi-Term Memory Patchification. Inspired by prior work \cite{far, hitvideo, framepack}, we leverage a simple observation: in autoregressive video generation, predicting future frames depends mostly on temporally nearby history for local motion and short-range continuity, whereas distant history primarily contributes coarse global context.

\begin{figure*}[!t]
  \centering
  \includegraphics[width=0.9\linewidth]{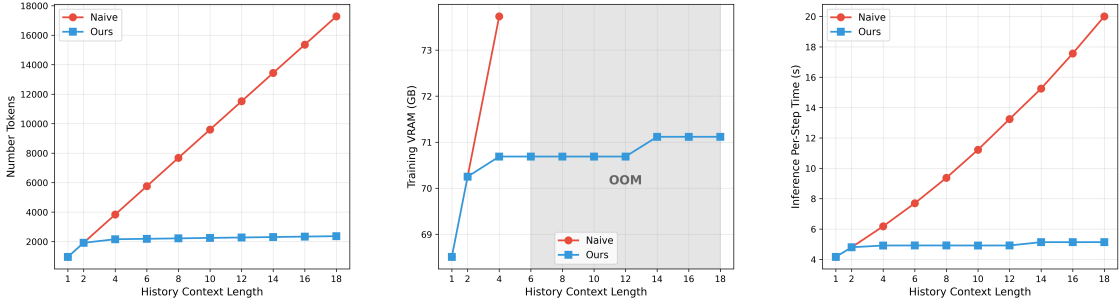}
  \caption{\textbf{Overhead reduction with Multi-Term Memory Patchification.} A hierarchical history window uses progressively larger kernels, keeping the token budget constant while extending the context length.}
  \label{figure_patchification}
\end{figure*}

Based on this observation, we adopt a hierarchical context window that partitions $X_{\text{Hist}}$ into three parts---short-, mid-, and long-term---containing $T_1$, $T_2$, and $T_3$ frames, respectively, where $0 < T_1 < T_2 < T_3$. For each part, we apply an independent Conv kernel $(p_t^{(i)}, p_h^{(i)}, p_w^{(i)})$ to compress spatiotemporal tokens, where $i \in \{1,2,3\}$ indexes the three parts. We increase the compression ratio with temporal distance; \eg, $p_t^{(1)} < p_t^{(2)} < p_t^{(3)}$, $p_h^{(1)} < p_h^{(2)} < p_h^{(3)}$, and $p_w^{(1)} < p_w^{(2)} < p_w^{(3)}$. After patchification, the number of tokens becomes:
\begin{equation}
  L_{\text{short}} = \frac{T_1 H W}{p_t^{(1)} p_h^{(1)} p_w^{(1)}}, \quad
  L_{\text{mid}} = \frac{T_2 H W}{p_t^{(2)} p_h^{(2)} p_w^{(2)}}, \quad
  L_{\text{long}} = \frac{T_3 H W}{p_t^{(3)} p_h^{(3)} p_w^{(3)}}.
\end{equation}
The total number of tokens in $X_{\text{Hist}}$ is:
\begin{equation}
  L_{\text{total}} = H W \left(\frac{T_1}{p_t^{(1)} p_h^{(1)} p_w^{(1)}} + \frac{T_2}{p_t^{(2)} p_h^{(2)} p_w^{(2)}} + \frac{T_3}{p_t^{(3)} p_h^{(3)} p_w^{(3)}}\right).
\end{equation}
As illustrated in Figure~\ref{figure_patchification}, this design keeps $L_{\text{total}}$ constant regardless of the target video length. Consequently, the model can retain substantially longer history under a fixed token budget, reducing both computational cost and memory footprint during training and inference. During training, we randomly zero out a certain proportion of the historical context to simulate T2V, I2V, and V2V during inference.

% 为了实现实时生成，我们首先针对historical context $X_{\text{Hist}}$的冗余性问题，引入\textbf{Multi-Term Memory Patchification}。受FAR、FramePack等工作的启发，该策略的核心设计基于一个关键观察：在视频自回归生成过程中,当前帧主要依赖相邻帧来捕捉局部运动和时间一致性,而较远的历史帧则更多地充当全局上下文记忆的角色。

% 基于此，我们采用了\textbf{分层上下文窗口策略},将历史上下文划分为短期、中期和长期三个层次，分别包含$T_1, T_2, T_3$帧，其中$0 < T_1 < T_2 < T_3$。

% 对于每个层次，我们采用不同尺寸的3D patchify核$(p_t^{(i)}, p_h^{(i)}, p_w^{(i)})$进行时空压缩，其中 $i \in \{1,2,3\}$对应三个层次，且压缩率递增：$p_t^{(1)} < p_t^{(2)} < p_t^{(3)}$，$p_h^{(1)} < p_h^{(2)} < p_h^{(3)}$，$p_w^{(1)} < p_w^{(2)} < p_w^{(3)}$。经过patchification后，每个层次的token数量为：

% \begin{equation}
% L_{\text{short}} = \frac{T_1 \cdot H \cdot W}{p_t^{(1)} \cdot p_h^{(1)} \cdot p_w^{(1)}}, \quad L_{\text{mid}} = \frac{T_2 \cdot H \cdot W}{p_t^{(2)} \cdot p_h^{(2)} \cdot p_w^{(2)}}, \quad L_{\text{long}} = \frac{T_3 \cdot H \cdot W}{p_t^{(3)} \cdot p_h^{(3)} \cdot p_w^{(3)}}
% \end{equation}

% historical context的token数量为：

% \begin{equation}
% L_{\text{total}} = H \cdot W \left(\frac{T_1}{p_t^{(1)} p_h^{(1)} p_w^{(1)}} + \frac{T_2}{p_t^{(2)} p_h^{(2)} p_w^{(2)}} + \frac{T_3}{p_t^{(3)} p_h^{(3)} p_w^{(3)}}\right)
% \end{equation}

% 如图XXX所示，该策略的优势在于：无论生成多长的视频，historical context的总长度 $L_{\text{total}}$ 始终保持固定，这使得模型能够以相同的计算开销存储数倍的历史上下文信息，从而显著降低了训练和推理过程中的计算成本与显存开销。

\subsubsection{Pyramid Unified Predictor Corrector}
\begin{figure*}[!t]
  \centering
  \includegraphics[width=1\linewidth]{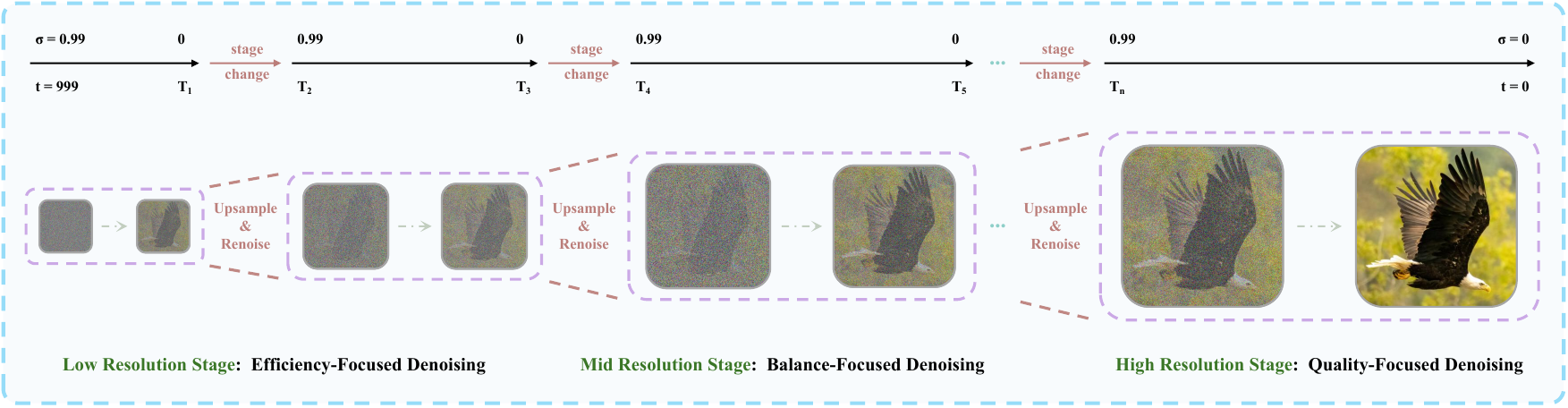}
  \caption{\textbf{Outline of Pyramid Unified Predictor Corrector.} The process consists of three stages: (i) Low-stage focuses on efficiency, (ii) Mid-stage balances quality and efficiency, and (iii) High-stage prioritizes quality.}
  \label{figure_pyramid}
\end{figure*}

To reduce redundancy in the noisy context $X_{\text{Noisy}}$, we propose Pyramid Unified Predictor Corrector, a multi-scale variant of the Unified Predictor Corrector (UniPC) sampler \cite{unipc}, as shown in Figure~\ref{figure_pyramid}. Inspired by prior works \cite{fdm, bottleneck_sampling, var, infinitystar, pyramidflow}, we observe that early sampling steps are dominated by strong noise and thus mainly determine global structure (\eg, layout and color), whereas later steps primarily refine fine-grained details (\eg, edges and textures). Accordingly, we adopt a coarse-to-fine schedule: we sample in a low-resolution latent space at early stages and progressively transition to full resolution. Concretely, \textit{\ours} learn multi-scale velocity fields that define an ODE-based generative process. Starting from low-resolution Gaussian noise $\epsilon \in \mathbb{R}^{B \times C \times T \times h \times w}$, we integrate the ODE to obtain a coarse-to-fine trajectory and progressively upsample it to get the full-resolution clean sample $x_0 \in \mathbb{R}^{B \times C \times T \times H \times W}$, where $h \ll H$ and $w \ll W$.

\noindent\myparagraph{Training.} We partition the generative process into $K$ stages with increasing spatial resolutions, where stage $k$ operates at resolution $(h^k, w^k)$. To learn a direct transport direction from scale $k\!-
\!1$ to scale $k$, we construct a linear interpolation path that serves as a continuous transition between the two scales:
\begin{equation}
  x_t^k = (1-\lambda_t)\, x^k + \lambda_t\, \mathrm{Up}(x^{k-1}),
  \label{eq: get_xt}
\end{equation}
where $k \in \{1,2,\ldots,K\}$ and $\lambda_t \in [0,1]$ controls the noise level. We use the same $\lambda_t$ schedule across stages to keep flow matching consistent across scales. The timestep $T \in [0,1000]$ associated with $\lambda_t$ is partitioned into stage boundaries $T_0 = 1000 > T_1 > \cdots > T_K = 0$, so that stage $k$ operates only on $[T_k,\, T_{k-1}]$. For the boundary conditions, when $k=1$ we start from noise, i.e., $\mathrm{Up}(x^{k-1})=\epsilon$ with $\epsilon \sim \mathcal{N}(0,I)$; when $k=K$, we recover the full-resolution sample, i.e., $x^k=x_0$. Along the linear path, the ground-truth velocity is constant:
\begin{equation}
  v^k = x^k - \mathrm{Up}(x^{k-1}).
\end{equation}
We parameterize the velocity field as $u_\theta^k(\cdot)$ and minimize the velocity-matching objective:
\begin{equation}
  \mathcal{L}
  = \mathbb{E}_{k,\lambda_t,x_t^k,\mathrm{Up}(x^{k-1}),y}
  \left[
    \left\|
    u_\theta^k(x_t^k, y, \lambda_t, k) - v^k
    \right\|_2^2
  \right],
\end{equation}
where $y$ denotes the conditioning input. In practice, we set $K=3$ to balance quality and efficiency.

\noindent\myparagraph{Inference.} We similarly partition sampling into $K$ stages and allocate $(N_1, N_2, \ldots, N_K)$ steps to each stage, resulting in total steps $N = \sum_{k=1}^{K} N_k$. At stage $k$, we sample at discrete timesteps $\{t_k^n\}_{n=0}^{N_k}$ and update:
\begin{equation}
  x^k_{t_k^n}
  = x^k_{t_k^{n-1}}
  + u_\theta^k\!\left(x^k_{t_k^{n-1}},\, y,\, t_k^{n-1}\right)
  \left(t_k^{n} - t_k^{n-1}\right).
\end{equation}
When transitioning from stage $k\!-\!1$ to $k$, naively upsampling the terminal state may introduce artifacts and break path continuity. Following PyramidFlow \cite{pyramidflow}, we upsample the terminal state using nearest-neighbor interpolation and then correct the injected noise and its covariance to maintain distributional consistency across scales. From a computational perspective, single-scale inference with $N$ steps costs $\mathcal{O}(H W N)$. In contrast, multi-scale sampling distributes steps across stages and processes fewer tokens in early stages. Under a standard pyramid (\eg, halving the resolution at each stage), the total number of processed tokens is:
\begin{equation}
  \left(
    H \times W
    + \frac{H}{2} \times \frac{W}{2}
    + \frac{H}{4} \times \frac{W}{4}
    + \cdots
    + \frac{H}{2^{K-1}} \times \frac{W}{2^{K-1}}
  \right)\times \frac{N}{K}.
\end{equation}
Finally, UniPC \cite{unipc} reuses predictions from previous steps to correct the current update. However, since prediction tensors change shape across different stages, cached predictions cannot be reused across transitions. We therefore reset the state buffer at each stage transition and re-accumulate the required state within the new stage; empirically, this preserves sampling stability while avoiding cross-scale correction artifacts.

\subsection{Deep Compression Flow - From Step View}
% Real-Time Generation
\begin{figure*}[!t]
  \centering
  \includegraphics[width=0.93\linewidth]{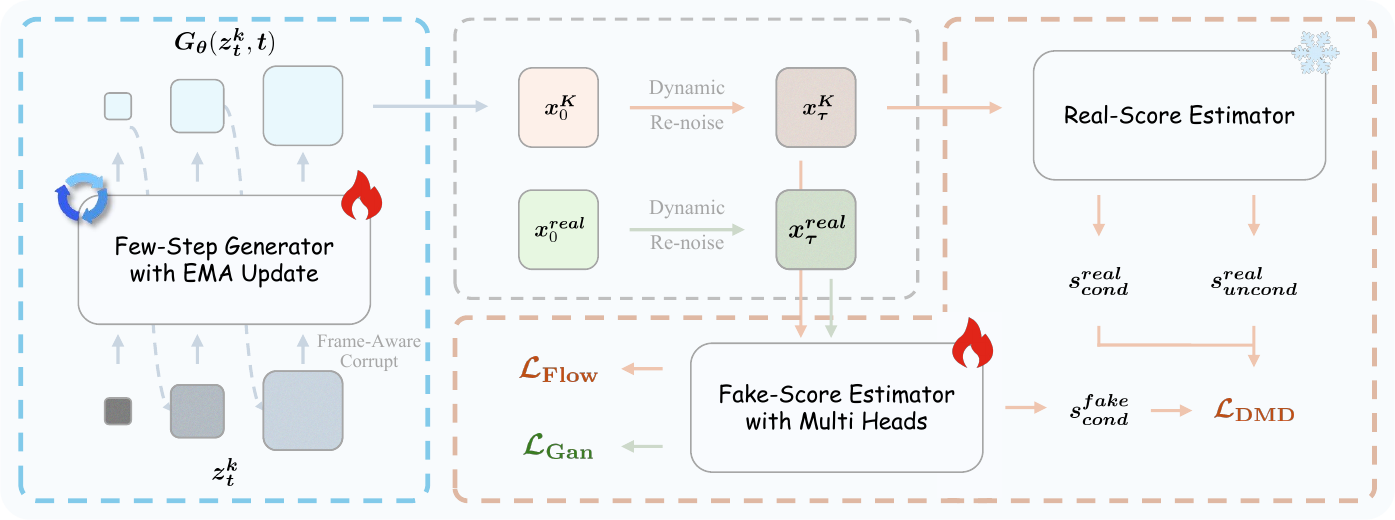}
  \caption{\textbf{Pipeline of Adversarial Hierarchical Distillation.} The framework is based on DMD \cite{DMD}, with improvements such as Pure Teacher Forcing, Staged Backward Simulation, Coarse-to-Fine Learning and Adversarial Post-Training.}
  \label{figure_distillation}
\end{figure*}

\subsubsection{Problem Formulation}
Step distillation is crucial for building real-time generative models. Among existing approaches \cite{rcm, LCM, sdxl-lightning, progressivedistillation}, Distribution Matching Distillation (DMD) \cite{DMD} is widely adopted and well established. In DMD, we first sample noise $\epsilon$ and feed it to a few-step generator $G_{\theta}$. Using $x_0$ prediction and backward simulation, the generator produces a clean sample $x_0$. We then sample a noise level $\lambda_{\tau} \sim \mathcal{U}[0,1]$ and perturb $x_0$ to obtain a noisy sample $x_{\tau}$. Next, we evaluate $x_{\tau}$ with a real-score estimator $p_{\text{real}}$ and a fake-score estimator $p_{\text{fake}}$, yielding scores $s_{\text{real}}$ and $s_{\text{fake}}$. The real score is computed via classifier-free guidance by combining conditional and unconditional predictions, i.e., $\mathrm{CFG}(s_{\text{real}}^{\text{cond}}, s_{\text{real}}^{\text{uncond}})$, whereas the fake score uses only the conditional branch, i.e., $s_{\text{fake}}^{\text{cond}}$. Their difference defines the distribution-matching gradient used to update $G_{\theta}$. In addition, we train $p_{\text{fake}}$ with a flow-matching loss $\mathcal{L}_{\text{Flow}}$ on $x_{\tau}$ to improve stability. However, \textit{\ours} changes the sampling procedure, so the standard pipeline is not directly applicable. We therefore propose Adversarial Hierarchical Distillation, a DMD-based framework (Figure~\ref{figure_distillation}) with the following improvements.

% Step distillation是构建实时生成模型的关键组成部分。其中，Distribution Matching Distillation (DMD) 是较为成熟的一种方案。给定噪声$\epsilon$，DMD首先将其输入 few-step generator $G_\theta$，通过 $x_0$-prediction与backward simulation得到对应的干净样本 $x_0$。随后，从区间 $[0,1]$ 上均匀采样噪声强度 $\lambda_\tau \sim \mathcal{U}[0,1]$，并将噪声注入$x_0$，得到带噪样本 $x_\tau$。接着，$x_\tau$ 分别输入 real-score estimator $p_{real}$ 与 fake-score estimator $p_{fake}$，得到真实分数与虚假分数。其中，真实分数$s^{real}$通过 classifier-free guidance 由条件与无条件分支组合得到$\mathrm{CFG}\big(s^{real}_{cond}},\, s^{real_{uncond})$。而虚假分数$s^{fake}$则为$s^{fake}_{cond}$。两者差值构成distribution matching gradient，用于更新$G_\theta$。与此同时，$p_{fake}$ 还需要对 $x_\tau$ 施加 flow matching loss $\mathcal{L}_{flow}$进行优化，以保证 fake-score estimator 的稳定性与有效性。然而，由于 \textit{\ours} 对 sampling 过程进行了修改，标准蒸馏流程无法直接适配。为解决这一问题，我们在 DMD 的基础上提出\textbf{Adversarial Hierarchical Distillation}（如图XXX所示），并引入如下改进：

\subsubsection{Adversarial Hierarchical Distillation}
\noindent\myparagraph{Pure Teacher Forcing with Autoregressive Teacher.}
Existing approaches \cite{selfforcing, selfforcing++, longlive, rollingforcing, rewardforcing} that apply DMD to real-time long-video generation typically discard real data entirely during training. For instance, Self-Forcing \cite{selfforcing} explicitly integrates the inference procedure of an autoregressive model into the training process: when generating the current section, previously generated sections are used as conditions to reduce the training--inference gap and mitigate drifting in video generation. However, we observe that the robustness of such methods against drifting is strongly dependent on the number of sections rolled out during training. Specifically, when training involves rollout of only five sections, the model frequently exhibits severe exposure bias during inference once the generated sequence exceeds this length. Motivated by this limitation, subsequent studies adopt a long self-rollout strategy \cite{selfforcing++, rollingforcing, longlive}, in which a large number of sections---corresponding to video durations of tens of seconds or even several minutes---are generated during training to enhance long-term stability. Nevertheless, this approach incurs substantial computational overhead, restricting existing methods to models with approximately 1.3B parameters \cite{wan}. To overcome this limitation, we employ real data exclusively as historical context during the distillation stage and require the generation of only a single section per training step, substantially improving training efficiency. Moreover, by incorporating the Easy Anti-Drifting mechanism proposed in Section~\ref{sec:easy_anti_drifting}, we achieve long-video anti-drifting performance comparable to that of long self-rollout strategies, without the need to perform such expensive rollouts. More importantly, we select \textit{\ours-Base}\footnote{We describe the details of \ours-Base in Section~\ref{sec:implementation}.} as the teacher model because it is already capable of generating high-quality long videos, whereas existing methods typically rely on Wan \cite{wan}, which is limited to synthesizing short videos.

% 现有将 DMD 应用于实时长视频生成模型的方法，通常在训练阶段完全移除对 ground-truth data 的使用，即不再将其作为 historical context。例如，Self-Forcing 方法将自回归模型的推理过程显式引入训练阶段：在生成当前 section 时，以之前生成的 sections 作为条件，从而试图缩小 training–inference gap 并缓解视频生成中的 drifting 问题。然而，我们观察到，此类方法的抗退化能力与训练阶段 rollout 的 section 数量高度相关。例如，当训练阶段仅 rollout 5 个 sections 时，模型在推理过程中往往会在超过 5 个 sections 后出现显著的 exposure bias。基于这一现象，后续工作采用 long self-rollout 策略，即在训练阶段生成与数十秒甚至数分钟视频长度相对应的大量 sections，以提升长时稳定性。然而，该策略带来了极高的计算开销，使得相关方法通常只能受限于约 1.3B 参数规模的 baseline 模型。为解决上述问题，我们在蒸馏阶段仅使用 ground-truth data 作为 historical context，并且每个训练 step 只需生成单个 section，极大提高了训练效率。同时，我们结合 Section \ref{sec:easy_anti_drifting} 中提出的 Easy Anti-Drifting 机制，在不引入 long self-rollout 的情况下，实现了与其相当的长视频抗退化效果。更重要的是我们选取\textit{\ours-Base}作为teacher model，which已经具备生成高质量长视频的能力，而不是像现有方法一样选择只能合成短视频的Wan2.1模型。

\noindent\myparagraph{Staged Backward Simulation.}
DMD performs backward simulation on a single flow trajectory to recover $x_0$. In contrast, we introduce Staged Backward Simulation, which decomposes the backward simulation into $K$ stages, producing intermediate estimates $\{x_0^k\}_{k=1}^K$; the final-stage output $x_0^K$ is used as $x_0$. At stage $k$, given the current state $x_t^k$ and the predicted velocity field $u_\theta^k(x_t^k, y, \lambda_t, k)$, we estimate the terminal state as
\begin{equation}
  x_0^k = x_t^k - \lambda_t \cdot u_\theta^k(x_t^k, y, \lambda_t, k),
  \label{eq: get_x0}
\end{equation}
where the update follows directly from the linear interpolation path in Eq.~\ref{eq: get_xt}. We then reconstruct $x_t^k$ using Eq.~\ref{eq: get_xt} and re-estimate $x_0^k$ via Eq.~\ref{eq: get_x0}, repeating this procedure until stage $k$ converges. The resulting estimate $x_0^k$ initializes stage $(k+1)$. After $K$ stages, we obtain $x_0 = x_0^K$.

% 不同于DMD在单一的全尺寸flow上进行backward simulation即可直接得到$x_0$，我们提出Staged Backward Simulation, which我们在$K$个阶段上分别执行backward simulation，得到一系列中间结果 $\{x_0^k\}_{k=1}^K$，并将最后一个阶段的输出$x_0^K$作为$x_0$。
% 具体而言，在第$k$个阶段的backward simulation过程中，当前状态$x_t^k$以及对应的速度场$u_\theta^k(x_t^k,\, y,\, \lambda_t,\, k)$ 均为已知。基于 Eq.~AAA 与 Eq.~BBB，可推导得到该阶段对应的最终状态估计：
% \begin{equation}
% x_0^k = x_t^k - \lambda_t \cdot u_\theta^k(x_t^k,\, y,\, \lambda_t,\, k) .
% \end{equation}
% 在得到$x_0^k$后，我们根据Eq.~BBB对$x_t^k$进行重新构造，并进一步结合Eq.~CCC再次计算$x_0^k$。通过重复执行上述过程，即可完成第$k$个阶段的 backward simulation。随后，将得到的$x_0^k$作为下一阶段初始状态估计，并在第 $k+1$ 个阶段中重复相同的 backward simulation 流程。经过 $K$ 个阶段的迭代后，最终得到$x_0^K$，并将其作为模型的最终输出。

\noindent\myparagraph{Coarse-to-Fine Learning.}
Compared with DMD, \textit{\ours} propagates gradients through $K$ stages and multiple flow trajectories, which increases optimization difficulty and may slow convergence, especially early in training. We therefore adopt three curriculum-style strategies that progressively increase task difficulty. \textit{(1) Staged ODE Init.} We construct a compact dataset of ODE solution pairs generated by \textit{\ours-Mid}, which is then used for initialization following \cite{causvid}. In contrast to prior work, our initialization is performed across $K$ stages. At each stage, only a single section needs to be generated rather than multiple sections, and an autoregressive teacher is employed to guide the process. \textit{(2) Dynamic Re-noise.} Uniformly sampling noise levels, as in standard DMD, is suboptimal in the hierarchical setting because different noise regimes contribute differently across training stages. Inspired by \cite{z_image, decoupleddmd}, we sample timesteps from a Beta distribution whose parameters follow a cosine decay schedule: it concentrates on high-noise timesteps early to learn coarse structure and becomes increasingly uniform later to emphasize medium- and low-noise timesteps for detail refinement.

% 不同于 DMD 仅包含单一且梯度路径较为简单的 flow 轨迹，\textit{\ours}的梯度传播需要覆盖 $K$ 个阶段的多条 flow 轨迹。虽然这种层级化设计有助于建模多阶段一致性，但也显著增加了优化难度，容易导致收敛变慢和训练不稳定，尤其是在训练初期。为此，我们针对设计了两种动态训练策略，通过从易到难的方式逐步降低优化难度，从而加速模型收敛并提升整体训练稳定性。
% \textbf{(1) Dynamic Length Section.} 在训练初期，模型对长程上下文和多latent-section之间的依赖关系建模能力有限，若直接引入较长的序列，容易造成梯度不稳定。基于这一观察，我们采用渐进式训练策略，通过动态调整latent-section数量来逐步增加任务难度。具体而言，在每个训练步中，latent-section数量从Beta分布中采样。在训练初期，模型更倾向于采样较少的latent-section，优先学习短程一致性；随着训练推进，采样分布逐渐向更长序列扩展，使模型能够逐步掌握跨多阶段的长程一致性建模能力。

\noindent\myparagraph{Adversarial Post-Training.}
DMD distills a multi-step teacher into a few-step student by matching the teacher-defined distribution; consequently, the student inherits the teacher's biases and is bounded by the teacher's expressive capacity. To relax this limitation, inspired by Spark-Wan \cite{spark_wan} and DMD2 \cite{DMD2}, we augment distillation with an additional GAN objective trained on real data. This auxiliary objective provides teacher-independent supervision and can further improve sample quality.

Concretely, we add multi-granularity classification branches $D$ to $p_{\text{fake}}$ and distribute them across DiT layers. We train these branches with the non-saturated GAN objective:
\begin{equation}
  \mathcal{L}_{D} = \mathbb{E} \left[\log D(x_\tau^{\text{real}}, \tau)\right] + \mathbb{E} \left[-\log D(x_\tau^K, \tau)\right].
\end{equation}
To stabilize discriminator training, we incorporate an approximate R1 regularizer (following APT \cite{apt}):
\begin{equation}
  \mathcal{L}_{\text{aR1}} = \left| D(x_\tau^{\text{real}}, \tau) - D(\mathcal{N}(x_\tau^{\text{real}}, \sigma_D \mathbf{I}), \tau) \right|_2^2.
\end{equation}
Finally, the full adversarial objectives are defined as:
\begin{equation}
  \mathcal{L}_{D} = \mathbb{E} \left[\log D(x_\tau^{\text{real}}, \tau)\right] + \mathbb{E} \left[-\log D(x_\tau^K, \tau)\right] + \lambda_D \cdot \mathbb{E} \left[\left| D(x_\tau^{\text{real}}, \tau) - D(\mathcal{N}(x_\tau^{\text{real}}, \sigma_D \mathbf{I}), \tau) \right|_2^2\right],
\end{equation}
\begin{equation}
  \mathcal{L}_{G} = \mathbb{E} \left[\log D(x_\tau^K, \tau)\right].
\end{equation}
In practice, we set $\lambda_D = 100$ and $\sigma_D = 0.1$. To reduce memory usage, we feed the discriminator a random crop of size $H' \times W'$ from $x_\tau^K$ instead of the full-resolution sample, where $H'=\frac{H}{2}$ and $W'=\frac{W}{2}$.

\noindent\myparagraph{Other Details.}
We initialize the few-step generator from \textit{\ours-Mid} and inherit both the real-score and fake-score estimators from \textit{\ours-Base}, which provides stable, high-fidelity supervision during training.

Based on this design, the objectives for Adversarial Hierarchical Distillation are:
\begin{equation}
  \mathcal{L}_{{G}_\theta} = \mathcal{L}_{\mathrm{DMD}} + w_G \cdot \mathcal{L}_{G},
\end{equation}
\begin{equation}
  \mathcal{L}_{p_{\text{fake}}} = \mathcal{L}_{\mathrm{Flow}} + w_D \cdot \mathcal{L}_{D}.
\end{equation}
Here, we set $w_G=5e-2$ and $w_D=1e-2$ as the weight coefficients for the respective losses. We follow CausVid \cite{causvid} and update $G_\theta$ only once every five updates of $p_{\text{fake}}$.

% Few-step generator的初始化权重继承自 \textit{\ours}，而real-score estimator与fake-score estimator则沿用baseline，从而在训练提供稳定且高保真度的监督信号。在奖励建模方面，我们选用 VideoAlign 作为 reward model，并将其中的 Motion Quality（MQ）、Visual Quality（VQ）以及 Text Alignment（VA）三个子指标进行加权求和平均，作为最终的reward score。基于上述设计，所提出的 Adversarial Reward-weighted Distillation 的整体优化目标定义为：
% \begin{equation}
%     \mathcal{L}_{\mathrm{HCD-G}} = \mathcal{R}_{\mathrm{rl}} \cdot \left( \mathcal{L}_{\mathrm{DMD}} + w_{mean-var} \cdot \mathcal{L}_{\mathrm{mean\text{-}vae}} + w_{G} \cdot \mathcal{L}_{\text{G}} \right),
% \end{equation}
% \begin{equation}
%     \mathcal{L}_{\mathrm{HCD-D}} = \mathcal{L}_{\mathrm{flow}} + w_{D} \cdot \mathcal{L}_{\mathrm{D}}.
% \end{equation}
% 其中, $w_{mean-var}$、$w_{G}$和$ w_{D}$ 分别表示各个损失的权重系数, $\mathcal{R}_{\mathrm{rl}}$对应的reward score定义为：
% \begin{equation}
%     \frac{w_{mq} \cdot MQ + w_{vq} \cdot VQ + w_{va} \cdot VA}{3},
% \end{equation}
% 其中 $w_{mq}$、$w_{vq}$ 和 $w_{va}$ 分别表示各奖励子项的权重系数，用于平衡不同维度的质量信号。需要注意的是，我们只有在训练后期才启用Reward Post-Training和Adversarial Post-Training。我们跟随XXX，每更新5次$p_{fake}$才更新一次$G_\theta$。

\subsection{Other Techniques}
In this section, we present several inference techniques that are training-free and parameter-free.

% In this section，介绍若干推理阶段使用的技巧，这些方法不引入额外参数，也无需训练。

\noindent\myparagraph{Adaptive Sampling.}
Figure~\ref{figure_degeneration_score} shows that drifting is accompanied by pronounced shifts in RGB statistics (mean and variance). Since the latent space is a compressed representation of RGB space, analogous distribution shifts also appear in latent statistics. This motivates an adaptive anti-drifting strategy. Let the RGB mean and variance of the $t$-th generated latent section be $\mu_t$ and $\sigma_t^2$. During inference, we maintain global statistics (global mean and variance), denoted by $\bar{\mu}_t$ and $\bar{\sigma}_t^2$, updated via an exponential moving average (EMA):
\begin{equation}
  \bar{\mu}_t = \rho_\mu \bar{\mu}_{t-1} + (1-\rho_\mu)\mu_t,
\end{equation}
\begin{equation}
  \bar{\sigma}_t^2 = \rho_\sigma \bar{\sigma}_{t-1}^2 + (1-\rho_\sigma)\sigma_t^2,
\end{equation}
where $\rho_\mu,\rho_\sigma \in (0,1)$ are smoothing coefficients. When the statistics of the current latent section deviate from the global statistics beyond preset thresholds $\delta_\mu$ and $\delta_\sigma$:
\begin{equation}
  \left\lVert \mu_t - \bar{\mu}_t \right\rVert_2 > \delta_\mu
  \quad \text{and} \quad
  \left\lVert \sigma_t^2 - \bar{\sigma}_t^2 \right\rVert_2 > \delta_\sigma,
\end{equation}
we treat the current section as exhibiting significant drift. When generating the next section, we apply Frame-Aware Corrupt to the historical context, perturbing the drifting frames in a targeted, training-free manner. This implicitly reduces the model's reliance on the biased history and encourages it to rely more on its intrinsic generative prior, thereby improving long-video quality and stability.

% 图~XXX 表明，当视频生成过程中出现drifting时，其对应的 RGB 空间统计量（均值与方差）会发生显著漂移。由于潜变量空间（latent space）本质上是对 RGB 空间的压缩表示，这种漂移会同步反映到 latent 的统计分布上，即其均值与方差同样发生变化，这启发我们可以设计自适应的Anti-Drifting策略。记第 $t$ 个生成 latent-section 的 RGB 均值与方差分别为$\mu_t,\;\sigma_t^2$，我们在推理阶段维护一组全局统计量（Global Mean 与 Global Variance）：$\bar{\mu}_t,\;\bar{\sigma}_t^2$，并通过指数滑动平均（EMA）进行更新：
% \begin{equation}
%     \bar{\mu}_t = \rho_\mu\, \bar{\mu}_{t-1} + (1-\rho_\mu)\, \mu_t,
% \end{equation}
% \begin{equation}
%     \bar{\sigma}_t^2 = \rho_\sigma\, \bar{\sigma}_{t-1}^2 + (1-\rho_\sigma)\, \sigma_t^2,
% \end{equation}
% 其中 $\rho \in (0,1)$ 为平滑系数。当当前 latent-section 的统计量偏离全局统计量超过预设阈值$\delta_\mu$和$\delta_\sigma$：
% \begin{equation}
%     \lVert \mu_t - \bar{\mu}_t \rVert_2 > \delta_\mu
%     \quad \text{and} \quad
%     \lVert \sigma_t^2 - \bar{\sigma}_t^2 \rVert_2 > \delta_\sigma,
% \end{equation}
% 我们认为该 latent-section 可能引入了显著的 drifting。在此情况下，在生成下一个 latent-section 时，我们对 historical context 施加Frame-Aware Corrupt操作。该操作以无参数的方式对包含 drifting 的历史帧进行有针对性的扰动，从而向模型隐式传达“降低对当前 historical context 依赖”的信号，引导模型更多依赖自身生成能力而非偏置上下文，最终提升长视频生成的整体质量与稳定性。

\noindent\myparagraph{Interactive Interpolation.}
Long-video generation enables interactive editing, where users can revise the prompt on the fly, requiring the model to adapt rapidly without introducing temporal artifacts. A naive solution is to switch to the new prompt embedding abruptly; however, this induces an instantaneous conditional shift and often causes visible discontinuities (e.g., flicker or sudden semantic jumps) around the editing boundary. Following Krea \cite{krea_realtime_14b}, we instead adopt prompt interpolation, which gradually transitions the conditioning from the current to the target prompt over multiple steps. This yields a smoother handover between conditions and improves perceived temporal coherence. Specifically, let the current prompt embedding be $\mathbf{e}^{(1)} \in \mathbb{R}^{\ell_{\text{Text}} \times D}$ and the target embedding be $\mathbf{e}^{(2)} \in \mathbb{R}^{\ell_{\text{Text}} \times D}$, where $\ell_{\text{Text}}$ is the text length and $D$ is the hidden dimension. We construct $M$ intermediate conditions $\{\mathbf{e}^{[j]}\}_{j=0}^{M-1}$ by linear interpolation:
\begin{equation}
  \mathbf{e}^{[j]} = (1-\lambda_j)\mathbf{e}^{(1)} + \lambda_j\mathbf{e}^{(2)},
  \quad
  \lambda_j = \frac{j}{M-1},
  \quad
  j = 0,1,\ldots,M-1,
\end{equation}
where $\lambda_j \in [0,1]$ increases linearly with $j$, so that $\mathbf{e}^{[0]}=\mathbf{e}^{(1)}$ and $\mathbf{e}^{[M-1]}=\mathbf{e}^{(2)}$. During generation, we feed these embeddings sequentially to gradually move the conditioning from $\mathbf{e}^{(1)}$ to $\mathbf{e}^{(2)}$ over $M$ steps. This gradual transition mitigates visual and semantic discontinuities caused by abrupt conditioning changes.

% 长视频生成使得交互式生成成为可能，其核心要求模型能够根据用户prompt的变化及时调整生成内容。一个直观方案是直接将原有 prompt embedding 替换为新的 prompt embedding，然而该做法会在切换时引入显著的条件突变，从而导致生成内容出现不自然的跳变。为缓解这一问题，我们受Krea的启发，引入prompt interpolation：不直接替换条件embedding，而是在当前条件与目标条件之间构造一条平滑的过渡路径，使模型能够逐步适应新的prompt。
% 具体而言，设当前prompt embedding为$\mathbf{e}^{(1)} \in \mathbb{R}^{\ell_\text{Text} \times D}$，目标prompt embedding为$\mathbf{e}^{(2)} \in \mathbb{R}^{\ell_\text{Text} \times D}$，其中$\ell_\text{Text}$表示text的序列长度、$D$为隐藏维度。我们通过线性插值构造一个包含$M$个中间条件的序列$\{\mathbf{e}^{[j]}\}_{j=0}^{M-1}$：
% \begin{equation}
% \mathbf{e}^{[j]} = (1-\lambda_j)\,\mathbf{e}^{(1)} + \lambda_j\,\mathbf{e}^{(2)},
% \quad
% \lambda_j = \frac{j}{M-1},
% \quad
% j = 0,1,\ldots,M-1,
% \end{equation}
% 其中$\lambda_j \in [0,1]$为插值权重，随索引$j$线性递增，因此该序列满足$\mathbf{e}^{[0]}=\mathbf{e}^{(1)}$ 且 $\mathbf{e}^{[M-1]}=\mathbf{e}^{(2)}$。在生成过程中，我们按顺序使用该插值序列逐步更新模型的条件输入，即在连续的若干生成步中将prompt embedding从$\mathbf{e}^{[0]}$平滑过渡到$\mathbf{e}^{[M-1]}$。当序列遍历结束后，模型完全切换至目标prompt并继续生成后续内容。该策略通过将条件变化分解为多个小幅度更新，有效避免了条件突变带来的视觉/语义跳变，从而使交互式编辑过程更加自然流畅。

% \noindent\myparagraph{History Cache.}
% Pass

% 由于Guidance Self Attention限制了$X_{\text{hist}}$只与自身交互，并且其timestep始终固定为0，这意味着当$X_{\text{hist}}$保持不变时，每次前向传播过程中计算得到的$\text{Attention}(Q'', K'', V'')$、$K''$和$V''$都是完全相同的。因此，我们在推理时可以只计算一次这些结果，然后在后续的所有去噪步骤中直接复用，这就是History Cache的核心思想。以常用的50步推理过程为例，通过History Cache机制，我们能够将$X_{\text{hist}}$部分的计算量减少至原来的1/50，相当于减少了49倍的冗余计算。

\section{Infrastructure}\label{sec:infrastructure}

\subsection{Workload Analysis}
Scaling DiT-based video generators to 14B parameters yields prohibitive compute and memory costs, even with batch size 1. The primary bottleneck is the quadratic complexity of 3D attention over temporal and spatial tokens. In practice, training such models on a single GPU typically requires extensive parallelism (\eg, CP, TP, SP) and parameter/activation sharding (\eg, FSDP, DeepSpeed).

In contrast, \textit{\ours} introduces Deep Compression Flow to compress both the historical and noisy contexts, enabling full forward and backward passes on a single GPU for the first two training stages, without parallelism or sharding. In Multi-Term Memory Patchification, we set $(p_t^{(1)},p_t^{(2)},p_t^{(3)})=(4,2,1)$, $(p_h^{(1)},p_h^{(2)},p_h^{(3)})=(8,4,2)$, and $(p_w^{(1)},p_w^{(2)},p_w^{(3)})=(8,4,2)$, with $(T_1,T_2,T_3)=(16,2,2)$. This reduces the historical-context token count from $5HW$ to $\frac{5}{8}HW$ (approximately $8\times$). With Pyramid Unified Predictor Corrector ($K=3$), the noisy-context token count decreases from $NHW$ to $\frac{7}{16}NHW$ (approximately $2.29\times$).

For a standard DiT, the per-layer complexity is approximately $\mathcal{O}(\alpha B \ell D^2 + \beta B \ell^2 D)$, where $\alpha$ and $\beta$ are the costs of linear layers and attention, respectively, and $\ell$ is the sequence length (the $\ell^2$ term dominates self-attention). The overall complexity scales as $\mathcal{O}(L(\alpha B \ell D^2 + \beta B \ell^2 D))$, while activation memory scales as $\mathcal{O}(\gamma L B \ell D)$, where $\gamma$ depends on the implementation. Therefore, the $8\times$ and $2.29\times$ token reductions translate to roughly $64\times$ and $5.2\times$ reductions in attention FLOPs for the historical and noisy contexts, respectively, and they linearly reduce activation and intermediate-state memory.

% 对于训练参数规模为14B的DiT视频生成模型，即使在batch size = 1的情况下，由于3D Attention在时间与空间维度上的二次复杂度，显存与计算开销依然极为高昂。实践中，社区通常需要依赖多种并行策略（如 CP、TP、SP）以及参数/激活分片框架（如 FSDP、DeepSpeed），才能勉强解决单卡显存不足、无法完成训练的问题。
% 相比之下，\textit{\ours}通过引入Deep Compression Flow，对historical context与noise context进行系统性压缩，对于第一阶段和第二阶段的训练,在不依赖任何并行或sharding基础设施的前提下，即可在单卡上完成完整的forward与backward。
% 具体而言，在Multi-Term Memory Patchification中，参数设置为$(p_t^{(1)},p_t^{(2)},p_t^{(3)})=(4,2,1)$，$(p_h^{(1)},p_h^{(2)},p_h^{(3)})=(8,4,2)$，$(p_w^{(1)},p_w^{(2)},p_w^{(3)})=(8,4,2)$，history contex的长度为$(T_1,T_2,T_3)=(16,2,2)$，该设计将 historical context的token数从原始的 $5\times H\times W$压缩至$\frac{5}{8}\times H\times W$，实现约$8\times$的序列长度缩减；在Pyramid Unified Predictor-Corrector中，参数设置为$K=3$，将noise context的token数从$N\times H\times W$降低至$\frac{7}{16}\times N\times H\times W$，对应约$2.29\times$的压缩比例。
% 对于标准DiT，其单层计算复杂度可近似表示为$\mathcal{O}(\alpha \times B \times \ell \times D^2 + \beta \times B \times \ell^2 \times D)$，其中，$\alpha$对应于线性层的成本，$\beta$对应于注意力层的成本，$\ell$为序列长度，且$\ell^2$项主导自注意力计算；整体计算复杂度随层数总层数$L$线性放大为$\mathcal{O}(L(\alpha \times B \times \ell \times D^2 + \beta \times B \times \ell^2 \times D))$，而显存占用可近似表示为$\mathcal{O}(\gamma \times L \times B \times \ell \times D)$,其中γ 取决于 DiT 层的实现方式。因此，\textit{\ours}对historical context的$8\times$压缩与对noise context的$2.29\times$压缩分别带来了约$64\times$与$5.2\times$的attention FLOPs降低，并使DiT激活与中间状态的显存消耗随token数线性下降。

\subsection{Memory Optimization}
In the first two training stages, the GPU needs to load only three components: the VAE, the text encoder, and the DiT. By offloading VAE latents and text embeddings to disk, the GPU effectively retains only the DiT, enabling single-GPU batch sizes comparable to those used for image diffusion models.

In the third stage, the memory demand increases substantially: the GPU must host four 14B models (the few-step generator, real-score estimator, fake-score estimator, and EMA model), as well as multiple GAN heads. Under an 80GB memory budget, this configuration can exceed capacity even for inference, and training further increases memory usage due to activations and intermediate states. We therefore adopt the following strategies to enable training under strict memory constraints.

\begin{figure*}[!t]
  \centering
  \includegraphics[width=0.80\linewidth]{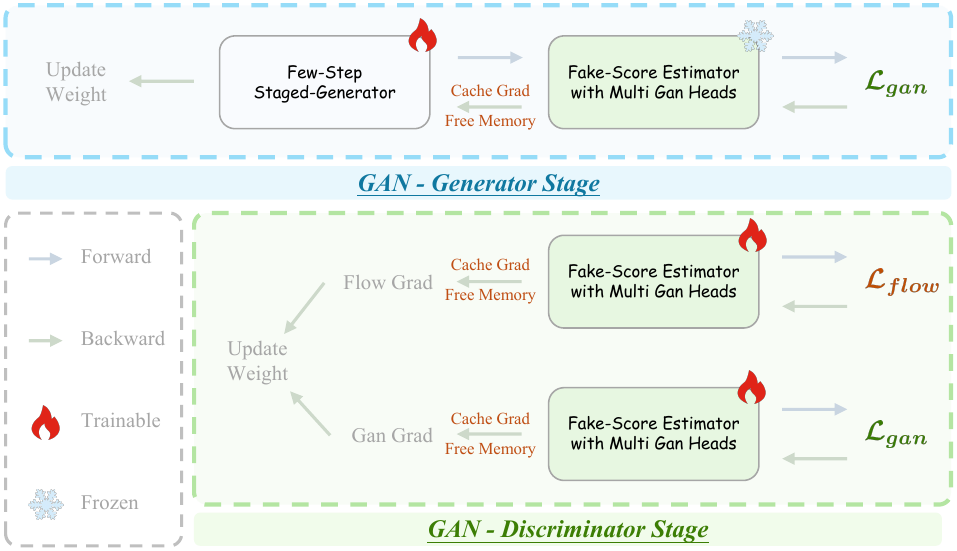}
  \caption{\textbf{Execution of Cache Grad for GAN.} We cache discriminator gradients w.r.t. inputs to decouple backpropagation and free intermediate activations early, substantially reducing peak memory.}
  \label{figure_cache_grad}
\end{figure*}

\noindent\myparagraph{Sharded EMA.}
Exponential moving average (EMA) stabilizes training by smoothing parameter updates and is typically stored in FP32 for numerical robustness. A naive implementation replicates the full FP32 EMA copy on every GPU, which incurs substantial memory overhead. Following OpenSora-Plan \cite{opensoraplan}, we instead shard the EMA parameters across GPUs using ZeRO-3, so that each device stores only a fraction of the EMA states. For a 14B-parameter model sharded over $Z$ GPUs, each device stores approximately $\frac{14\times 4}{Z}$~GiB of EMA parameters. This removes redundant replicas and improves memory efficiency.

% Exponential Moving Average（EMA）在训练过程中用于对模型参数进行时间维度上的平滑，从而缓解权重更新过程中的震荡。为保证数值稳定性，EMA 参数通常以 FP32 格式存储。然而，在常规实现中，每个GPU都需要额外维护一份与相同的FP32 EMA副本，这会带来显著的显存开销。为此，我们对 EMA 模型采用 ZeRO-3策略进行GPU端分片存储。借助 ZeRO-3，每个GPU仅需持有完整FP32 EMA参数的一小部分，而非完整副本，从而显著提升显存利用效率。具体而言，对于一个14B的模型，当其被分片到$Z$个GPU上时，每张GPU仅需存储约$\frac{14 \times 4}{Z}$ GiB 的EMA参数。这种分片方式不仅有效消除了 EMA 带来的冗余显存占用，使得计算成本随着GPU数量的增加而线性下降。

\noindent\myparagraph{Asynchronous VRAM Freeing.}
In stage-3 training, we sequentially execute multiple large models: we feed noise $z_t$ to the few-step staged generator to obtain $x_0^{\text{staged}}$, and then re-noise and evaluate the sample with the real-score estimator $p_{\text{real}}$ and fake-score estimator $p_{\text{fake}}$ to compute $\mathcal{L}_{\mathrm{DMD}}$, $\mathcal{L}_{\mathrm{GAN}}$, and $\mathcal{L}_{\mathrm{Flow}}$. Because these computations are serialized, only one model needs to reside on the GPU at a time during the forward pass.

Moreover, under the two time-scale update rule (TTUR) \cite{DMD}, each iteration updates either the fake-score estimator $p_{\text{fake}}$ (via $\mathcal{L}_{\mathrm{Flow}}$ and $\mathcal{L}_{\mathrm{GAN}}$) or the few-step generator ${G}_\theta$ (via $\mathcal{L}_{\mathrm{DMD}}$). We exploit this structure to asynchronously offload unused models to host memory, limiting peak VRAM to roughly that of training a single 14B model. With pinned host memory, non-blocking transfers, and careful CPU--GPU scheduling, we maintain throughput close to GPU-only execution despite frequent transfers.

% 一个完整的第三阶段训练流程如下：噪声$z_t$输入给few-step staged generator，生成$x_0^{staged}$。接着，$x_0^{staged}$分别输入到VAE和Reward Model，得到$L_{rl}$；之后，$x_0^{staged}$经过renoise处理，再输入到real-score estimator和fake-score estimator，分别得到$\mathcal{L}_{\mathrm{DMD}}$和$\mathcal{L}_{\mathrm{Flow}}$。可以看到，各个模型的运行是串行进行的，因此在forward过程中，同一时间内GPU上只需要加载一个模型即可。此外，由于使用了Two Time-scale Update Rule（TTUR），在同一时间内，我们只需要使用$\mathcal{L}_{\mathrm{Flow}}$来更新fake-score estimator，或者使用$\mathcal{L}_{\mathrm{DMD}}$和$L_{rl}$来更新few-step staged generator。因此，这种方式使得显存峰值与训练一个14B参数的模型相当。更重要的是，通过使用pin_memory操作将模型放入物理内存、非阻塞传输（non-blocking）以及精心设计CPU-GPU的传输位置，即使在频繁进行CPU-GPU传输的情况下，训练速度仍然保持与仅在GPU上训练时相似，训练效率接近于无损。

\noindent\myparagraph{Cache Grad for GAN.}
\textit{(1) Updating the generator.} Standard autodiff requires keeping all activations of the few-step generator ${G}_\theta$ and the fake-score estimator $p_{\text{fake}}$ until backpropagation finishes, which can make the subsequent computation of $\mathcal{L}_{\text{DMD}}$ infeasible under memory limits. We therefore decouple the fake-score estimator $p_{\text{fake}}$ from the default backward pass by caching the discriminator gradient with respect to its input during the forward pass. Specifically, we immediately free the estimator's intermediate activations after its forward pass and reuse the cached input gradients during backpropagation, avoiding the need to retain the full computation graph. This reduces peak memory to that of a single 14B model.

\textit{(2) Updating the fake-score estimator.} We combine gradient accumulation with batched execution. We first compute $\mathcal{L}_{\mathrm{Flow}}$ in a separate forward/backward pass, accumulate its gradients into the discriminator, and immediately release its activations. We then concatenate the real, fake, and perturbed samples (for $\mathcal{L}_{\text{aR1}}$) into one batch and run a single forward/backward pass to compute the remaining loss terms. Compared with jointly computing all losses, this scheduling substantially reduces peak memory.

% \textbf{(1)} 在 GAN 训练中，当对 few-step generator 进行优化时，传统自动微分机制需要在前向传播阶段持续保留 few-step generator 与 fake-score estimator 的全部中间激活，直至反向传播结束才能释放。这导致显存峰值过高,使得后续无法执行获取 $\mathcal{L}_{\text{DMD}}$ 所需的步骤。为缓解这一问题，我们设计了一种自定义反向传播机制，将 fake-score estimator 的梯度计算与标准反向传播过程解耦：在前向传播阶段即计算并缓存判别器关于输入的梯度，并在完成后立即释放 fake-score estimator 的中间激活；在反向传播阶段则直接复用缓存的梯度，从而避免构建和保留完整计算图。借助该策略，判别器在前向传播结束后即可从 GPU 中卸载，使整体显存峰值降低至仅与单个 14B 参数模型相当。
% \textbf{(2)} 当对 fake-score estimator 进行优化时，我们进一步采用了梯度累积与批量前向传播相结合的显存优化策略。具体而言，首先对去噪损失单独进行前向与反向传播，并将对应梯度累积到判别器参数中，从而在完成该步骤后立即释放相关中间激活；随后，将真假样本以及 R1、R2 正则化所需的扰动样本合并为一个批次进行统一的前向传播，并执行一次反向传播以获取剩余损失项的梯度。相较于对所有损失项同时进行前向与反向传播的常规做法，该策略显著降低了训练过程中的显存峰值。

\subsection{Efficiency Optimization}
To further accelerate training and inference, we replace multiple native PyTorch operations with custom implementations spanning both forward and backward propagation, thereby improving computational efficiency.

% 为了进一步加快训练和推理过程，我们重新实现了多个操作替换Pytorch原生实现，包括前向和反向传播过程，从而提高了效率。

\begin{table}[!t]
  \centering
  \caption{\textbf{The detailed training hyperparameters of Stage-1 and Stage-2.}}
  \label{tab:training_hyperparameters-1-2}
  \resizebox{\textwidth}{!}{
    \begin{tabular}{lcccc}
      \toprule
      \textbf{Configuration} & \textbf{Stage-1-init} & \textbf{Stage-1-post} & \textbf{Stage-2-init} & \textbf{Stage-2-post} \\
      \midrule
      Global Batch Size & 128 & 128 & 256 & 192 \\
      Optimizer & \makecell[c]{AdamW, $\beta_1=0.9$, $\beta_2=0.999$ \\ $\epsilon=1e-08$, weight\_decay=1e-04} & \makecell[c]{AdamW, $\beta_1=0.9$, $\beta_2=0.999$ \\ $\epsilon=1e-08$, weight\_decay=1e-04} & \makecell[c]{AdamW, $\beta_1=0.9$, $\beta_2=0.999$ \\ $\epsilon=1e-08$, weight\_decay=1e-04} & \makecell[c]{AdamW, $\beta_1=0.9$, $\beta_2=0.999$ \\ $\epsilon=1e-08$, weight\_decay=1e-04} \\
      Learning Rate & 5e-5 & 3e-5 & 1e-4 & 3e-5 \\
      Learning Rate Schedule & Constant & Constant & Constant & Constant \\
      Training Steps & 5.5k & 7.5k & 16k & 20k \\
      Gradient Clipping & 1.0 & 1.0 & 1.0 & 1.0 \\
      LoRA Rank & 128 & 128 & 256 & 256 \\
      LoRA Alpha & 128 & 128 & 256 & 256 \\
      Numerical Precision & BFloat16 & BFloat16 & BFloat16 & BFloat16 \\
      GPU Usage & 64 NVIDIA H100 & 64 NVIDIA H100 & 64 NVIDIA H100 & 64 NVIDIA H100 \\
      History Corrupt Ratio & \multicolumn{4}{c}{$p_a=0.0, p_b=0.8, p_c=0.1, p_d=0.1; b_{\min}=0, b_{\max}=0.33, c_{\min}=0, c_{\max}=0.1$} \\
      % \midrule
      % History Size & (16, 2, 1) & (16, 2, 1) & (16, 2, 1) & (16, 2, 1) \\
      % History Compression Ratio & (256, 32, 8) & (256, 32, 8) & (256, 32, 8) & (256, 32, 8) \\
      % Pyramid Stage & - & - & 3 & 3 \\
      % Noise Corrupt Ratio & 0 $\sim$ 1/3 & 0 $\sim$ 1/3 & 0 $\sim$ 1/3 & 0 $\sim$ 1/3 \\
      % Resolution Corrupt Ratio & 0 $\sim$ 1/10 & 0 $\sim$ 1/10 & 0 $\sim$ 1/10 & 0 $\sim$ 1/10 \\
      \bottomrule
    \end{tabular}
  }
\end{table}

\begin{table}[!t]
  \centering
  \caption{\textbf{The detailed training hyperparameters of Stage-3.}}
  \label{tab:training_hyperparameters-3}
  \resizebox{\textwidth}{!}{
    \begin{tabular}{lcc}
      \toprule
      \textbf{Configuration} & \textbf{Stage-3-ode} & \textbf{Stage-3-post} \\
      \midrule
      Global Batch Size          & 128 & 128 \\
      Real-score initialization  & - & \ours-Base \\
      Fake-score initialization  & - & \ours-Base \\
      Real-score CFG weight      & - & 3.0 \\
      Optimizer ($G_\theta$ \& $p_{fake}$)   & \makecell[c]{AdamW, $\beta_1=0.0$, $\beta_2=0.999$ \\ $\epsilon=1e-08$, weight\_decay=1e-03} & \makecell[c]{AdamW, $\beta_1=0.0$, $\beta_2=0.999$ \\ $\epsilon=1e-08$, weight\_decay=1e-03} \\
      Learning Rate $(G_\theta)$                        & 2.0e-6 & 2.0e-6 \\
      Learning Rate $(p_{fake})$                        & -      & 4.0e-7 \\
      Learning Rate Schedule ($G_\theta$ \& $p_{fake}$) & Constant & Constant \\
      Learning Rate Warmup Step  & -    & -    \\
      Gradient Clipping ($G_\theta$ \& $p_{fake}$) & 10.0 & 10.0 \\
      LoRA Rank $(G_\theta)$     & 256  & 256  \\
      LoRA Rank $(p_{fake})$     & -    & 256  \\
      LoRA Alpha $(G_\theta)$    & 256  & 256  \\
      LoRA Alpha $(p_{fake})$    & -    & 256  \\
      TTUR                       & -    & 5    \\
      GAN Head Layers            & -    & 5, 15, 25, 35, 39 \\
      GAN Head Dim               & -    & 768 \\
      GAN Start Step             & -    & 1000 \\
      EMA Decay                  & 0.99 & 0.99 \\
      EMA Start Step             & 250  & 750 \\
      Training Steps             & 3759 & 2250 \\
      Numerical Precision        & BFloat16 & BFloat16 \\
      GPU Usage                  & 128 NVIDIA H100 & 128 NVIDIA H100 \\
      History Corrupt Ratio      & \multicolumn{2}{c}{$p_a=0.4, p_b=0.4, p_c=0.0, p_d=0.2; a_{\min}=0.3, a_{\max}=1.7, b_{\min}=0, b_{\max}=0.33$} \\
      \bottomrule
    \end{tabular}
  }
\end{table}

\noindent\myparagraph{Flash Normalization.}
We implement kernel fusion for LayerNorm and RMSNorm using Triton, following \cite{unsloth, hsu2025ligerkernel}. By consolidating mean and variance computation, normalization, and affine transformations into a single kernel, we minimize memory traffic and leverage optimized primitives such as tl.math.rsqrt. To reduce memory footprint, we cache only scalar statistics (row-wise $\text{inv\_var} \in \mathbb{R}^{B \times \ell}$ and $\mu \in \mathbb{R}^{B \times \ell}$) for the backward pass, avoiding storage of the full normalized tensor $\mathbf{z} \in \mathbb{R}^{B \times \ell \times D}$. This approach reduces the memory complexity of intermediate activations from $\mathcal{O}(B\ell D)$ to $\mathcal{O}(B\ell)$. Furthermore, we adopt a mixed-precision strategy where internal computations are performed in FP32 for numerical stability, while inputs and outputs retain their original data types (\eg, bfloat16). Finally, we maximize GPU bandwidth utilization through row-wise parallelism---mapping one program instance per token---and coalesced memory access patterns.

% 我们通过triton对LayerNorm和RMSNorm进行kernel融合优化：通过将均值计算、方差计算、归一化和仿射变换融合到单个GPU kernel中，减少数据在GPU和内存间的传输次数，并利用tl.math.rsqrt等优化的数学运算提升计算效率。显存优化方面，反向传播时仅保存标量统计量（每行的$\text{inv_var} \in \mathbb{R}^{B \times \ell}$和$\mu \in \mathbb{R}^{B \times \ell}$），而非完整的归一化张量$\mathbf{z} \in \mathbb{R}^{B \times \ell \times D}$，将中间激活显存从$O(B\ellD)$降至$O(B\ell)$。前向传播中采用混合精度策略，计算过程使用float32保证数值稳定性，输入输出保持原始dtype（如bfloat16），进一步减少显存开销。此外，通过按行并行处理（每个token对应一个program）和内存合并访问模式，最大化GPU带宽利用率。

\noindent\myparagraph{Flash Rotary Position Embedding.}
We implement kernel fusion optimization for Rotary Positional Embeddings (RoPE) using Triton. By consolidating complex number decomposition, rotation matrix multiplication, and result reconstruction into a single GPU kernel, we eliminate the memory fragmentation and data copying overheads inherent in PyTorch’s native unflattening, chunking, and indexing operations. Specifically, we flatten the input $\mathbf{x} \in \mathbb{R}^{B \times \ell \times H \times D}$ to $\mathbb{R}^{(B \cdot \ell \cdot H) \times D}$. We parallelize execution by mapping one program instance per attention head, using interleaved memory access to directly retrieve real and imaginary components. The rotation is applied using pre-computed $\cos$ and $\sin$ values—$\text{out}_{\text{real}} = x_{\text{real}} \cdot \cos - x_{\text{imag}} \cdot \sin$ and $\text{out}_{\text{imag}} = x_{\text{real}} \cdot \sin + x_{\text{imag}} \cdot \cos$—with results written back in-place. For backward propagation, we reuse the forward kernel to perform inverse rotation by simply negating the sine component ($\sin_{\text{neg}} = -\sin$). This strategy obviates the need to store full intermediate tensors, requiring only $\cos \in \mathbb{R}^{B \times \ell \times (D/2)}$ and $\sin \in \mathbb{R}^{B \times \ell \times (D/2)}$. Consequently, we reduce the memory complexity of intermediate activations from $\mathcal{O}(B\ell HD)$ to $\mathcal{O}(B\ell D)$, where $B$, $\ell$, $H$, and $D$ denote batch size, sequence length, head count, and head dimension, respectively.

% 我们通过triton对RoPE进行kernel融合优化：将RoPE的复数分解、旋转矩阵乘法和结果重组融合到单个GPU kernel中，避免了PyTorch原生实现中多次张量unflatten、chunk和索引操作导致的内存碎片和额外拷贝开销。具体而言，输入$\mathbf{x} \in \mathbb{R}^{B \times \ell \times H \times D}$被展平为$\mathbb{R}^{(B \cdot \ell \cdot H) \times D}$，每个attention head对应一个program并行处理，通过交错访问模式直接读取实部和虚部，利用预计算的$\cos$和$\sin$执行旋转变换：$\text{out}{\text{real}} = x{\text{real}} \cdot \cos - x_{\text{imag}} \cdot \sin$，$\text{out}{\text{imag}} = x{\text{real}} \cdot \sin + x_{\text{imag}} \cdot \cos$，并原地写回结果。反向传播时，通过取负sin（$\sin_{\text{neg}} = -\sin$）复用前向kernel实现逆旋转，无需保存完整的归一化张量，仅保存$\cos \in \mathbb{R}^{B \times \ell \times (D/2)}$和$\sin \in \mathbb{R}^{B \times \ell \times (D/2)}$，将中间激活显存从$O(BLHD)$降至$O(B\ellD)$。其中$B$为batch size，$\ell$为序列长度，$H$为attention head数，$D$为head维度。

\subsection{Other Techniques.}
By eliminating the need for causal masking, \textit{\ours} seamlessly integrates with high-efficiency attention backends such as FlashAttention \cite{flashattention2, flashattention}, thereby achieving superior throughput and reduced latency.

% 得益于无需引入causal mask，\textit{\ours} 可以无缝结合 FlashAttention 等高效 attention backend，实现更高的 计算吞吐与更低的推理延迟。

\section{Experiments}\label{sec:experiment}

\subsection{Implementation Details}\label{sec:implementation}

\begin{table}[!t]
  \centering
  \caption{\textbf{Quantitative comparisons on 81-frames short videos.} \ours achieves a superior speed-quality trade-off over existing methods, which are either computationally prohibitive or yield suboptimal results. ``$\uparrow$'' higher is better.
    % Best results in \textbf{bold}, second-best \uline{underlined}.
  }
  \label{tab:performance_short}
  \resizebox{\textwidth}{!}{
    \begin{tabular}{c|cc|c|ccccc}
      \toprule
      \textbf{Model} & \textbf{\#Params} & \makecell{\textbf{Throughput}\\\textbf{(FPS) $\uparrow$}} & \makecell{\textbf{Total}\\\textbf{$\uparrow$}} & \makecell{\textbf{Aesthetic}\\\textbf{$\uparrow$}} & \makecell{\textbf{Dynamic}\\\textbf{$\uparrow$}} & \makecell{\textbf{Smoothness}\\\textbf{$\uparrow$}} & \makecell{\textbf{Semantic}\\\textbf{$\uparrow$}} & \makecell{\textbf{Naturalness}\\\textbf{$\uparrow$}} \\
      \midrule
      \midrule
      \rowcolor{myblue!30}
      \multicolumn{9}{l}{\textit{Bidirectional Models}}\\
      SANA Video \cite{sana_video}                    & 2B     & 1.34  & 4.60 & 9 & 7 & 9 & 5 & 1  \\
      CogVideoX \cite{cogvideox}                      & 2B     & 1.95  & 5.55 & 7 & 10 & 7 & 4 & 5  \\
      CogVideoX 1.5 \cite{cogvideox}                  & 5B     & 1.47  & 5.15 & 7 & 5 & 8 & 5 & 4  \\
      Mochi-1 \cite{mochi}                            & 10B    & 0.53  & 5.35 & 6 & 7 & 9 & 5 & 4  \\
      HV Video \cite{hunyuanvideo}                    & 13B    & 0.36  & 6.00 & 8 & 8 & 9 & 5 & 5  \\
      HV Video 1.5 \cite{hunyuanvideo15}              & 8.3B   & 0.24  & 6.90 & 8 & 10 & 9 & 5 & 7  \\
      Wan 2.1 1.3B \cite{wan}                         & 1.3B   & 1.60  & 6.10 & 8 & 10 & 8 & 6 & 4  \\
      Wan 2.2 5B \cite{wan}                           & 5B     & 3.32  & 6.05 & 7 & 7 & 8 & 5 & 6  \\
      Wan 2.1 14B \cite{wan}                          & 14B    & 0.33  & 6.15 & 8 & 6 & 9 & 6 & 5  \\
      Wan 2.2 14B \cite{wan}                          & 14B    & 0.33  & 6.35 & 8 & 8 & 9 & 5 & 6  \\
      LTX Video \cite{ltx}                            & 1.9B     & 15.03 & 4.45 & 6 & 4 & 10 & 1 & 6  \\
      LTX Video 2 \cite{ltx2}                        & 19B    & 3.84  & 6.00 & 8 & 8 & 9 & 6 & 4  \\
      Kandinsky 5 lite \cite{kandinsky5}              & 2B     & 1.03  & 6.25 & 8 & 6 & 10 & 5 & 6  \\
      Kandinsky 5 pro \cite{kandinsky5}               & 19B    & 0.49  & 7.35 & 8 & 10 & 10 & 6 & 7  \\
      StepVideo T2V \cite{stepvideo}                  & 30B    & 0.27  & 4.00 & 6 & 4 & 9 & 1 & 5  \\
      \midrule
      FastVideoWan 2.1 \cite{fastvideo}               & 14B    & 5.37  & 5.25 & 9 & 3 & 9 & 5 & 4  \\
      TurboDiffusion 2.1 \cite{turbodiffusion}        & 14B    & 10.15 & 5.10 & 8 & 6 & 9 & 4 & 4  \\
      TurboDiffusion 2.1-Quant \cite{turbodiffusion}  & 14B   & 8.48  & 5.45 & 8 & 6 & 9 & 4 & 5  \\
      \midrule
      \midrule
      \rowcolor{myblue!30}
      \multicolumn{9}{l}{\textit{Autoregressive Models}}\\
      NOVA \cite{nova}                           & 0.6B  & 1.18  & 3.70 & 5 & 2 & 9 & 4 & 2  \\
      Pyramid Flow \cite{pyramidflow}            & 2B    & 3.11  & 4.55 & 8 & 3 & 10 & 4 & 3  \\
      MAGI-1 \cite{magi}                         & 4.5B  & 0.37  & 5.25 & 8 & 3 & 10 & 6 & 3  \\
      InfinityStar \cite{infinitystar}           & 8B    & 3.38  & 5.30 & 8 & 8 & 9 & 6 & 2  \\
      SkyReelsV2-DF \cite{skyreelsv2}            & 1.3B  & 0.55  & 5.65 & 8 & 8 & 9 & 5 & 4  \\
      SkyReelsV2-DF \cite{skyreelsv2}            & 14B   & 0.12  & 5.55 & 8 & 7 & 9 & 5 & 4  \\
      \midrule
      CausVid \cite{causvid}                     & 1.3B  & 24.41 & 4.50 & 8 & 7 & 9 & 4 & 2  \\
      Self Forcing \cite{selfforcing}            & 1.3B  & 21.20 & 5.75 & 8 & 9 & 9 & 5 & 4  \\
      Rolling Forcing \cite{rollingforcing}      & 1.3B  & 19.47 & 5.25 & 8 & 4 & 9 & 5 & 4  \\
      LongLive \cite{longlive}                   & 1.3B  & 18.05 & 5.80 & 8 & 5 & 10 & 5 & 5  \\
      Infinite Forcing \cite{infinite_forcing}   & 1.3B  & 22.19 & 5.10 & 8 & 6 & 9 & 4 & 4  \\
      Reward Forcing \cite{rewardforcing}        & 1.3B  & 22.13 & 5.55 & 8 & 7 & 9 & 5 & 4  \\
      Causal Forcing \cite{causalforcing}        & 1.3B  & 20.98 & 5.40 & 8 & 10 & 8 & 5 & 3 \\
      Dummy Forcing Long \cite{dummyforcing}     & 1.3B  & 20.10 & 5.45 & 9 & 4 & 10 & 5 & 4  \\
      SANA Video Long \cite{sana_video}          & 2B    & 13.24 & 3.85 & 9 & 2 & 10 & 4 & 1  \\
      Krea \cite{krea_realtime_14b}              & 14B   & 6.74  & 5.95 & 9 & 10 & 9 & 5 & 4  \\
      \midrule
      \midrule
      \rowcolor{myblue!30}
      \multicolumn{9}{l}{\textit{Autoregressive Video Continuation Models}}\\
      LongCat-Video \cite{longcat_video}        & 13.6B  & 0.33  & 6.30 & 9 & 10 & 9 & 6 & 4  \\
      \textbf{Helios-Base}                    & 14B    & 0.54  & 6.35 & 8 & 8  & 9 & 5 & 6  \\
      \textbf{Helios-Mid}                    & 14B    & 1.05  & 6.25 & 8 & 7  & 9 & 5 & 6  \\
      \textbf{Helios-Distilled}                    & 14B    & 19.53 & 6.00 & 8 & 7  & 10 & 5 & 5  \\
      \bottomrule
    \end{tabular}
  }
\end{table}

\begin{figure*}[!t]
  \centering
  \includegraphics[width=1\linewidth]{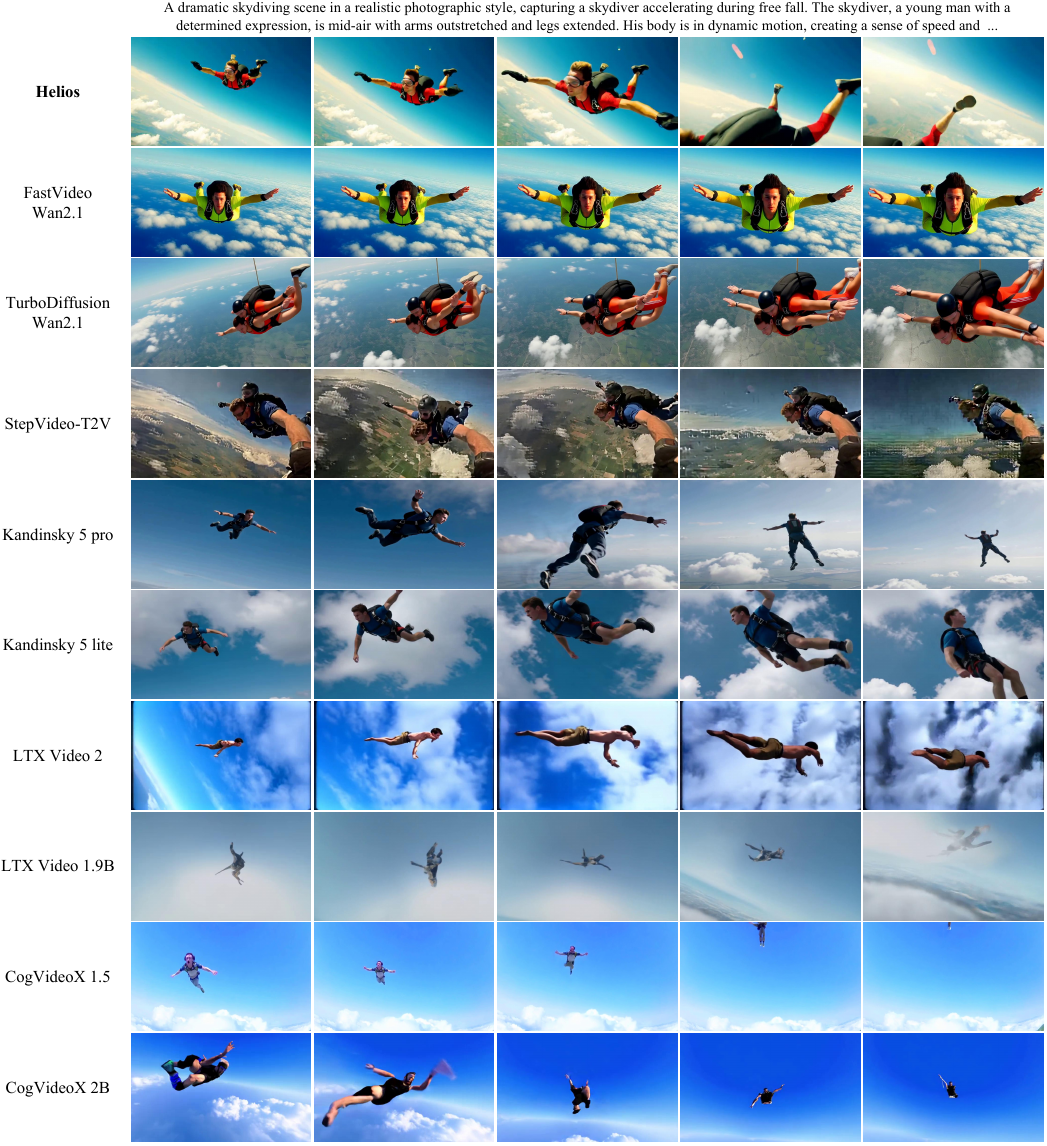}
  \caption{\textbf{Qualitative comparisons on 81-frames short videos (Part-1).} Even as a distilled model, \textit{Helios} matches or even surpasses the base models in terms of visual quality, motion dynamics, and naturalness.}
  \label{figure_qualitative_comparison_short-1}
\end{figure*}
\begin{figure*}[!t]
  \centering
  \includegraphics[width=1\linewidth]{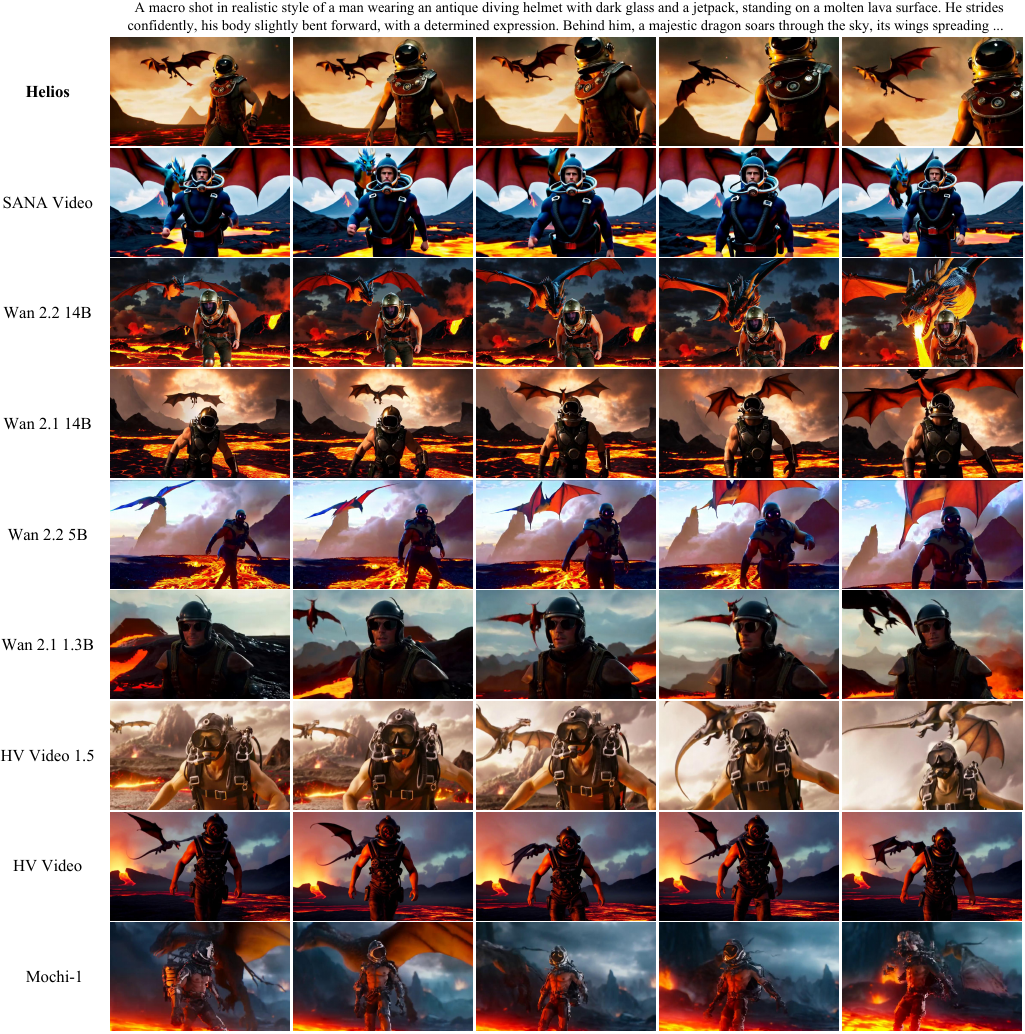}
  \caption{\textbf{Qualitative comparisons on 81-frames short videos (Part-2).} Despite being a distilled model, \textit{Helios} matches or surpasses the base models in visual fidelity, text alignment, and overall realism.}
  \label{figure_qualitative_comparison_short-2}
\end{figure*}

\noindent\myparagraph{Training.}
We initialize from Wan-2.1-T2V-14B \cite{wan} and train on 0.8M clips of duration $<10$ seconds using a three-stage progressive pipeline. \textbf{Stage~1 (Base)} performs architectural adaptation: we apply Unified History Injection, Easy Anti-Drifting, and Multi-Term Memory Patchification to convert the bidirectional pretrained model into an autoregressive generator. \textbf{Stage~2 (Mid)} targets token compression by introducing Pyramid Unified Predictor Corrector, which aggressively reduces the number of noisy tokens and thus the overall computation. \textbf{Stage~3 (Distilled)} applies Adversarial Hierarchical Distillation, reducing the sampling steps from 50 to 3 and eliminating the need for classifier-free guidance (CFG). Throughout training, we apply dynamic shifting to all timestep-dependent operations to match the noise schedule to the latent size. We cap the resolution at $384 \times 640$ and extract 109-frame clips from each video. More details are in Tables \ref{tab:training_hyperparameters-1-2} and \ref{tab:training_hyperparameters-3}.

% 我们选择Wan-2.1-T2V-14B作为baseline，基于OpenSoraPlan数据（视频片段均在10秒以内）构建了一个三阶段渐进式训练pipeline。第一阶段聚焦于模型架构转换，使用了Unified History Injection、Easy Anti-Drifting和Multi-Term Memory Patchification，将预训练模型改造为自回归生成模式；第二阶段致力于token压缩优化，结合Pyramid Unified Predictor-Corrector对noisy token进行深度压缩以提升效率；第三阶段则通过Adversarial Hierarchical Distillation将推理步数从50步大幅降至3步，并完全移除了对CFG的依赖。在数据处理层面，我们统一采用384x640的最大分辨率标准，从每个视频中提取连续的109帧序列进行训练。在训练过程中，我们对所有涉及timestep的操作（包括蒸馏阶段）进行了dynamic shifting，使得噪声调度和latent尺寸相适应。

\noindent\myparagraph{Inference.}
For Stages~1--2, we adopt UniPC~\cite{unipc} scheduler with 50 sampling steps, a classifier-free guidance (CFG) scale of 5.0, and $v$-prediction. For Stage~2, instead of standard CFG, we employ CFG-Zero-Star \cite{cfgzerostar}. For Stage~3, we use $x_0$-prediction with 3 sampling steps and a CFG scale of 1.0.

% 对于第一和第二阶段的推理，我们选择UniPC作为采样器，采样步数为50，CFG scale设置为5.0进行v-prediction采样。对于第三阶段的推理，由于是x0-prediction，因此无需采样器，采样步数设置为3，CFG scale设置为1.0。

\noindent\myparagraph{HeliosBench.}
Because no open-source benchmark targets real-time long-video generation, we build \emph{HeliosBench}, a test set of 240 LLM-refined prompts from Self-Forcing \cite{selfforcing}. We evaluate four duration tiers: very short (81 frames), short (240 frames), medium (720 frames), and long (1440 frames). For automated evaluation, existing benchmarks are only weakly aligned with human preference \cite{vbench, vbench2, vbench++, chronomagicbench}; however, they remain the best available options. Following \cite{vbench, chronomagicbench, opens2v}, we report five dimensions: (1) \textbf{Aesthetic}, measured by the LAION aesthetic predictor \cite{aesthecit_predictor}; (2) \textbf{Dynamic}, computed using the Farneb\"ack algorithm \cite{opens2v}; (3) \textbf{Motion Smoothness}, measured by RAFT \cite{raft}; (4) \textbf{Semantic}, measured by ViCLIP \cite{viclip} for video--text alignment; and (5) \textbf{Naturalness}, measured by OpenS2V-Eval \cite{opens2v}. We additionally follow \cite{framepack} to quantify drifting on Aesthetic, Motion Smoothness, Semantic, and Naturalness. Since these metrics are noisy and their raw scores may correlate poorly with human perception, we map each metric to a 10-point scale using its empirical score distribution for improved robustness. To measure throughput (FPS), we report end-to-end speed at $384 \times 640$ under default frame lengths, including the latency of both the VAE and the text encoder. For each model, we enable its officially supported acceleration techniques (e.g., FlashAttention, \texttt{torch.compile}, KV-cache, and warm-up) to achieve the best possible throughput. More details are provided in the Appendix \ref{sec:calculation_of_bench}.

For baselines, we compare \textit{\ours} with a broad set of open-source video generation models, including (1) \textbf{base models}: SANA Video \cite{sana_video}, CogVideoX \cite{cogvideox}, Mochi \cite{mochi}, HV Video \cite{hunyuanvideo, hunyuanvideo15}, Wan \cite{wan}, LTX Video \cite{ltx, ltx2}, Kandinsky \cite{kandinsky5}, StepVideo \cite{stepvideo}, NOVA \cite{nova}, Pyramid Flow \cite{pyramidflow}, MAGI \cite{magi}, InfinityStar \cite{infinitystar}, SkyReelsV2 \cite{skyreelsv2}, and LongCat-Video \cite{longcat_video}; and (2) \textbf{distilled models}: FastVideo \cite{fastvideo}, TurboDiffusion \cite{turbodiffusion}, CausVid \cite{causvid}, Self-Forcing \cite{selfforcing}, Rolling Forcing \cite{rollingforcing}, LongLive \cite{longlive}, Infinite Forcing \cite{infinite_forcing}, Reward Forcing \cite{rewardforcing}, Causal Forcing \cite{causalforcing}, Dummy Forcing \cite{dummyforcing}, SANA Video Long \cite{sana_video}, and Krea \cite{krea_realtime_14b}. For models that only support short-video generation, we construct a subset by matching the 240 prompts to each model's default length. For a fair comparison, we truncate outputs from long-video models to the first 81 frames.

% 由于现有的benchmark与人类感知对齐度较低，且缺少开源的长视频benchmark，我们提出了新的评估方法。对于测试用例，我们首先从Self-Forcing收集了240个LLM-refined的prompt，并将视频划分为四个时长范围：极短视频（81帧）、短视频（240帧）、中视频（720帧）和长视频（1440帧）。在自动化指标方面，我们遵循XXX的方法，选择了六个维度来评估视频的基础素质：(1). Throughput (FPS)：评估模型在384x640分辨率及官方默认视频长度下生成视频的端到端速度（例如，只记录KV-Cache填满后的速度），包括VAE和Text Encoder的处理速度; (2). Aesthetic：通过LAION aesthetic predictor衡量视频的美学价值; (3). Dynamic：通过插帧模型测量视频中的运动平滑度; (4). Motion Smoothness：利用RAFT（Recurrent All-Pairs Field Transforms）评估视频中的运动平滑性; (5). Semantic：通过ViCLIP衡量生成视频与prompt的相关性; (6). Naturalness：通过OpenS2V-Eval评估生成内容是否与符合物理规律。此外，我们采用了XXX的drifting measurement方法，基于Aesthetic、Motion Smoothness、Semantic和Natural这四个维度来评估视频的drifting程度。然而，由于Aesthetic、Dynamic和Motion Smoothness的稳定性较低，某些优质视频反而得分较低，而一些较差的视频可能得分较高。为了解决这个问题，我们根据各维度得分的特性，将原始分数转化为10分制的评分系统，以提高评分的可靠性。具体细节可以参见附录。
% 其中一些只能生成短视频，因此为了对这些模型进行测试，我们构建了一个子集，及将先前构建的240个prompt对应的视频长度都设置为模型的默认视频长度，并且将长视频模型生成的视频截取前81帧用于和这些模型进行公平评测。

\subsection{Qualitative and Quantitative Comparison}

\noindent\myparagraph{Short Video Generation.}
First, we benchmark the ability of various models to generate 81-frame ultra-short video clips using Aesthetic, Dynamic, Smoothness, Semantic, and Naturalness scores, as well as their weighted sum. As shown in Table \ref{tab:performance_short}, our method achieves an overall score of 6.00, surpassing all distilled models and matching the performance of most base models of the same size. Notably, distilled models tend to produce videos with higher saturation and smaller motion amplitudes, which leads to higher Aesthetic and Smoothness scores compared to base models; however, this does not necessarily correlate with superior quality. Consequently, we focus primarily on Semantic and Naturalness. The experimental results indicate that \textit{Helios} excels in these aspects, either matching or surpassing models such as Wan 14B \cite{wan} and HV Video \cite{hunyuanvideo, hunyuanvideo15}, thereby demonstrating strong video generation quality. In addition, \textit{Helios} achieves a better balance between Dynamic and Motion Smoothness: it avoids the overly static motion patterns often seen in distilled models, while not introducing the temporal jitter or local inconsistency that can appear in aggressive acceleration settings. Furthermore, \textit{Helios} achieves a real-time generation speed of 19.53 FPS on a single H100 GPU. In contrast, while SANA Video Long \cite{sana_video} is a distilled model and seven times smaller than \textit{Helios}, its generation speed is 1.28 times slower. Compared with the same-sized FastVideo \cite{fastvideo} and TurboDiffusion \cite{turbodiffusion}, \textit{Helios} is 2--3$\times$ faster and outperforms Wan 14B \cite{wan} by a factor of 52. These results show that \textit{Helios} not only advances video generation quality but also leads in generation speed among existing large-scale video generation models. A comprehensive qualitative analysis further supports the effectiveness of our method. Specifically, Figure \ref{figure_qualitative_comparison_short-1} and Figure \ref{figure_qualitative_comparison_short-2} illustrate that \textit{\ours} generates high-quality videos that are more natural and better aligned with human perception than those from distilled models, while remaining comparable to base models.

% This observation also highlights the limited robustness of commonly used metrics \cite{vbench, chronomagicbench}.

% 首先，我们使用Aesthetic、Dynamic、Smoothness、Semantic、Naturalness以及这些分数的加权求和，来benchmark各个模型生成81帧超短视频片段的能力。如表\ref{tab:performance_short}所示，我们的方法取得了XXX的总体视频质量得分，超越了所有蒸馏模型，并且达到了大部分同尺寸基础模型的水平。需要注意的是，蒸馏模型通常偏向于生成高饱和度和较小运动幅度的视频，这使得它们在Aesthetic和Smoothness得分上普遍高于基础模型，但这并不意味着更好。因此，我们的研究主要集中在Semantic和Naturalness方面。从实验结果来看，Helios在这些维度上表现优异，基本达到或超过了如Wan 14B和HV Video等模型的水平，展现出卓越的视频生成质量。此外，Helios在单个H100 GPU上的实时生成速度为19.53 FPS，令人瞩目。相比之下，尽管SANA Video Long是一个蒸馏模型，且尺寸小于Helios 7倍，但其生成速度仅为Helios的1.28倍。而与同尺寸的FastVideo和TurboDiffusion相比，Helios的生成速度快了2到3倍，甚至比Wan 14B快了52倍。由此可见，Helios不仅在视频生成质量上取得了显著突破，而且在生成速度方面也领先于现有的超大规模视频生成模型，成为目前最快的模型之一。对结果的定性分析证实了我们方法的有效性。具体而言，图XXX展示了\textit{\ours}能够和base models一样生成比蒸馏模型更加自然、更符合人类感知的高质量视频，同时验证了现有常用指标 \cite{vbench, chronomagicbench}的鲁棒性较差的问题。

\begin{table}[!t]
  \centering
  \caption{\textbf{Quantitative comparisons on 120, 240, 720 and 1440-frames long videos.} \ours consistently exceeds existing real-time long video generation methods. ``$\uparrow$'' denotes higher is better.
    % Best results in \textbf{bold}, second-best \uline{underlined}.
  }
  \label{tab:performance_long}
  \resizebox{\textwidth}{!}{
    \begin{tabular}{c|cc|cc|c|ccccc|cccc}
      \toprule
      \textbf{Model} & \textbf{\#Params} &
      \makecell{\textbf{Throughput}\\\textbf{(FPS) $\uparrow$}} & \makecell{\textbf{Total}\\\textbf{$\uparrow$}} &
      \makecell{\textbf{Throughput}\\\textbf{Score $\uparrow$}} &
      \makecell{\textbf{Total*}\\\textbf{$\uparrow$}} &
      \makecell{\textbf{Aesthetic}\\\textbf{$\uparrow$}} & \makecell{\textbf{Dynamic}\\\textbf{$\uparrow$}} & \makecell{\textbf{Smoothness}\\\textbf{$\uparrow$}} & \makecell{\textbf{Semantic}\\\textbf{$\uparrow$}} & \makecell{\textbf{Naturalness}\\\textbf{$\uparrow$}} &
      \makecell{\textbf{Drifting}\\\textbf{Aesthetic $\uparrow$}} & \makecell{\textbf{Drifting}\\\textbf{Smoothness $\uparrow$}} & \makecell{\textbf{Drifting}\\\textbf{Semantic $\uparrow$}} & \makecell{\textbf{Drifting}\\\textbf{Naturalness $\uparrow$}} \\
      \midrule
      \midrule
      \rowcolor{myblue!30}
      \multicolumn{14}{l}{\textit{Autoregressive Models}}\\
      NOVA \cite{nova}                         & 0.6B  & 1.18  & 2.48 & 1 & 2.38 & 1 & 1 & 9 & 1 & 2 & 1 & 9 & 1 & 2 \\
      Pyramid Flow \cite{pyramidflow}          & 2B    & 3.11  & 2.85 & 1 & 2.75 & 6 & 3 & 9 & 3 & 1 & 1 & 8 & 1 & 2 \\
      MAGI-1 \cite{magi}                       & 4.5B  & 0.37  & 4.92 & 1 & 4.82 & 7 & 3 & 10 & 6 & 2 & 1 & 10 & 6 & 5 \\
      InfinityStar \cite{infinitystar}         & 8B    & 3.38  & 2.63 & 2 & 2.43 & 4 & 4 & 8 & 2 & 1 & 1 & 9 & 1 & 1 \\
      SkyReelsV2-DF \cite{skyreelsv2}          & 1.3B  & 0.55  & 3.29 & 1 & 3.19 & 6 & 6 & 9 & 4 & 1 & 1 & 9 & 2 & 1 \\
      SkyReelsV2-DF \cite{skyreelsv2}          & 14B   & 0.12  & 3.87 & 1 & 3.77 & 6 & 6 & 8 & 5 & 2 & 1 & 8 & 4 & 1 \\
      \midrule
      CausVid \cite{causvid}                   & 1.3B  & 24.41 & 5.38 & 8 & 4.58 & 8 & 6 & 9 & 3 & 1 & 6 & 10 & 7 & 6 \\
      Self Forcing \cite{selfforcing}          & 1.3B  & 21.20 & 5.00 & 7 & 4.30 & 7 & 5 & 9 & 5 & 2 & 1 & 10 & 5 & 3 \\
      Rolling Forcing \cite{rollingforcing}    & 1.3B  & 19.47 & 6.86 & 7 & 6.16 & 8 & 3 & 9 & 5 & 4 & 7 & 10 & 9 & 7 \\
      LongLive \cite{longlive}                 & 1.3B  & 18.05 & 6.82 & 6 & 6.22 & 8 & 5 & 9 & 5 & 4 & 8 & 10 & 9 & 6 \\
      Infinite Forcing \cite{infinite_forcing} & 1.3B  & 22.19 & 6.50 & 7 & 5.80 & 8 & 6 & 9 & 4 & 4 & 7 & 10 & 9 & 5 \\
      Reward Forcing \cite{rewardforcing}      & 1.3B  & 22.13 & 6.88 & 7 & 6.18 & 8 & 7 & 9 & 5 & 4 & 7 & 10 & 9 & 6 \\
      Causal Forcing \cite{causalforcing}      & 1.3B  & 20.98 & 3.86 & 7 & 3.61 & 7 & 10 & 8 & 4 & 1 & 1 & 9 & 3 & 3 \\
      Dummy Forcing Long \cite{dummyforcing}   & 1.3B  & 20.10 & 6.14 & 7 & 5.44 & 8 & 4 & 9 & 5 & 3 & 7 & 10 & 8 & 3 \\
      SANA Video Long \cite{sana_video}        & 2B    & 13.24 & 6.03 & 5 & 5.53 & 9 & 2 & 10 & 5 & 1 & 7 & 10 & 9 & 8 \\
      Krea \cite{krea_realtime_14b}            & 14B   & 6.74  & 4.10 & 3 & 3.80 & 7 & 10 & 9 & 5 & 1 & 1 & 10 & 3 & 1 \\
      \midrule
      \rowcolor{myblue!30}
      \multicolumn{14}{l}{\textit{Autoregressive Video Continuation Models}}\\
      LongCat-Video \cite{longcat_video}       & 13.6B  & 0.33  & 6.54 & 1 & 6.44 & 8 & 7 & 9  & 6 & 4 & 7 & 10 & 8 & 7 \\
      \textbf{Helios-Base}                     & 14B    & 0.54  & 6.57 & 1 & 6.47 & 8 & 6 & 9  & 6 & 5 & 7 & 10 & 8 & 5 \\
      \textbf{Helios-Mid}                      & 14B    & 1.05  & 6.05 & 1 & 5.95 & 7 & 5 & 9  & 5 & 5 & 5 & 10 & 7 & 6 \\
      \textbf{Helios-Distilled}                & 14B    & 19.53 & 6.94 & 6 & 6.34 & 8 & 6 & 10 & 5 & 5 & 7 & 10 & 7 & 7 \\
      \bottomrule
    \end{tabular}
  }
\end{table}

\begin{figure*}[!t]
  \centering
  \includegraphics[width=1\linewidth]{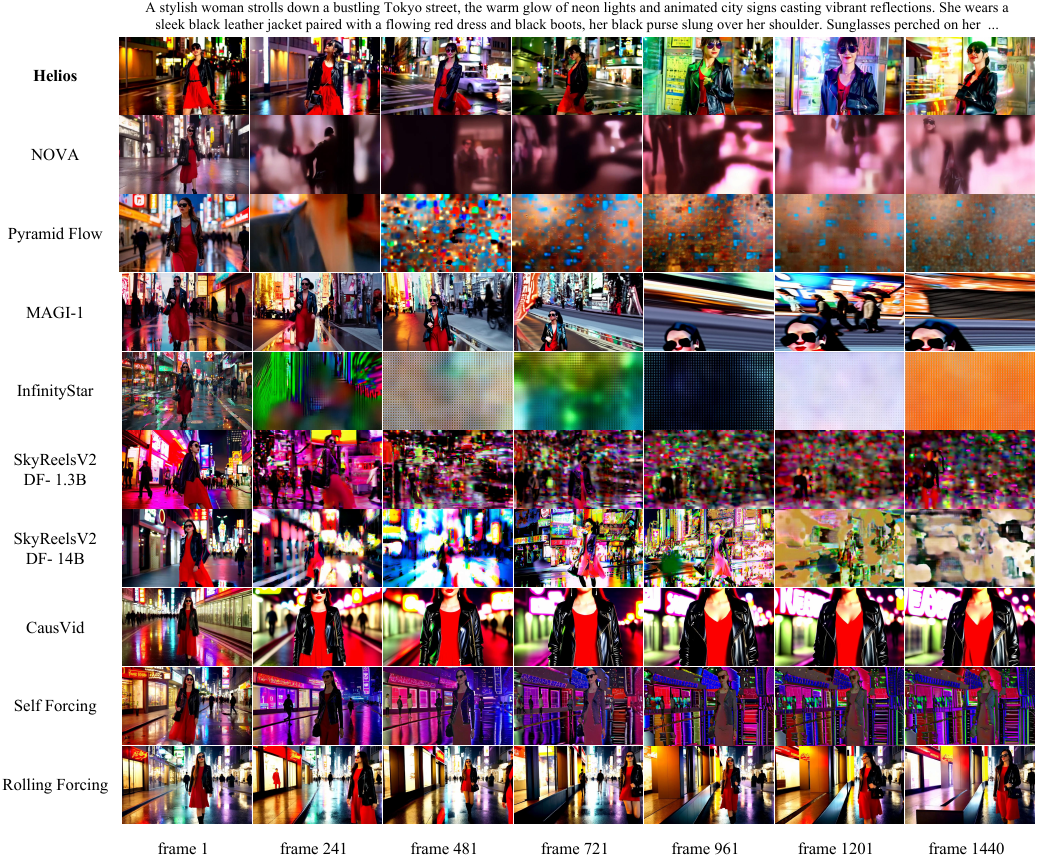}
  \caption{\textbf{Qualitative comparisons on 120, 240, 720 and 1440-frames long videos (Part-1).} It is clear that \ours consistently outperforms the baseline models in terms of realism and naturalness.}
  \label{figure_qualitative_comparison_long-1}
\end{figure*}
\begin{figure*}[!t]
  \centering
  \includegraphics[width=1\linewidth]{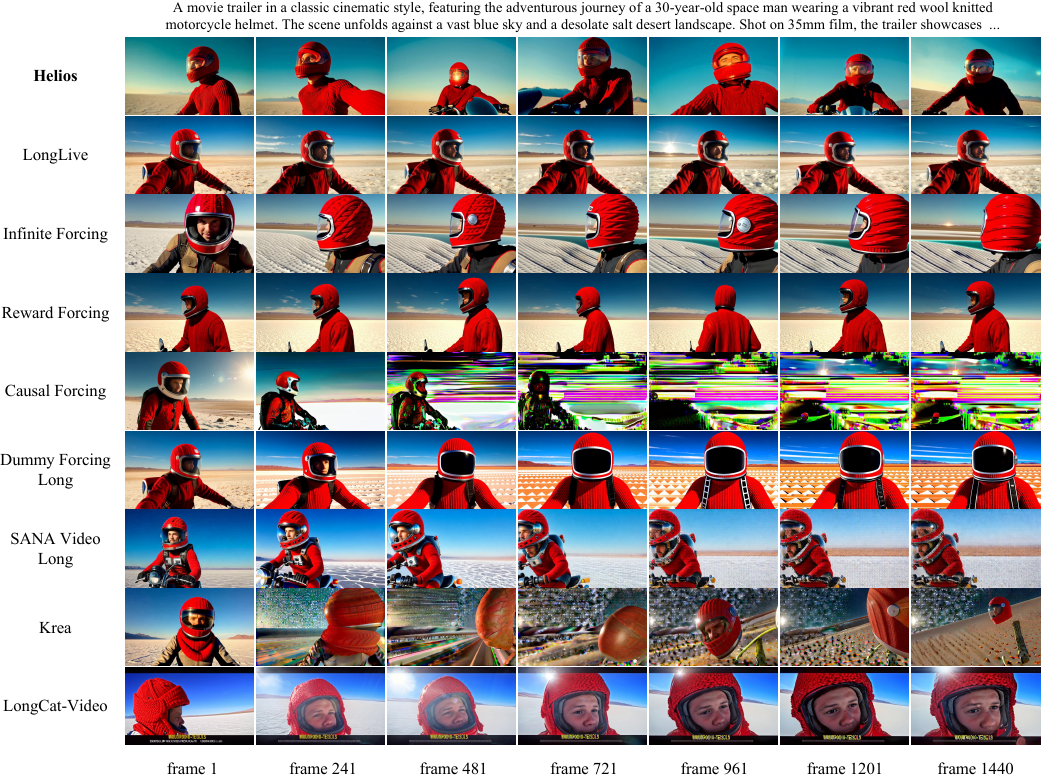}
  \caption{\textbf{Qualitative comparisons on 120, 240, 720 and 1440-frames long videos (Part-2).} It is clear that \ours consistently outperforms the baseline models in terms of text alignment and dynamics.}
  \label{figure_qualitative_comparison_long-2}
\end{figure*}

\noindent\myparagraph{Long Video Generation.}
Next, building on Table~\ref{tab:performance_short}, we introduce \emph{Throughput Score} and \emph{Drifting Score} to evaluate long-video generation across different durations. As shown in Table~\ref{tab:performance_long}, our method achieves a total score of 7.08, outperforming the strongest baseline, Reward Forcing \cite{rewardforcing} (6.88), while maintaining competitive runtime. In particular, we obtain a higher Naturalness score (6), avoiding the over-saturated appearance commonly observed in distilled models, and we simultaneously improve Dynamic and Motion Smoothness, yielding more vivid yet physically plausible motion. We also observe that \textit{\ours} exhibits consistently lower drifting across multiple dimensions (Aesthetic, Semantic, and Naturalness), indicating that the model better preserves content identity and scene layout as the video extends to hundreds or even thousands of frames. Moreover, although \textit{\ours}-Stage1 and \textit{\ours}-Stage2 do not rely on Self-Forcing \cite{selfforcing} or Error-Banks \cite{SVI}, they still effectively mitigate drifting over long horizons, suggesting a complementary route to improve long-term consistency. Figures~\ref{figure_qualitative_comparison_long-1} and \ref{figure_qualitative_comparison_long-2} corroborate these quantitative results: \textit{\ours} preserves visual quality over time, whereas baseline methods exhibit noticeable degradation and inconsistencies.

% 接下来，在表\ref{tab:performance_short}的基础上，我们进一步引入了Throughput Score和Drifting Score，用于基准评测各个模型生成不同长度视频片段的能力。如表\ref{tab:performance_long}所示，我们的方法在总分上达到了XXX，显著优于当前最先进的基线方法Reward Forcing（6.88）。特别值得注意的是，我们的方法在Naturalness（6）的表现上明显优于现有方法，避免了distilled models常见的生成高饱和度视频的问题，且能够生成更具动态变化且符合物理规律的视频。此外，我们还观察到，尽管Helios-Stage1和Helios-Stage2未采用Self-Forcing和Error-Bank策略，我们的模型依然能够在长时间段的视频生成中有效避免drifting的现象，从而为社区提供了一种新的解决drifting问题的思路。对结果的定性分析进一步证实了我们方法的有效性。具体而言，图XXX展示了\textit{\ours}在长期生成过程中保持了优异的一致性，而与之对比的基线方法则随着时间的推移出现了明显的质量下降和不一致。

\subsection{User Study}
\begin{figure*}[!t]
  \centering
  \includegraphics[width=0.49\linewidth]{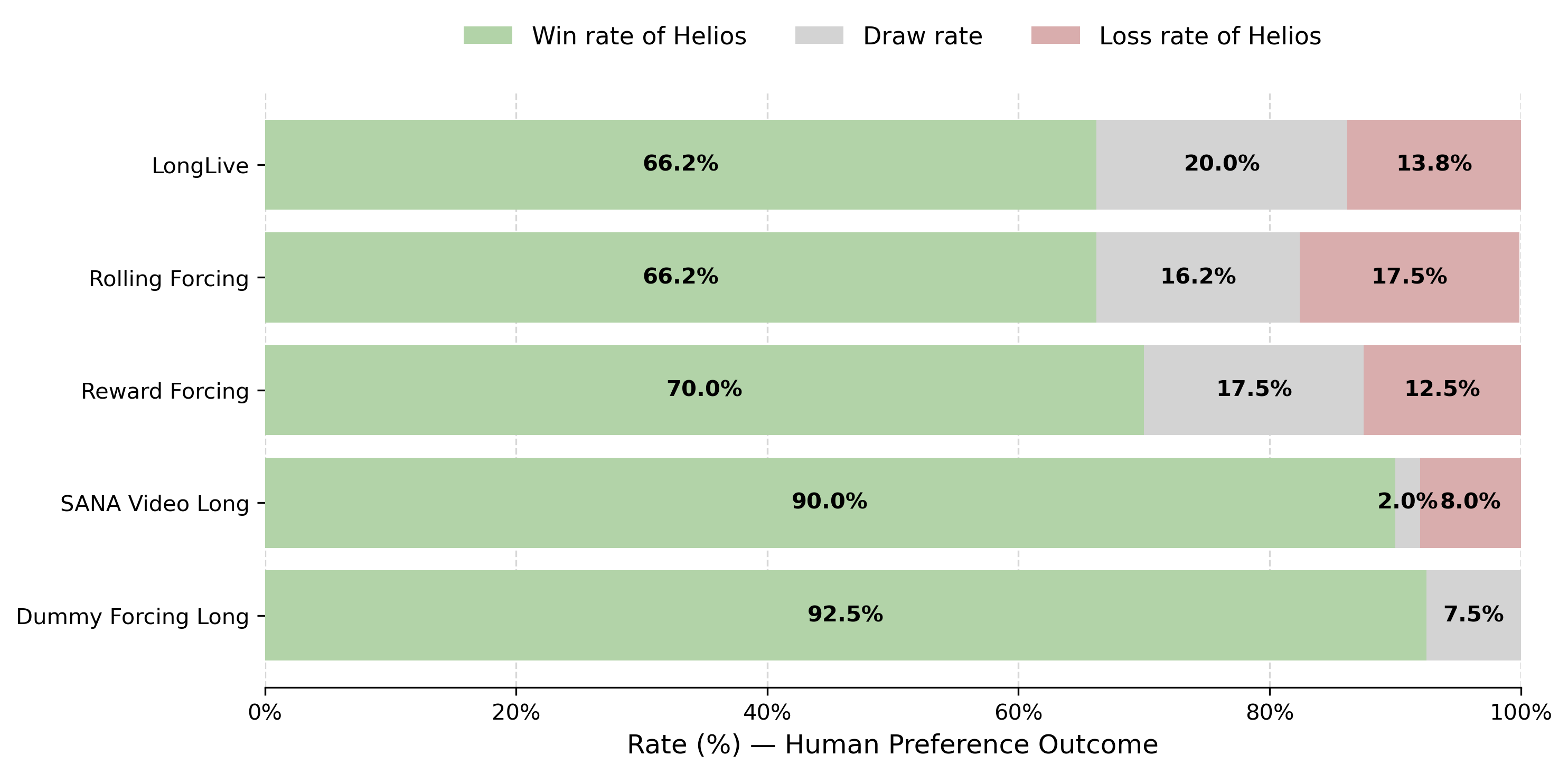}
  \hfill
  \includegraphics[width=0.49\linewidth]{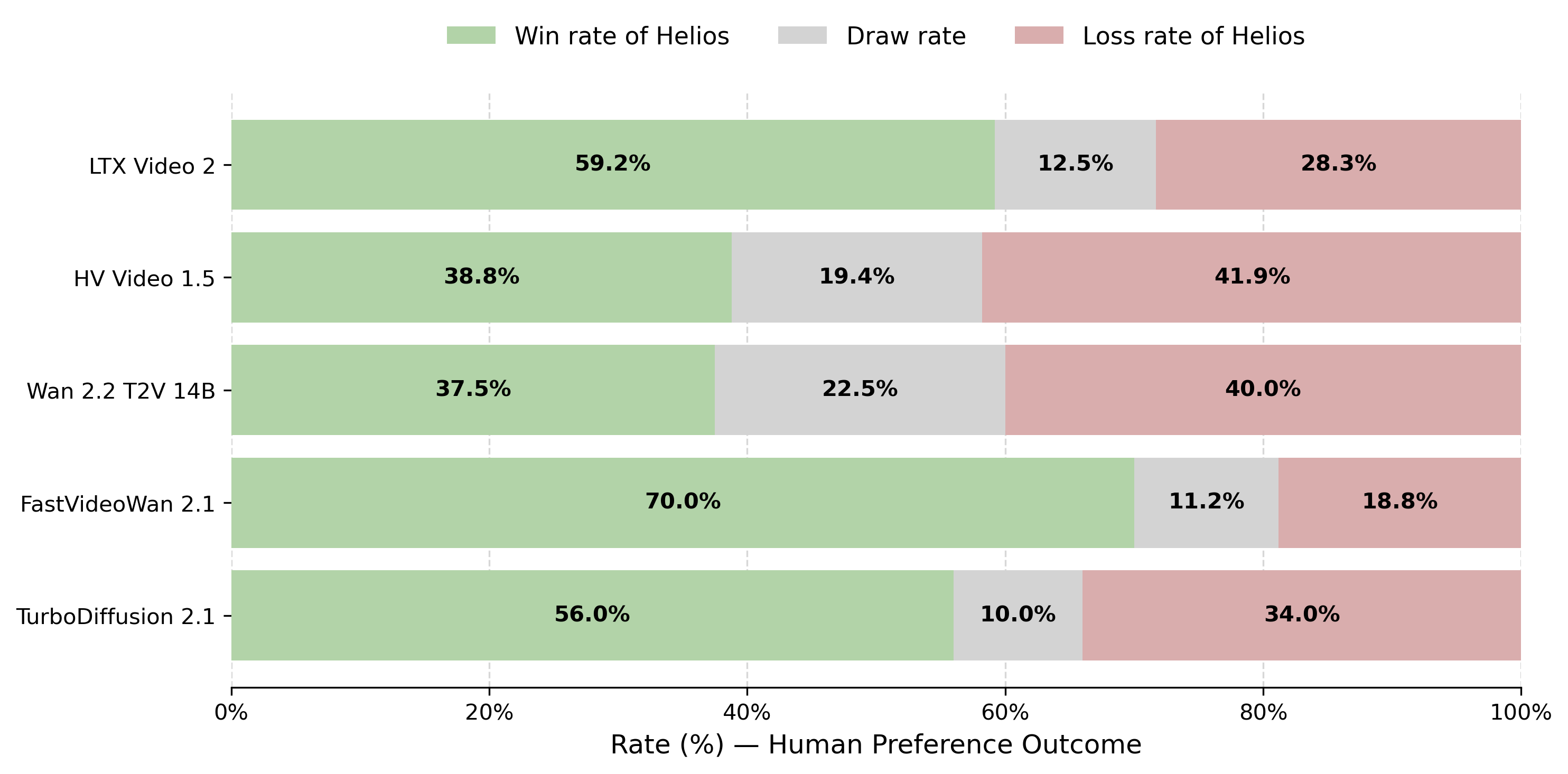}
  \caption{\textbf{Side-by-side human evaluation of \textit{Helios} versus counterparts.} Left: Long Video; Right: Short Video}
  \label{figure_side-by-side_evaluation}
\end{figure*}

We conduct a side-by-side user study with five representative models for real-time long-video generation \cite{rewardforcing, rollingforcing, longlive, sana_video, dummyforcing} and five for short-video generation \cite{longcat_video, wan, hunyuanvideo15, ltx2, sana_video}. In each trial, participants view two videos and indicate whether one is better or whether they are comparable. Each questionnaire contains 40 pairwise comparisons, requiring participants to watch 80 clips in total. We randomize both the presentation order and the left--right placement to reduce bias. Each participant completes only one questionnaire to avoid information leakage and improve engagement. We collect 200 valid responses. As shown in Figure~\ref{figure_side-by-side_evaluation}, \textit{\ours} consistently outperforms prior methods on both long- and short-video generation.

% 在延续先前工作的基础上，我们选择了五个代表性模型的实时长视频生成模型进行Side-by-Side（SBS）评估。在该评估中，给定两段视频，要求评估员判断哪一段视频更好或两者不相上下。每份问卷包含关于两个模型的40个问题，参与者需要观看80个视频片段。来自不同模型的生成内容会随机混合显示，并且左右位置也会随机排序。每个参与者只能填写一份问卷，这种设置旨在防止信息泄露，并提高参与度和问卷的有效性。由于评估需要大量的参与者，我们成功收集了XXX份有效问卷。结果如Figure XXX所示，表明我们的方法显著优于现有的长视频实时生成方法，进一步验证了Helios的先进性。

\subsection{Ablation Study}

We evaluate key components through qualitative and quantitative comparisons. For simplicity, we omit the Throughput score from the quantitative results below.

\begin{table}[!t]
  \centering
  \caption{\textbf{Quantitative ablation of key components.} Changing any component leads to a significant degradation in quality while exacerbating temporal drifting, thereby impairing \ours’s ability to generate stable long videos. \textit{w Guidance Attention*} indicates the additional use of a causal mask for self-attention. \textit{w Staged Backward Simulation*} means incorporating multi-scale $x_0^k$ into real/fake-score estimators during training.}
  \label{tab:performance_ablation}
  \resizebox{\textwidth}{!}{
    \begin{tabular}{c|cccccc|cccc}
      \toprule
      \textbf{Model} &
      \makecell{\textbf{Total}\\\textbf{$\uparrow$}} & \makecell{\textbf{Aesthetic}\\\textbf{$\uparrow \downarrow$}} & \makecell{\textbf{Dynamic}\\\textbf{$\uparrow \downarrow$}} & \makecell{\textbf{Smoothness}\\\textbf{$\uparrow \downarrow$}} & \makecell{\textbf{Semantic}\\\textbf{$\uparrow$}} & \makecell{\textbf{Naturalness}\\\textbf{$\uparrow$}} &
      \makecell{\textbf{Drifting}\\\textbf{Aesthetic $\uparrow$}} & \makecell{\textbf{Drifting}\\\textbf{Smoothness $\uparrow$}} & \makecell{\textbf{Drifting}\\\textbf{Semantic $\uparrow$}} & \makecell{\textbf{Drifting}\\\textbf{Naturalness $\uparrow$}} \\
      \midrule
      \rowcolor{myblue!30} \textbf{Helios-Base}       & 6.47 & 8 & 6 & 9  & 6 & 5 & 7 & 10 & 8 & 5 \\
      \textit{w} Guidance Attention*                  & \multicolumn{10}{c}{\textit{unstable training process}} \\
      \textit{w/o} Guidance Attention                 & 6.23 & 9 & 4 & 7  & 7 & 5 & 6 & 10 & 8 & 2 \\
      \textit{w/o} First Frame Anchor                 & 5.51 & 8 & 5 & 8  & 7 & 4 & 3 & 10 & 6 & 2 \\
      \textit{w/o} Frame-Aware Corrupt                & 4.70 & 7 & 4 & 8  & 6 & 4 & 2 & 10 & 3 & 1 \\
      \midrule
      \rowcolor{myblue!30} \textbf{Helios-Distilled}  & 6.34 & 8 & 6 & 10 & 5 & 5 & 7 & 10 & 7 & 7 \\
      \textit{w} Self-Forcing                         & 6.11 & 9 & 6 & 10 & 4 & 5 & 8 & 10 & 7 & 6 \\
      \textit{w} Bidirectional Teacher                & 4.75 & 8 & 6 & 9  & 3 & 4 & 5 & 10 & 4 & 4 \\
      \textit{w} Staged Backward Simulation*          & \multicolumn{10}{c}{\textit{unstable training process}} \\
      \textit{w/o} Coarse-to-Fine Learning            & 5.31 & 8 & 4 & 8  & 4 & 4 & 8 & 9  & 6 & 4 \\
      \textit{w/o} Adversarial Post-Training          & 6.31 & 8 & 8 & 9  & 5 & 4 & 7 & 10 & 7 & 9 \\
      % \midrule
      \textit{w} Decouple DMD                         & 5.21 & 7 & 9  & 7 & 4 & 4 & 7 & 8  & 6 & 4 \\
      \textit{w} Reward-weighted Regression           & 6.23 & 8 & 10 & 9 & 5 & 5 & 8 & 10 & 7 & 4 \\
      \bottomrule
    \end{tabular}
  }
\end{table}

\begin{figure*}[!t]
  \centering
  \includegraphics[width=0.95\linewidth]{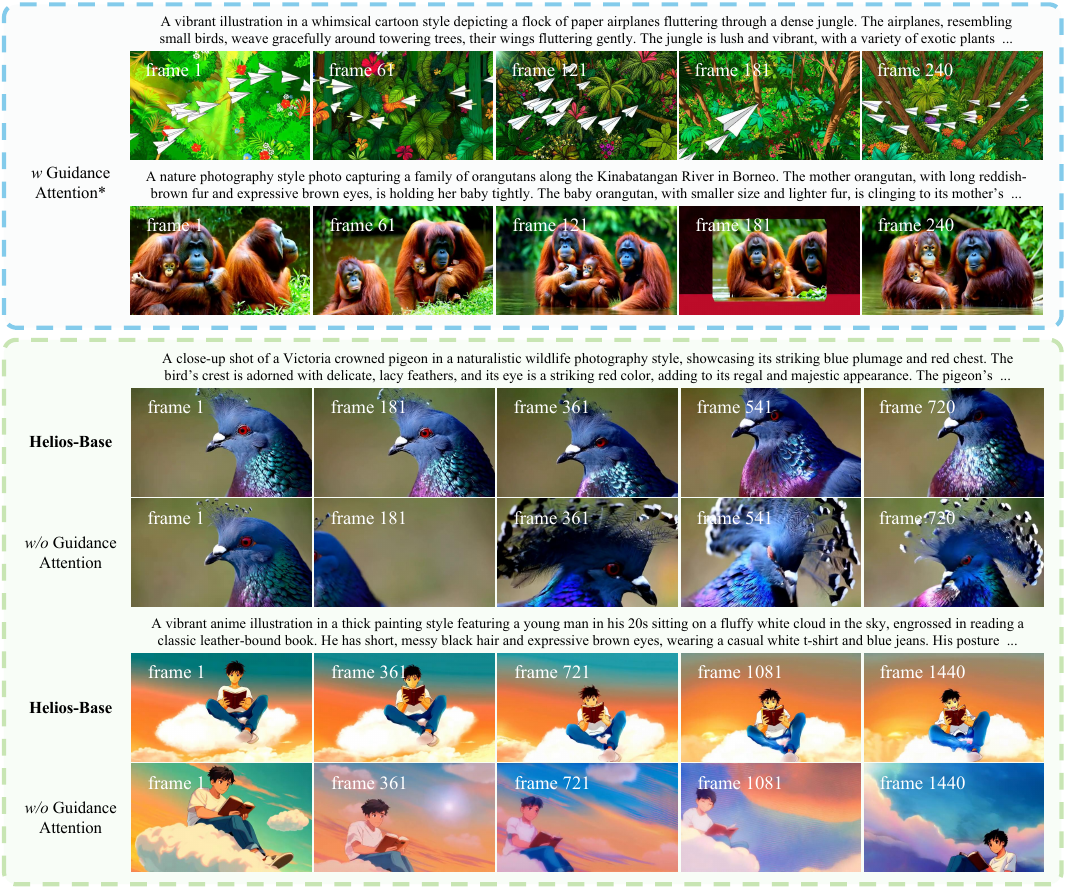}
  \caption{\textbf{Qualitative ablation on Guidance Attention with \textit{Helios}-Base.} Introducing causal mask hinders the ability to learn temporal coherence, causing each generated section to appear independent. Conversely, removing Guidance Attention results in excessive semantic accumulation over time (e.g., a progressively enlarged bird crest). * indicates the additional usage of causal mask to self-attention to prevent noisy context from affecting historical context.}
  \label{figure_qualitative_ablation_part1}
\end{figure*}

\subsubsection{Impact of Guidance Attention}
To study the effect of causal masking on autoregressive generation, we augment Guidance Attention with a causal mask in self-attention, following \cite{causvid, selfforcing}. This mask prevents the noisy context from interfering with the historical context. As shown in Table~\ref{tab:performance_ablation} and Figure~\ref{figure_qualitative_ablation_part1}, causal masking substantially reduces representational capacity and makes optimization harder. We attribute this to the fact that causal masking limits cross-section interactions, undermining temporal coherence across sections; as a result, each section tends to generate an independent new scene. To further evaluate Guidance Attention, we remove it from \textit{\ours-Base}. Without Guidance Attention, the model progressively accumulates semantic content over time, leading to artifacts such as an abnormally enlarged bird crest or steadily increasing saturation.

% 为了验证 causal mask 对自回归模型性能的影响，我们首先在 Guidance Attention 的基础上，参考 XXXX 的做法，在 self-attention 中加入 causal mask，以避免 noisy context 反向干扰 historical context。结果如 Table XXX 和 Figure XXX 所示：引入 causal mask 后，模型的表征能力明显减弱，并且训练过程更难收敛。这表明 causal mask 会限制模型对跨 section 信息的建模能力，使其难以学习不同 section 之间的时序连贯性，从而导致每个 section 更倾向于生成一个相互独立的全新场景。为了进一步验证 Guidance Attention 的有效性，我们将其从\textit{\ours-Base}中移除。结果显示，缺少 Guidance Attention 后，模型容易出现语义信息随时间推移不断累积的问题，例如鸟冠逐渐异常增大，或整体画面饱和度持续升高。

\subsubsection{Impact of First Frame Anchor}
\begin{figure*}[!t]
  \centering
  \includegraphics[width=0.95\linewidth]{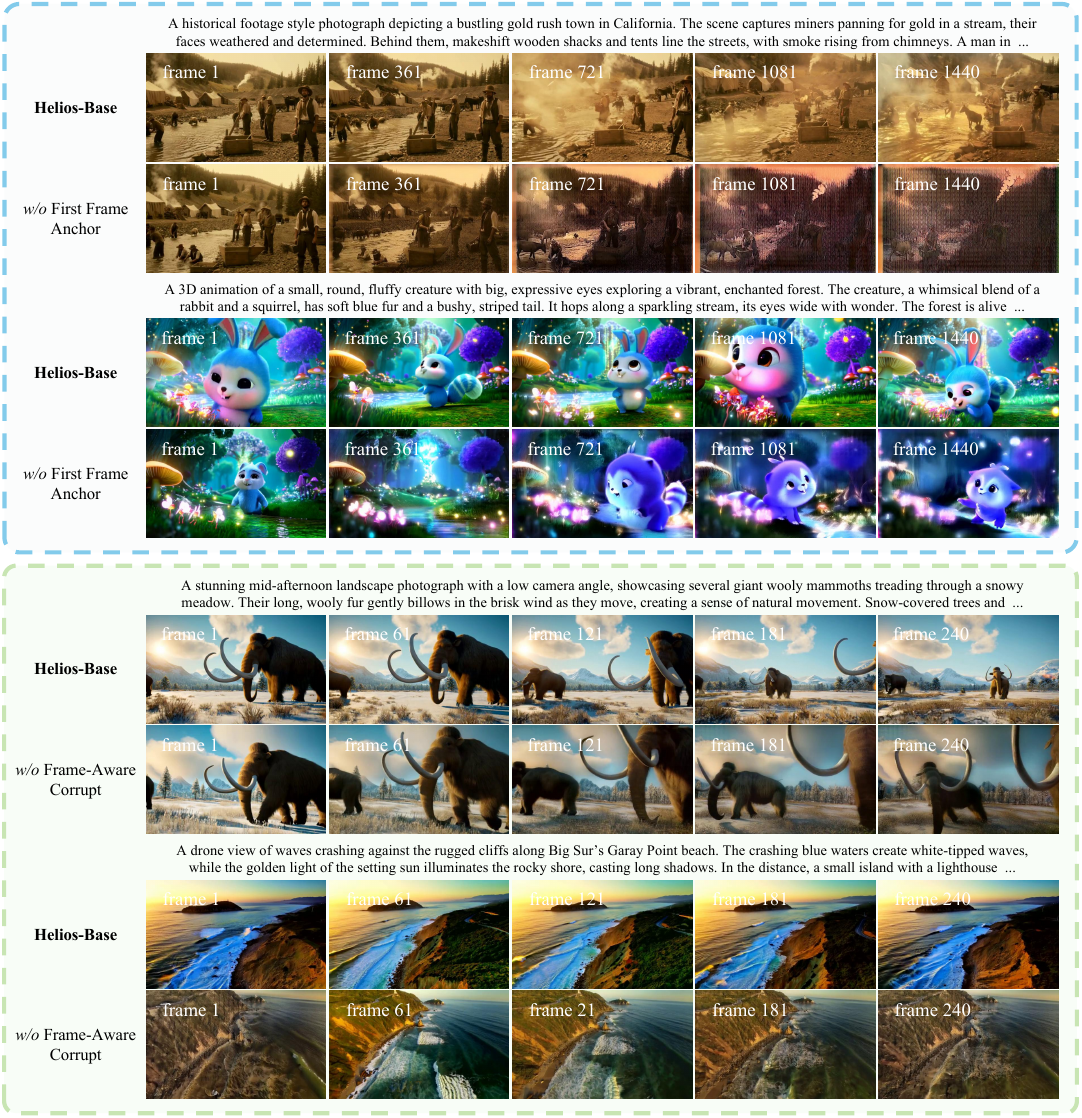}
  \caption{\textbf{Qualitative ablation on First Frame Anchor and Frame-Aware Corrupt with \textit{Helios}-Base.} Removing the First Frame Anchor not only introduces drifting but also causes the subject to deviate from the one in the initial frame. Removing Frame-Aware Corruption leads to noticeable drifting even in videos as short as 240 frames.
  }
  \label{figure_qualitative_ablation_part2}
\end{figure*}

Building on \textit{\ours-Base}, we study the effect of the First Frame Anchor on long-video generation. As shown in Table~\ref{tab:performance_ablation} and Figure~\ref{figure_qualitative_ablation_part2}, this component constrains the global appearance distribution and is crucial for maintaining color consistency. Removing it leads to noticeable degradation as early as frame 720, with errors compounding over longer sequences. The component also improves subject consistency: without it, the subject gradually deviates from the identity established in the first frame, causing cumulative identity drifting.

% 在 \textit{\ours-Base} 的基础上，我们进一步分析 First Frame Anchor 对长视频生成性能的影响。实验结果如 Table XXX 与 Figure XXX 所示：该组件能够有效约束生成序列的全局外观分布，尤其在 全局色彩一致性 的维持上起到关键作用。当移除该组件后，模型在生成至第 720 帧时便可能出现明显的视觉退化，更不用说生成更长的视频序列。此外，First Frame Anchor 还能显著增强模型对 主体一致性 的保持能力。具体而言，缺失该组件会导致主体外观随时间推移逐渐偏离初始帧中的主体特征，从而引发累积性的身份漂移问题。

\subsubsection{Impact of Frame-Aware Corrupt}
As shown in Table~\ref{tab:performance_ablation} and Figure~\ref{figure_qualitative_ablation_part2}, Frame-Aware Corrupt is essential for mitigating error accumulation in long sequences. Removing it causes severe drifting even at 240 frames, leading to a sharp drop in Aesthetic, Semantic, and Naturalness. The degradation becomes more pronounced for minute-scale generation.

% 实验结果如 Table XXX 与 Figure XXX 所示：Frame-Aware Corrupt对抑制长序列生成中的累积误差起到了关键作用。当移除该组件后，即使仅生成 240 帧的视频也会出现严重的退化现象，导致生成质量在 aesthetic、semantic 以及 naturalness 等指标上急剧下降，更不用说生成 minutes-scale 的长视频序列。

\subsubsection{Impact of Multi-Term Memory Patchification}
As shown in Figure~\ref{figure_patchification}, Multi-Term Memory Patchification addresses the poor scalability of naive historical-context modeling. With naive designs, increasing the history length sharply increases the token count, GPU memory footprint, and inference time; in particular, out-of-memory (OOM) errors occur when the context length reaches 6. In contrast, \textit{\ours} decomposes the memory into long-, mid-, and short-term scales and applies scale-specific compression ratios. This design extends the history length to 18 while keeping compute and memory costs stable, thereby avoiding OOM and improving inference speed.

% 如Figure XXX所示，Multi-Term Memory Patchification有效地解决了朴素方法对于historical context的扩展性问题。朴素方法会随着history context的增加而导致标记使用量、显存占用和推理时间急剧上升，特别是当history context length达到6时就会出现OOM。而\textit{\ours}通过将memory分成long-, mid-, shor-term三个尺度并使用不同的压缩比例，即使将history context length扩展到18也能使得这些指标维持在稳定水平，并且避免显存溢出问题并确保更快的推理速度.

\subsubsection{Impact of Pyramid Unified Predictor Corrector}
As shown in Tables~\ref{tab:performance_short} and \ref{tab:performance_long}, extending a single flow trajectory into multiple multi-scale trajectories enables Pyramid Unified Predictor Corrector to nearly double throughput while incurring only a modest performance drop. This gap is further narrowed in Stage~3 via Adversarial Hierarchical Distillation with \textit{\ours-Base} as the teacher; we refer to the resulting model as \textit{\ours-Mid}.

% 如表 XXX 和表 XXX 所示，通过将single flow trajectory 扩展为多尺度的多条 flow trajectories，Pyramid Unified Predictor-Corrector 在仅引入轻微性能损失的情况下，实现了近 2 倍的吞吐率提升。此外，该性能差距可在 Stage-3 中通过以 \textit{\ours-Base} 作为教师模型的 Adversarial Hierarchical Distillation 得到有效弥补。因此，我们将该阶段训练得到的模型称为 \textit{\ours-Mid}。

\begin{figure*}[!t]
  \centering
  \includegraphics[width=0.95\linewidth]{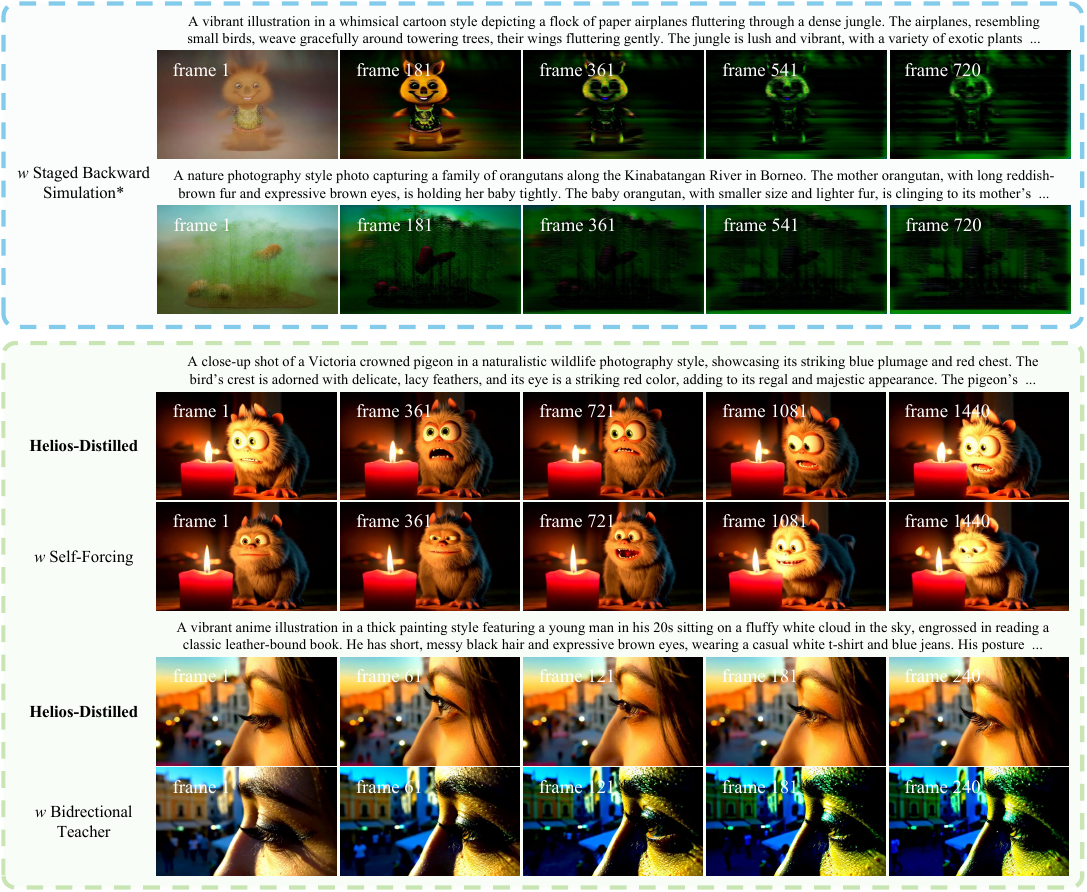}
  \caption{\textbf{Qualitative ablation on Stage Backward Simulation and Pure Teacher Forcing with \textit{Helios}-Distilled.} Feeding multi-scale $x_0^k$ into the fake-score estimator $p_{fake}$ causes the model to converge toward incorrect directions. Pure Teacher Forcing achieves comparable robustness against long-video drifting as Self-Forcing \cite{selfforcing}. * indicates that multi-scale $x_0^k$, rather than only the full-resolution $x_0^K$, is provided to the real/fake-score estimator.
  }
  \label{figure_qualitative_ablation_part3}
\end{figure*}

\begin{figure*}[!t]
  \centering
  \includegraphics[width=0.95\linewidth]{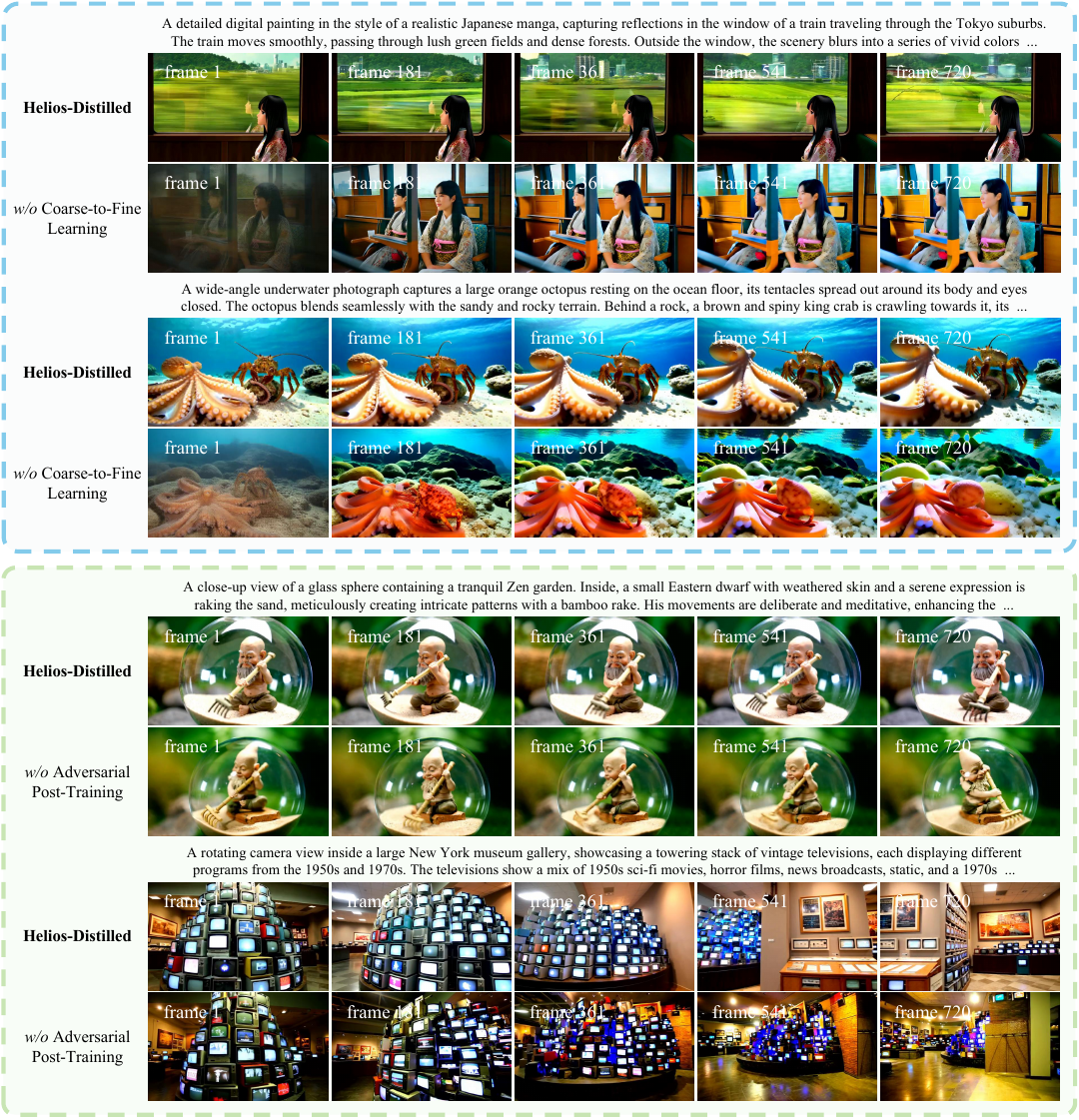}
  \caption{\textbf{Qualitative ablation on Coarse-to-Fine Learning and Adversarial Post-Training with \textit{Helios}-Distilled.} Removing Coarse-to-Fine Learning prevents the model from converging, with particularly unacceptable quality in the first generated section. Removing Adversarial Post-Training leads to degradation in visual quality.
  }
  \label{figure_qualitative_ablation_part4}
\end{figure*}

\begin{figure*}[!t]
  \centering
  \includegraphics[width=0.95\linewidth]{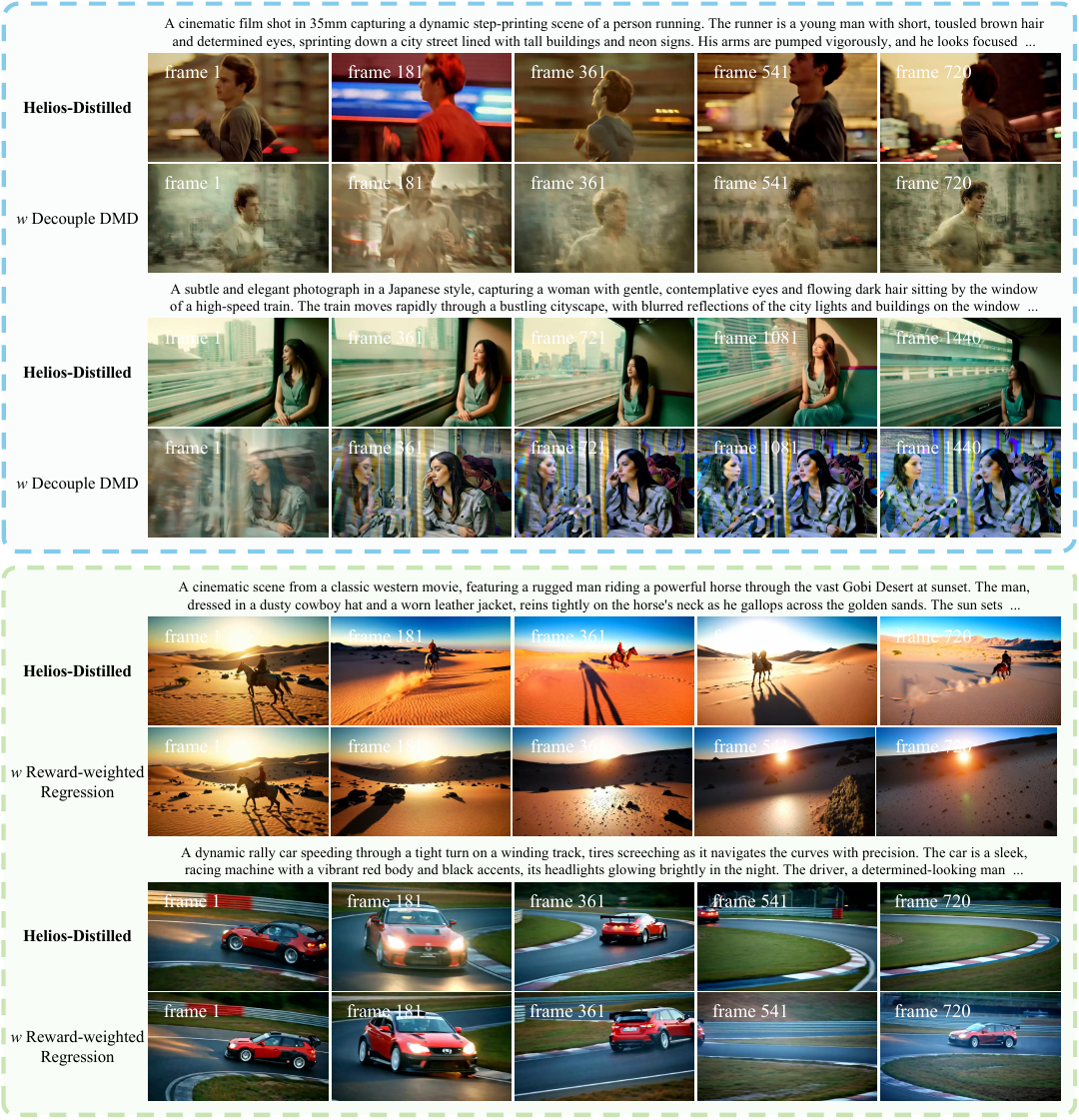}
  \caption{\textbf{Qualitative ablation on Decouple DMD \cite{decoupleddmd} and Reward-weighted Regression \cite{rewardforcing} with \textit{Helios}-Distilled.} The former may hinder convergence and cause grayish outputs, whereas the latter may intensify video flickering.
  }
  \label{figure_qualitative_ablation_part5}
\end{figure*}

\subsubsection{Impact of Pure Teacher Forcing with Autoregressive Teacher}
To evaluate Pure Teacher Forcing, we compare against Self-Forcing \cite{selfforcing} with long rollouts \cite{longlive, rollingforcing, selfforcing++, rewardforcing}. As shown in Figure~\ref{figure_qualitative_ablation_part3} and Table~\ref{tab:performance_ablation}, \textit{\ours-Distilled} achieves robustness to long-horizon drift comparable to Self-Forcing. This result suggests that we can obtain similar benefits without long rollouts, substantially reducing training overhead. We further study the impact of teacher architecture by replacing the Autoregressive Teacher with a Bidirectional Teacher (Wan-2.1-T2V-14B \cite{wan}). For a fair comparison, we feed 21-frame segments to the real/fake score estimators at each step, following \cite{selfforcing, longlive, rollingforcing, selfforcing++, rewardforcing}. Overall, \textit{\ours-Distilled} with an Autoregressive Teacher outperforms the variant that uses a Bidirectional Teacher.

% To evaluate Pure Teacher Forcing, we compare it with long-rollout self-forcing \cite{longlive, selfforcing++}, using the bidirectional Wan-2.1-T2V-14B \cite{wan} as the teacher. For a fair comparison, we feed the real/fake score estimators 21-frame sections at each step. As shown in Figure~\ref{figure_qualitative_ablation_part3} and Table~\ref{tab:performance_ablation}, \textit{\ours-Distilled} trained with an autoregressive teacher achieves better overall performance and exhibits robustness to long-horizon drifting comparable to self-forcing. Importantly, it attains these gains without long rollouts, substantially reducing training overhead.

% 为了验证 Pure Teacher Forcing 机制的有效性，我们将其与采用 Long-rollout 的 Self-forcing 策略进行了对比实验。如 Figure XXX 和 Table XXX 所示，\textit{\ours-Distilled} 在长视频生成任务中展现出了与 Self-forcing 相当的抗退化能力。这一结果表明，该模型无需依赖高开销的 Long-rollout 机制即可达到同等性能，从而显著降低了训练成本。在此基础上，为进一步探讨教师模型架构的影响，我们将原有的自回归教师模型（Autoregressive Teacher）替换为双向教师模型（Bidirectional Teacher，即 Wan-2.1-T2V-14B）。为确保评估条件的一致性，在该设置下，我们统一了输入至 Real/Fake-score Estimator 的分段（Section）帧数（均为 21 帧）。实验对比显示，采用自回归教师模型的 \textit{\ours-Distilled} 在整体性能上仍优于基于双向教师模型的方案。

\subsubsection{Impact of Staged Backward Simulation}
Staged Backward Simulation produces multi-scale estimates $x_0^k$. We can further interpolate these intermediate results to the $t=0$ (i.e., noise-free) state using the following formula and feed them to $p_{real}$ and $p_{fake}$:
\begin{equation}
  x_0^{k'} = (x_0^k - \lambda_t \cdot \epsilon) / (1 - \lambda_t).
\end{equation}
In principle, incorporating multi-scale $x_0^k$ during training can provide richer supervision and improve robustness across resolutions. In practice, however, we feed only the full-scale $x_0^K$ to the real/fake score estimators. As shown in Figure~\ref{figure_qualitative_ablation_part3}, directly providing multi-scale $x_0^k$ to the fake-score estimator $p_{\text{fake}}$ causes the optimization to converge to an undesirable solution, resulting in a significant performance drop.

% 在 Staged Backward Simulation 中，我们可以获得多尺度的 $x_0^k$。进一步地，这些中间结果可通过如下公式插值至 $t=0$（即无噪声）状态，并作为输入送入 real/fake-score estimator：
% \begin{equation}
% x_0^{k'} = (x_0^k - \lambda_t \cdot \epsilon) / (1 - \lambda_t).
% \end{equation}
% 尽管从理论上看，将多尺度的 $x_0^k$ 一并用于训练能够使学习过程更加充分且合理，但在实际实现中，我们仅将 full-scale 的 $x_0^K$ 输入至 real/fake-score estimator。实验发现，若将多尺度 $x_0^k$ 直接输入至 fake-score estimator $p_{\text{fake}}$，模型的优化过程会偏向错误的收敛方向，导致性能显著退化，如 Figure XXX 所示。

\subsubsection{Impact of Coarse-to-Fine Learning}
As shown in Figure~\ref{figure_qualitative_ablation_part4} and Table~\ref{tab:performance_ablation}, Coarse-to-Fine Learning not only improves early-stage performance but also is critical for stable convergence. With the same training budget, \textit{\ours-Base} already generates videos that better align with human perception. In contrast, removing Coarse-to-Fine Learning disrupts convergence and prevents quality improvements, especially for the first generated section.

% 如图XXX和表XXX所示，我们的Coarse-to-Fine Learning不仅显著提升了初始训练阶段的性能，还对能否正常收敛起到不可或缺的作用。具体而言，训练相同的step，\textit{\ours-Base}早已能够生成符合人类感知的视频，而移除Coarse-to-Fine Learning会阻碍模型收敛，导致视频质量难以得到提升，尤其是在生成的第一个section时。

\begin{figure*}[!t]
  \centering
  \includegraphics[width=0.885\linewidth]{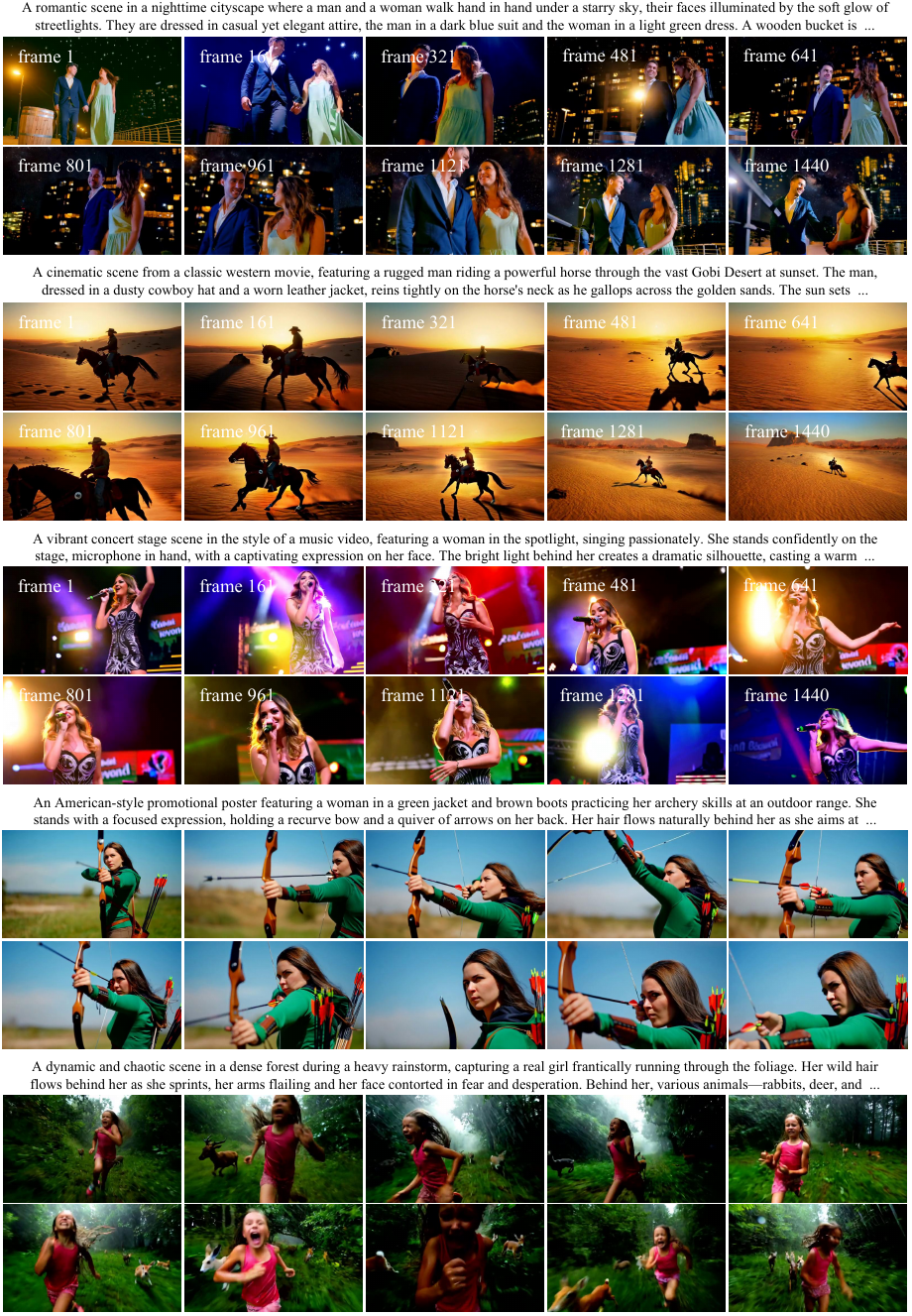}
  \caption{\textbf{Text-to-Video showcases.} Part of the prompts are sourced from \cite{selfforcing}.
  }
  \label{figure_showcases_t2v}
\end{figure*}

\begin{figure*}[!t]
  \centering
  \includegraphics[width=0.885\linewidth]{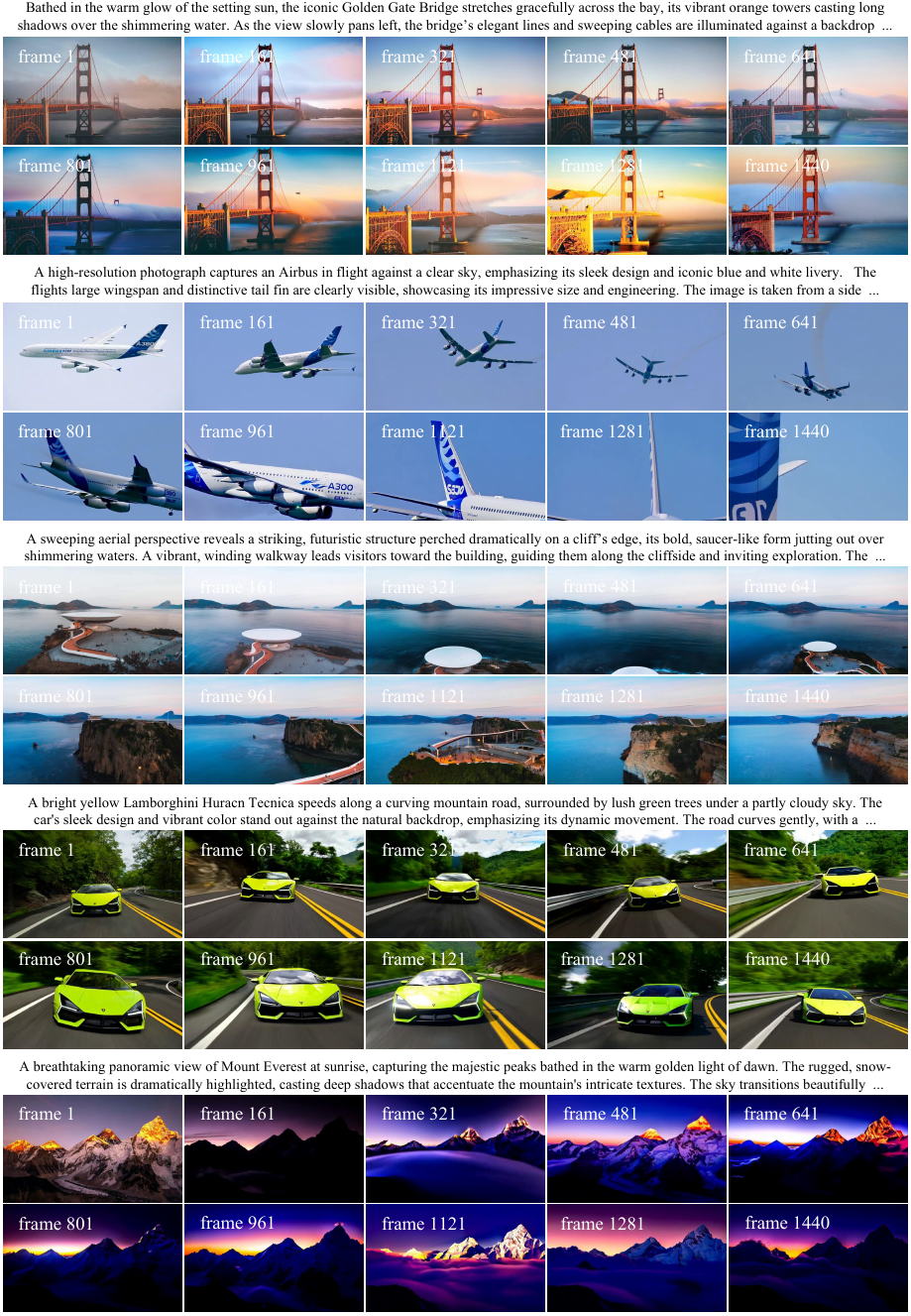}
  \caption{\textbf{Image-to-Video showcases.} Part of the images are sourced from \cite{vbench, SVI}.
  }
  \label{figure_showcases_i2v}
\end{figure*}

\begin{figure*}[!t]
  \centering
  \includegraphics[width=0.93\linewidth]{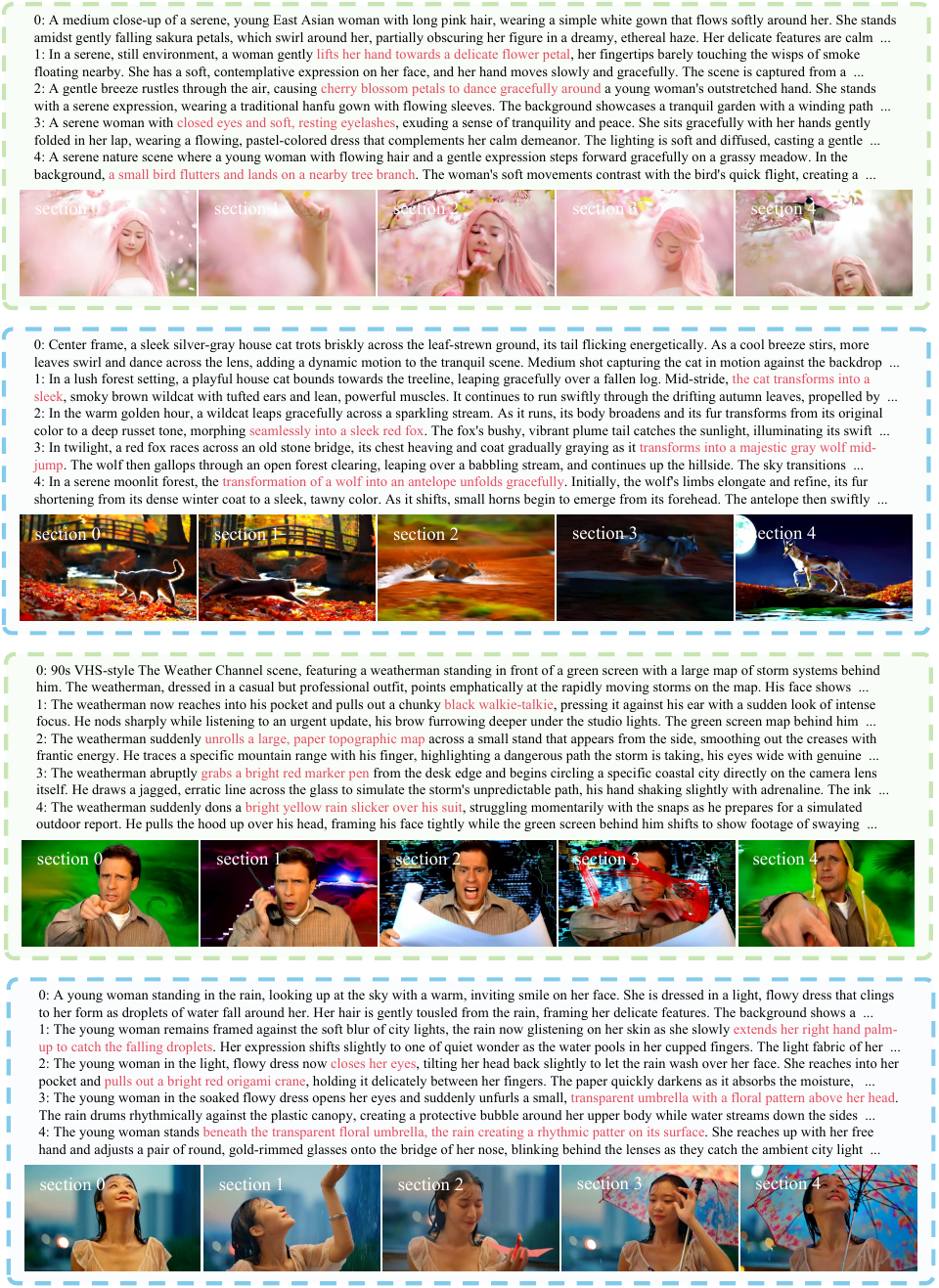}
  \caption{\textbf{Interactive-to-Video showcases.} Part of the prompts are sourced from \cite{longlive, vidprom}.
  }
  \label{figure_showcases_interactive2v}
\end{figure*}

\subsubsection{Impact of Adversarial Post-Training}
Adversarial post-training leverages real data to mitigate the performance ceiling imposed by the teacher model during distillation. By introducing an adversarial objective in the distilled stage, the student model is encouraged not only to mimic the teacher's outputs but also to better align with the distribution of real data, thereby enhancing realism beyond what pure distillation can achieve. To assess its impact, we remove it in an ablation study. As shown in Figure~\ref{figure_qualitative_ablation_part4} and Table~\ref{tab:performance_ablation}, disabling this object noticeably degrades visual quality, particularly in naturalness and realism, highlighting its importance for improving perceptual fidelity.

% Adversarial Post-Training 的初衷是利用真实数据进行训练，从而避免蒸馏模型的性能受限于教师模型的表达能力。为验证该策略是否能够为 \textit{\ours} 带来性能增益，我们将 Adversarial Post-Training 移除并进行消融实验，结果如图 XXX 和表 XXX 所示。可以观察到，移除 Adversarial Post-Training 会导致生成结果的视觉质量明显下降，具体表现为结果不够自然和真实。这表明 该策略在提升模型的感知质量和逼真度方面起到了关键作用。

\subsubsection{Impact of Flash Normalization and Flash RoPE}
To evaluate Flash Normalization and Flash RoPE, we measure the inference and training time of Wan-2.1-T2V-14B \cite{wan} for 50 steps at $384 \times 640$ with 81-frame inputs. As reported in Table~\ref{tab:performance_triton}, Triton-optimized kernels substantially improve throughput, which is crucial for real-time long-video generation.

% 为了验证Flash Normalization and Flash RoPE的有效性，我们记录了Wan-2.1-T2V-14B在384x640分辨率及81帧视频设置下，推理50步和训练50步所需时间的对比。结果如表XXX所示，通过使用triton重写的kernels能够有效增加吞吐量，which对于实时视频生成至关重要。

\subsubsection{Ablation on Decouple DMD}
% Decouple DMD \cite{decoupleddmd} rewrites the DMD objective \cite{DMD, DMD2} as the sum of CFG Augmentation (CA) and Distribution Matching (DM), with separate weights, and reports improved performance. Following \cite{decoupleddmd}, we adapt this formulation from image to video generation. As shown in Figure~\ref{figure_qualitative_ablation_part5} and Table~\ref{tab:performance_ablation}, it converges more slowly and produces grayish outputs with occasional local jitter and grid-like artifacts. We therefore do not use Decouple DMD in the final model.

Decouple DMD \cite{decoupleddmd} reformulates the original DMD objective \cite{DMD, DMD2} as the weighted sum of two disentangled components: CFG Augmentation (CA) and Distribution Matching (DM). By assigning separate weights to these terms, it provides more flexible control over the optimization process and has been shown to yield improved performance in image generation settings. Following \cite{decoupleddmd}, we extend this formulation from image to video generation and integrate it into our framework. However, as illustrated in Figure~\ref{figure_qualitative_ablation_part5} and Table~\ref{tab:performance_ablation}, this variant exhibits noticeably slower convergence compared to our default setting. Moreover, the generated videos tend to suffer from grayish tones, occasional local jitter, and grid-like artifacts, indicating suboptimal temporal and spatial consistency. Given these limitations, we do not adopt Decouple DMD in our final model.

% Decouple DMD将DMD training objective分解为CFG Augmentation (CA)和Distribution Matching (DM)两个term并使用不同的加权系数，并宣称该training objective能够取得更好的效果。Following Decouple DMD，我们将该方法从image generation迁移到video generation以验证是否能获得增益。结果如图XXX和表XXX所示，我们发现采用该方法会导致收敛慢，生成的视频偏灰，且偶尔有局部抖动、网格现象，因此我们最后并未采用该方法。

\begin{table}[!t]
    \centering
    \caption{\textbf{Quantitative ablation on Flash Normalization and Flash RoPE.} We report the total runtime of the DiT component in Helios, measured over 50 forward passes and 50 forward-backward passes on a single NVIDIA H100 GPU.}
    \label{tab:performance_triton}
    \resizebox{0.6\textwidth}{!}{
        \begin{tabular}{c|cc}
            \toprule
            & \textbf{Inference Time (s)} & \textbf{Training Time (s)} \\
            \midrule
            % Baseline & 1.083 & 4.302 \\
            % \textit{w} Flash Normalization & 0.9742 & 3.8406 \\
            % \textit{w} Flash RoPE & 1.0208 & 4.0924 \\
            % \textit{w} Flash Normalization and Flash RoPE & 0.9093 & 3.5968 \\
            % Baseline & 54.15 & 215.10 \\
            % \textit{w} Flash Normalization & 48.71 & 192.03 \\
            % \textit{w} Flash RoPE & 51.04 & 204.62 \\
            % \textit{w} Flash Normalization and Flash RoPE & 45.46 & 179.84 \\
            Wan-2.1-T2V-14B \cite{wan}                    & 98.68 & 398.03 \\
            \textit{w} Flash Normalization                & 89.91 & 360.77 \\
            \textit{w} Flash RoPE                         & 93.39 & 378.77 \\
            \textit{w} Flash Normalization and Flash RoPE & 84.41 & 340.38 \\
            \bottomrule
        \end{tabular}
    }
\end{table}

\subsubsection{Ablation on Reinforcement Post-Training}
Beyond adversarial objectives that improve the upper bound, recent post-training methods attempt to raise the lower bound via reinforcement learning (RL) \cite{dancegrpo, flowgrpo, videoalign} Reward Forcing \cite{rewardforcing}, we weight the standard DMD \cite{DMD, DMD2}, which can be interpreted as a form of reward-weighted regression \cite{peters2007reinforcement, videoalign}:
\begin{equation}
  \mathbb{E}_{y,x_0,}
  \Big[
    \exp\!\left(\frac{r(x_0,y)}{\beta}\right)
    \cdot
    \log \frac{p_{\mathrm{fake}}(x_0 \mid y)}{p_{\mathrm{real}}(x_0 \mid y)}
  \Big]
  \;=\;
  \mathbb{E}_{y,x_0}
  \Big[
    \mathcal{R}_{\mathrm{rl}}
    \cdot
    \mathcal{L}_{\mathrm{DMD}}
  \Big].
\end{equation}
Here, $p_{\mathrm{ref}}$ denotes the reference distribution (e.g., the teacher), and $\beta$ controls the trade-off between reward maximization and distribution shift. We use VideoAlign \cite{videoalign} as the reward model and its Motion Quality as the score. As shown in Figure~\ref{figure_qualitative_ablation_part5} and Table~\ref{tab:performance_ablation}, reinforcement learning consistently degrades performance: semantic and aesthetic scores drop, and the outputs exhibit severe flickering. We therefore exclude RL.

% 后训练阶段除了通过添加adversarial object提高模型上限外，新兴的做法是使用reinforcement learning提高模型的下限。Following Reward Forcing，我们在\textit{\ours-Distilled}的基础上使用奖励函数对标准 DMD 损失进行加权优化，which是reward-weighted regression的一个变种：
% \begin{equation}
%     \mathbb{E}_{y,x_0,}
%     \Big[
%     \exp\!\left(\frac{r(x_0,y)}{\beta}\right)
%     \cdot
%     \log \frac{p_{\mathrm{fake}}(x_0 \mid y)}{p_{\mathrm{real}}(x_0 \mid y)}
%     \Big]
%     \;=\;
%     \mathbb{E}_{y,x_0}
%     \Big[
%     \mathcal{R}_{\mathrm{rl}}
%     \cdot
%     \mathcal{L}_{\mathrm{DMD}}
%     \Big].
% \end{equation}
% 其中 $p_{\mathrm{ref}}$ 表示参考分布（如教师模型），$\beta$ 控制奖励优化与分布偏移之间的权衡。我们选用 VideoAlign 作为 reward model，并将其中的Motion Quality子指标作为最终的reward score。结果如图XXX和表XXX所示，可见采用reinforcement learning后模型的结果不升反降，除了导致semantic、aesthetic等下降外，更严重的是会导致生成的视频出现明显的闪烁现象，因此我们最后并未采用该方法。

\section{Application}\label{sec: application}
Benefiting from the proposed Representation Control, \textit{\ours} continues to adopt the conventional text-to-video pipeline for data preparation and model optimization during training. However, since the historical context is randomly zeroed out with a certain probability throughout training, the model can naturally generalize at inference time and seamlessly support T2V, I2V, and V2V tasks. The showcases are presented in Figure \ref{figure_showcases_i2v} and Figure \ref{figure_showcases_t2v}, demonstrating satisfactory quality. Moreover, by incorporating Interactive Interpolation, \textit{\ours} further enables, in a zero-shot manner, a key capability of world models -- Interactive Generation. This mechanism allows users to dynamically modify the input prompt during the generation process, thereby providing real-time control over the generated content. Some representative showcases are shown in Figure \ref{figure_showcases_interactive2v}.

% 得益于所提出的 Representation Control，尽管 \textit{\ours} 在训练阶段仍按照传统的 text-to-video 流程进行数据准备与模型优化，但由于训练过程中会以一定比例对 historical context 进行置零处理，因此模型在推理阶段能够自然泛化并同时支持 T2V、I2V 和 V2V 任务，相关效果展示如图 XXX 所示。此外，通过引入 Interactive Interpolation，\textit{\ours} 进一步以 zero-shot 的方式支持世界模型中的一项关键任务——交互式生成。该机制允许用户在生成过程中动态修改输入 prompt，从而实现对输出内容的实时控制。具体示例如图 XXX 所示。

\section{Conclusion}\label{sec: conclusion}

In this paper, we presented \ours, the first 14B video generation model that runs at 19.5 FPS on a single NVIDIA H100 GPU and supports minute-scale generation, without relying on conventional anti-drifting strategies or acceleration techniques. \ours adopts a unified input representation that supports text-to-video, image-to-video, and video-to-video tasks within a single framework. To mitigate drifting in long-video generation, we analyze typical failure modes and develop a set of simple yet effective training strategies that explicitly simulate drifting. In terms of efficiency, we reduce computational cost to a level comparable to---or even lower than---that of image diffusion models by aggressively compressing the historical context, the noisy context, and the number of sampling steps. Extensive experiments show that \ours consistently outperforms existing methods across multiple dimensions for both short and long videos.

% 本文提出了一种名\ours的神经网络结构，旨在无需常见抗退化策略及加速技巧的前提下，解决超大规模视频生成模型无法实时生成高质量长视频的难题。\ours采用统一输入表示,同时支持T2V、I2V和V2V任务。针对长视频生成的drifting问题,我们分析了其典型表现并设计了一系列简洁有效的策略,在训练阶段主动模拟退化过程。在计算效率上,我们通过深度压缩历史上下文、当前上下文及采样步数,使计算开销与图像扩散模型相当甚至更低。大量的实验表明,无论短视频还是长视频,\ours都在各个维度显著超越现有方法。

\section{Limitations and Future Work}\label{sec: limitation and future work}
(1) Existing metrics \cite{vbench, vbench2, vbench++} are insufficient for accurately assessing the performance of video generation models. Although \ours can produce realistic and natural videos conditioned on text prompts, widely used metrics (e.g., Aesthetic and Smoothness) exhibit only marginal differences compared to prior approaches. A promising direction is to develop perceptually aligned metrics that better reflect human judgment.

(2) Similar to existing autoregressive models \cite{selfforcing++, longlive, rollingforcing, rewardforcing, causvid, framepack, causalforcing}, the generated segments may suffer from flickering artifacts at stitching boundaries. Nevertheless, our method consistently outperforms previous approaches. Future work could explore reinforcement learning techniques that explicitly optimize smoothness-related objectives to further enhance temporal consistency.

(3) Due to resource constraints, our experiments are limited to a resolution of $384 \times 640$, leaving higher-resolution settings unexplored. This work prioritizes real-time long-video generation as a core capability for world models \cite{sora, genie, genie3, stableworld, yume, yume15, hv_world15, lingbot_world, magictime, rethinking_video_generation} and video generation \cite{fsvideo, opensoraplan, opensora, hunyuanvideo, hunyuanvideo15, wan, ltx, ltx2, skyreelsv2, skyreelsv3, stepvideo, kandinsky5, mochi}, without a dedicated design for long-video memory. Both remain important directions for future work.

% (1). 现有指标无法准确评估不同长视频生成模型的性能。虽然\ours可以根据文本提示生成逼真自然的视频，但常用指标（例如Aesthetic和Smoothness）得到的结果与以往方法相比差异甚微，一个有前景的方法是开发一种更符合人类感知的指标。
% (2). 与现有的自回归模型（如self-forcing++、longlive、rollingforcing、self-forcing、causvid、framepack等）一样，生成的视频片段在拼接处有可能会出现闪烁现象，但仍然比现有方法好，未来或许可以考虑使用与Smoothness相关的RL算法进行改进。
% (3). 由于资源限制，我们仅在384×640分辨率下进行了快速验证，未能探索更高分辨率的效果。此外，本工作聚焦于世界模型与视频生成领域中的实时生成这一核心特性，暂未针对长视频记忆机制进行深入设计。上述两点均为未来工作的重要优化方向。

\section{Declaration of LLM Usage}
We used advanced natural language processing models, such as GPT, to assist in preparing this manuscript. Specifically, GPT supported language-related tasks including grammar correction, spell-checking, and expression refinement, improving the paper's precision and readability. GPT also helped with data organization, selection, and preliminary figure design, as well as drafting alternative phrasing for clarity. All academic content, analyses, and conclusions were independently developed, validated, and interpreted by the authors.

% 我们采用了先进的自然语言处理模型，如GPT，来协助本文的撰写。具体来说，GPT 被用于处理文本中的语言任务，包括语法修正、拼写校对和表达优化，从而提升文稿的表达精度和可读性。此外，GPT 还参与了数据的整理、筛选以及图表的初步设计。所有的学术内容、分析过程以及结论均由作者独立进行构思、验证和解释。

\section{Ethics Statement}
\ours is capable of generating high-quality, realistic videos. However, it also carries potential negative consequences, as the technology could be exploited to create deceptive content for fraudulent purposes. It is crucial to acknowledge that all technologies are vulnerable to misuse.

\section{Acknowledgment}\label{sec: acknowledgment}
We thank all collaborators for their constructive comments and support throughout this project. This project is for research purposes only and is not integrated into any ByteDance products.

\clearpage

\bibliographystyle{plainnat}
\setlength{\bibhang}{0pt}
\setlength\bibindent{0pt}
\bibliography{main}

@String(CVPR= {IEEE Conf. Comput. Vis. Pattern Recog.})

@String(ICLR = {Int. Conf. Learn. Represent.})

@String(CVPR  = {CVPR})

@String(ICLR  = {ICLR})

@inproceedings{sora,
  title={Video generation models as world simulators},
  author={Tim Brooks and Bill Peebles and Connor Holmes and Will DePue and Yufei Guo and Li Jing and David Schnurr and Joe Taylor and Troy Luhman and Eric Luhman and Clarence Ng and Ricky Wang and Aditya Ramesh},
  year={2024},
  booktitle={openai},
  url={https://openai.com/research/video-generation-models-as-world-simulators},
}

@inproceedings{vbench,
  title={Vbench: Comprehensive benchmark suite for video generative models},
  author={Huang, Ziqi and He, Yinan and Yu, Jiashuo and Zhang, Fan and Si, Chenyang and Jiang, Yuming and Zhang, Yuanhan and Wu, Tianxing and Jin, Qingyang and Chanpaisit, Nattapol and others},
  booktitle={Proceedings of the IEEE/CVF Conference on Computer Vision and Pattern Recognition},
  pages={21807--21818},
  year={2024}
}

@article{rewardforcing,
    title={Reward forcing: Efficient streaming video generation with rewarded distribution matching distillation},
    author={Lu, Yunhong and Zeng, Yanhong and Li, Haobo and Ouyang, Hao and Wang, Qiuyu and Cheng, Ka Leong and Zhu, Jiapeng and Cao, Hengyuan and Zhang, Zhipeng and Zhu, Xing and others},
    journal={arXiv preprint arXiv:2512.04678},
    year={2025}
}

@article{dummyforcing,
    title={Efficient Autoregressive Video Diffusion with Dummy Head}, 
    author={Hang Guo and Zhaoyang Jia and Jiahao Li and Bin Li and Yuanhao Cai and Jiangshan Wang and Yawei Li and Yan Lu},
    year={2026},
    journal={arXiv preprint arXiv:2601.20499}, 
}

@article{LoL,
  title={LoL: Longer than Longer, Scaling Video Generation to Hour},
  author={Cui, Justin and Wu, Jie and Li, Ming and Yang, Tao and Li, Xiaojie and Wang, Rui and Bai, Andrew and Ban, Yuanhao and Hsieh, Cho-Jui},
  journal={arXiv preprint arXiv:2601.16914},
  year={2026}
}

@article{RoPE,
  title={Roformer: Enhanced transformer with rotary position embedding},
  author={Su, Jianlin and Ahmed, Murtadha and Lu, Yu and Pan, Shengfeng and Bo, Wen and Liu, Yunfeng},
  journal={Neurocomputing},
  volume={568},
  pages={127063},
  year={2024},
  publisher={Elsevier}
}

@article{attention,
  title={Attention is all you need},
  author={Vaswani, Ashish and Shazeer, Noam and Parmar, Niki and Uszkoreit, Jakob and Jones, Llion and Gomez, Aidan N and Kaiser, {\L}ukasz and Polosukhin, Illia},
  journal={Advances in neural information processing systems},
  volume={30},
  year={2017}
}

@article{z_image,
  title={Z-Image: An Efficient Image Generation Foundation Model with Single-Stream Diffusion Transformer},
  author={Cai, Huanqia and Cao, Sihan and Du, Ruoyi and Gao, Peng and Hoi, Steven and Hou, Zhaohui and Huang, Shijie and Jiang, Dengyang and Jin, Xin and Li, Liangchen and others},
  journal={arXiv preprint arXiv:2511.22699},
  year={2025}
}

@article{qwen_image,
  title={Qwen-image technical report},
  author={Wu, Chenfei and Li, Jiahao and Zhou, Jingren and Lin, Junyang and Gao, Kaiyuan and Yan, Kun and Yin, Sheng-ming and Bai, Shuai and Xu, Xiao and Chen, Yilei and others},
  journal={arXiv preprint arXiv:2508.02324},
  year={2025}
}

@misc{flux_2,
    author={Black Forest Labs},
    title={{FLUX.2: Frontier Visual Intelligence}},
    year={2025},
    howpublished={\url{https://bfl.ai/blog/flux-2}},
}

@article{gan,
  title={Generative adversarial networks},
  author={Goodfellow, Ian and Pouget-Abadie, Jean and Mirza, Mehdi and Xu, Bing and Warde-Farley, David and Ozair, Sherjil and Courville, Aaron and Bengio, Yoshua},
  journal={Communications of the ACM},
  volume={63},
  number={11},
  pages={139--144},
  year={2020},
  publisher={ACM New York, NY, USA}
}

@article{longcat_video,
  title={Longcat-video technical report},
  author={Team, Meituan LongCat and Cai, Xunliang and Huang, Qilong and Kang, Zhuoliang and Li, Hongyu and Liang, Shijun and Ma, Liya and Ren, Siyu and Wei, Xiaoming and Xie, Rixu and others},
  journal={arXiv preprint arXiv:2510.22200},
  year={2025}
}

@article{stableworld,
  title={StableWorld: Towards Stable and Consistent Long Interactive Video Generation},
  author={Yang, Ying and Lv, Zhengyao and Pan, Tianlin and Wang, Haofan and Yang, Binxin and Yin, Hubery and Li, Chen and Liu, Ziwei and Si, Chenyang},
  journal={arXiv preprint arXiv:2601.15281},
  year={2026}
}

@article{captain_cinema,
  title={Captain cinema: Towards short movie generation},
  author={Xiao, Junfei and Yang, Ceyuan and Zhang, Lvmin and Cai, Shengqu and Zhao, Yang and Guo, Yuwei and Wetzstein, Gordon and Agrawala, Maneesh and Yuille, Alan and Jiang, Lu},
  journal={arXiv preprint arXiv:2507.18634},
  year={2025}
}

@article{yang2023gated,
  title={Gated linear attention transformers with hardware-efficient training},
  author={Yang, Songlin and Wang, Bailin and Shen, Yikang and Panda, Rameswar and Kim, Yoon},
  journal={arXiv preprint arXiv:2312.06635},
  year={2023}
}

@article{team2025kimi,
  title={Kimi linear: An expressive, efficient attention architecture},
  author={Team, Kimi and Zhang, Yu and Lin, Zongyu and Yao, Xingcheng and Hu, Jiaxi and Meng, Fanqing and Liu, Chengyin and Men, Xin and Yang, Songlin and Li, Zhiyuan and others},
  journal={arXiv preprint arXiv:2510.26692},
  year={2025}
}

@article{yan2025flashsparseattentionalternative,
  title={Flash Sparse Attention: More Efficient Natively Trainable Sparse Attention},
  author={Yan, Ran and Jiang, Youhe and Yuan, Binhang},
  journal={arXiv preprint arXiv:2508.18224},
  year={2025}
}

@article{sageattention,
  title={Sageattention: Accurate 8-bit attention for plug-and-play inference acceleration},
  author={Zhang, Jintao and Wei, Jia and Huang, Haofeng and Zhang, Pengle and Zhu, Jun and Chen, Jianfei},
  journal={arXiv preprint arXiv:2410.02367},
  year={2024}
}

@article{zhang2025spargeattn,
  title={Spargeattn: Accurate sparse attention accelerating any model inference},
  author={Zhang, Jintao and Xiang, Chendong and Huang, Haofeng and Wei, Jia and Xi, Haocheng and Zhu, Jun and Chen, Jianfei},
  journal={arXiv preprint arXiv:2502.18137},
  year={2025}
}

@article{li2025radial,
  title={Radial Attention: Sparse Attention with Energy Decay for Long Video Generation},
  author={Li, Xingyang and Li, Muyang and Cai, Tianle and Xi, Haocheng and Yang, Shuo and Lin, Yujun and Zhang, Lvmin and Yang, Songlin and Hu, Jinbo and Peng, Kelly and others},
  journal={arXiv preprint arXiv:2506.19852},
  year={2025}
}

@article{magcache,
  title={MagCache: Fast Video Generation with Magnitude-Aware Cache},
  author={Ma, Zehong and Wei, Longhui and Wang, Feng and Zhang, Shiliang and Tian, Qi},
  journal={arXiv preprint arXiv:2506.09045},
  year={2025}
}

@inproceedings{teacache,
  title={Timestep Embedding Tells: It's Time to Cache for Video Diffusion Model},
  author={Liu, Feng and Zhang, Shiwei and Wang, Xiaofeng and Wei, Yujie and Qiu, Haonan and Zhao, Yuzhong and Zhang, Yingya and Ye, Qixiang and Wan, Fang},
  booktitle={Proceedings of the Computer Vision and Pattern Recognition Conference},
  pages={7353--7363},
  year={2025}
}

@inproceedings{omnicache,
  title={OmniCache: A Trajectory-Oriented Global Perspective on Training-Free Cache Reuse for Diffusion Transformer Models},
  author={Chu, Huanpeng and Wu, Wei and Feng, Guanyu and Zhang, Yutao},
  booktitle={Proceedings of the IEEE/CVF International Conference on Computer Vision},
  pages={16302--16312},
  year={2025}
}

@article{sparse_videogen,
  title={Sparse videogen: Accelerating video diffusion transformers with spatial-temporal sparsity},
  author={Xi, Haocheng and Yang, Shuo and Zhao, Yilong and Xu, Chenfeng and Li, Muyang and Li, Xiuyu and Lin, Yujun and Cai, Han and Zhang, Jintao and Li, Dacheng and others},
  journal={arXiv preprint arXiv:2502.01776},
  year={2025}
}

@article{turbodiffusion,
  title={TurboDiffusion: Accelerating Video Diffusion Models by 100-200 Times},
  author={Zhang, Jintao and Zheng, Kaiwen and Jiang, Kai and Wang, Haoxu and Stoica, Ion and Gonzalez, Joseph E and Chen, Jianfei and Zhu, Jun},
  journal={arXiv preprint arXiv:2512.16093},
  year={2025}
}

@article{sd35,
  title={Scaling Rectified Flow Transformers for High-Resolution Image Synthesis},
  author={Esser, Patrick and Kulal, Sumith and Blattmann, Andreas and Entezari, Rahim and M{\"u}ller, Jonas and Saini, Harry and Levi, Yam and Lorenz, Dominik and Sauer, Axel and Boesel, Frederic and others},
  journal={arXiv preprint arXiv:2403.03206},
  year={2024}
}

@article{uniworldv2,
  title={Uniworld-v2: Reinforce image editing with diffusion negative-aware finetuning and mllm implicit feedback},
  author={Li, Zongjian and Liu, Zheyuan and Zhang, Qihui and Lin, Bin and Wu, Feize and Yuan, Shenghai and Yan, Zhiyuan and Ye, Yang and Yu, Wangbo and Niu, Yuwei and others},
  journal={arXiv preprint arXiv:2510.16888},
  year={2025}
}

@article{uniworld,
  title={Uniworld: High-resolution semantic encoders for unified visual understanding and generation},
  author={Lin, Bin and Li, Zongjian and Cheng, Xinhua and Niu, Yuwei and Ye, Yang and He, Xianyi and Yuan, Shenghai and Yu, Wangbo and Wang, Shaodong and Ge, Yunyang and others},
  journal={arXiv preprint arXiv:2506.03147},
  year={2025}
}

@article{sana_video,
  title={Sana-video: Efficient video generation with block linear diffusion transformer},
  author={Chen, Junsong and Zhao, Yuyang and Yu, Jincheng and Chu, Ruihang and Chen, Junyu and Yang, Shuai and Wang, Xianbang and Pan, Yicheng and Zhou, Daquan and Ling, Huan and others},
  journal={arXiv preprint arXiv:2509.24695},
  year={2025}
}

@article{streamdiffusionv2,
  title={StreamDiffusionV2: A Streaming System for Dynamic and Interactive Video Generation},
  author={Feng, Tianrui and Li, Zhi and Yang, Shuo and Xi, Haocheng and Li, Muyang and Li, Xiuyu and Zhang, Lvmin and Yang, Keting and Peng, Kelly and Han, Song and others},
  journal={arXiv preprint arXiv:2511.07399},
  year={2025}
}

@article{vbench2,
  title={Vbench-2.0: Advancing video generation benchmark suite for intrinsic faithfulness},
  author={Zheng, Dian and Huang, Ziqi and Liu, Hongbo and Zou, Kai and He, Yinan and Zhang, Fan and Gu, Lulu and Zhang, Yuanhan and He, Jingwen and Zheng, Wei-Shi and others},
  journal={arXiv preprint arXiv:2503.21755},
  year={2025}
}

@article{chronomagicbench,
  title={Chronomagic-bench: A benchmark for metamorphic evaluation of text-to-time-lapse video generation},
  author={Yuan, Shenghai and Huang, Jinfa and Xu, Yongqi and Liu, Yaoyang and Zhang, Shaofeng and Shi, Yujun and Zhu, Rui-Jie and Cheng, Xinhua and Luo, Jiebo and Yuan, Li},
  journal={Advances in Neural Information Processing Systems},
  volume={37},
  pages={21236--21270},
  year={2024}
}

@article{wan,
  title={Wan: Open and advanced large-scale video generative models},
  author={Wan, Team and Wang, Ang and Ai, Baole and Wen, Bin and Mao, Chaojie and Xie, Chen-Wei and Chen, Di and Yu, Feiwu and Zhao, Haiming and Yang, Jianxiao and others},
  journal={arXiv preprint arXiv:2503.20314},
  year={2025}
}

@article{hunyuanvideo,
  title={Hunyuanvideo: A systematic framework for large video generative models},
  author={Kong, Weijie and Tian, Qi and Zhang, Zijian and Min, Rox and Dai, Zuozhuo and Zhou, Jin and Xiong, Jiangfeng and Li, Xin and Wu, Bo and Zhang, Jianwei and others},
  journal={arXiv preprint arXiv:2412.03603},
  year={2024}
}

@article{hunyuanvideo15,
  title={Hunyuanvideo 1.5 technical report},
  author={Wu, Bing and Zou, Chang and Li, Changlin and Huang, Duojun and Yang, Fang and Tan, Hao and Peng, Jack and Wu, Jianbing and Xiong, Jiangfeng and Jiang, Jie and others},
  journal={arXiv preprint arXiv:2511.18870},
  year={2025}
}

@article{magi,
  title={MAGI-1: Autoregressive Video Generation at Scale},
  author={Teng, Hansi and Jia, Hongyu and Sun, Lei and Li, Lingzhi and Li, Maolin and Tang, Mingqiu and Han, Shuai and Zhang, Tianning and Zhang, WQ and Luo, Weifeng and others},
  journal={arXiv preprint arXiv:2505.13211},
  year={2025}
}

@article{cogvideox,
  title={Cogvideox: Text-to-video diffusion models with an expert transformer},
  author={Yang, Zhuoyi and Teng, Jiayan and Zheng, Wendi and Ding, Ming and Huang, Shiyu and Xu, Jiazheng and Yang, Yuanming and Hong, Wenyi and Zhang, Xiaohan and Feng, Guanyu and others},
  journal={arXiv preprint arXiv:2408.06072},
  year={2024}
}

@article{opensora,
  title={Open-sora: Democratizing efficient video production for all},
  author={Zheng, Zangwei and Peng, Xiangyu and Yang, Tianji and Shen, Chenhui and Li, Shenggui and Liu, Hongxin and Zhou, Yukun and Li, Tianyi and You, Yang},
  journal={arXiv preprint arXiv:2412.20404},
  year={2024}
}

@article{opensoraplan,
  title={Open-sora plan: Open-source large video generation model},
  author={Lin, Bin and Ge, Yunyang and Cheng, Xinhua and Li, Zongjian and Zhu, Bin and Wang, Shaodong and He, Xianyi and Ye, Yang and Yuan, Shenghai and Chen, Liuhan and others},
  journal={arXiv preprint arXiv:2412.00131},
  year={2024}
}

@article{ltx,
  title={Ltx-video: Realtime video latent diffusion},
  author={HaCohen, Yoav and Chiprut, Nisan and Brazowski, Benny and Shalem, Daniel and Moshe, Dudu and Richardson, Eitan and Levin, Eran and Shiran, Guy and Zabari, Nir and Gordon, Ori and others},
  journal={arXiv preprint arXiv:2501.00103},
  year={2024}
}

@article{ltx2,
  title={LTX-2: Efficient Joint Audio-Visual Foundation Model},
  author={HaCohen, Yoav and Brazowski, Benny and Chiprut, Nisan and Bitterman, Yaki and Kvochko, Andrew and Berkowitz, Avishai and Shalem, Daniel and Lifschitz, Daphna and Moshe, Dudu and Porat, Eitan and others},
  journal={arXiv preprint arXiv:2601.03233},
  year={2026}
}

@article{yume,
  title={Yume: An interactive world generation model},
  author={Mao, Xiaofeng and Lin, Shaoheng and Li, Zhen and Li, Chuanhao and Peng, Wenshuo and He, Tong and Pang, Jiangmiao and Chi, Mingmin and Qiao, Yu and Zhang, Kaipeng},
  journal={arXiv preprint arXiv:2507.17744},
  year={2025}
}

@article{rethinking_video_generation,
  title={Rethinking Video Generation Model for the Embodied World},
  author={Deng, Yufan and Pan, Zilin and Zhang, Hongyu and Li, Xiaojie and Hu, Ruoqing and Ding, Yufei and Zou, Yiming and Zeng, Yan and Zhou, Daquan},
  journal={arXiv preprint arXiv:2601.15282},
  year={2026}
}

@article{magref,
  title={MAGREF: Masked Guidance for Any-Reference Video Generation},
  author={Deng, Yufan and Guo, Xun and Yin, Yuanyang and Fang, Jacob Zhiyuan and Yang, Yiding and Wang, Yizhi and Yuan, Shenghai and Wang, Angtian and Liu, Bo and Huang, Haibin and others},
  journal={arXiv preprint arXiv:2505.23742},
  year={2025}
}

@article{yume15,
  title={Yume-1.5: A Text-Controlled Interactive World Generation Model},
  author={Mao, Xiaofeng and Li, Zhen and Li, Chuanhao and Xu, Xiaojie and Ying, Kaining and He, Tong and Pang, Jiangmiao and Qiao, Yu and Zhang, Kaipeng},
  journal={arXiv preprint arXiv:2512.22096},
  year={2025}
}

@article{hv_world15,
  title={WorldPlay: Towards Long-Term Geometric Consistency for Real-Time Interactive World Modeling},
  author={Sun, Wenqiang and Zhang, Haiyu and Wang, Haoyuan and Wu, Junta and Wang, Zehan and Wang, Zhenwei and Wang, Yunhong and Zhang, Jun and Wang, Tengfei and Guo, Chunchao},
  journal={arXiv preprint arXiv:2512.14614},
  year={2025}
}

@article{lingbot_world,
    title={Advancing Open-source World Models}, 
    author={LingBot-World Team},
    journal={arXiv preprint arXiv:2601.20540},
    year={2026},
}

@article{mineworld,
  title={Mineworld: a real-time and open-source interactive world model on minecraft},
  author={Guo, Junliang and Ye, Yang and He, Tianyu and Wu, Haoyu and Jiang, Yushu and Pearce, Tim and Bian, Jiang},
  journal={arXiv preprint arXiv:2504.08388},
  year={2025}
}

@article{hv_world,
  title={Hunyuanworld 1.0: Generating immersive, explorable, and interactive 3d worlds from words or pixels},
  author={Team, HunyuanWorld and Wang, Zhenwei and Liu, Yuhao and Wu, Junta and Gu, Zixiao and Wang, Haoyuan and Zuo, Xuhui and Huang, Tianyu and Li, Wenhuan and Zhang, Sheng and others},
  journal={arXiv preprint arXiv:2507.21809},
  year={2025}
}

@inproceedings{genie,
  title={Genie: Generative interactive environments},
  author={Bruce, Jake and Dennis, Michael D and Edwards, Ashley and Parker-Holder, Jack and Shi, Yuge and Hughes, Edward and Lai, Matthew and Mavalankar, Aditi and Steigerwald, Richie and Apps, Chris and others},
  booktitle={Forty-first International Conference on Machine Learning},
  year={2024}
}

@article{motionstream,
  title={Motionstream: Real-time video generation with interactive motion controls},
  author={Shin, Joonghyuk and Li, Zhengqi and Zhang, Richard and Zhu, Jun-Yan and Park, Jaesik and Shechtman, Eli and Huang, Xun},
  journal={arXiv preprint arXiv:2511.01266},
  year={2025}
}

@article{matrixgame,
  title={Matrix-Game: Interactive World Foundation Model},
  author={Zhang, Yifan and Peng, Chunli and Wang, Boyang and Wang, Puyi and Zhu, Qingcheng and Kang, Fei and Jiang, Biao and Gao, Zedong and Li, Eric and Liu, Yang and others},
  journal={arXiv preprint arXiv:2506.18701},
  year={2025}
}

@article{astra,
  title={Astra: General Interactive World Model with Autoregressive Denoising},
  author={Zhu, Yixuan and Feng, Jiaqi and Zheng, Wenzhao and Gao, Yuan and Tao, Xin and Wan, Pengfei and Zhou, Jie and Lu, Jiwen},
  journal={arXiv preprint arXiv:2512.08931},
  year={2025}
}

@article{genie3,
  title={Genie 3: A new frontier for world models},
  author={Ball, Philip J and Bauer, Jakob and Belletti, Frank and Brownfield, B and Ephrat, A and Fruchter, S and Gupta, A and Holsheimer, K and Holynski, A and Hron, J and others},
  journal={Google DeepMind Blog},
  pages={253--279},
  year={2025}
}

@article{hunyuan_Gamecraft,
  title={Hunyuan-GameCraft: High-dynamic Interactive Game Video Generation with Hybrid History Condition},
  author={Li, Jiaqi and Tang, Junshu and Xu, Zhiyong and Wu, Longhuang and Zhou, Yuan and Shao, Shuai and Yu, Tianbao and Cao, Zhiguo and Lu, Qinglin},
  journal={arXiv preprint arXiv:2506.17201},
  year={2025}
}

@inproceedings{consisid,
  title={Identity-preserving text-to-video generation by frequency decomposition},
  author={Yuan, Shenghai and Huang, Jinfa and He, Xianyi and Ge, Yunyang and Shi, Yujun and Chen, Liuhan and Luo, Jiebo and Yuan, Li},
  booktitle={Proceedings of the Computer Vision and Pattern Recognition Conference},
  pages={12978--12988},
  year={2025}
}

@inproceedings{omnihuman,
  title={Omnihuman-1: Rethinking the scaling-up of one-stage conditioned human animation models},
  author={Lin, Gaojie and Jiang, Jianwen and Yang, Jiaqi and Zheng, Zerong and Liang, Chao and Zhang, Yuan and Liu, Jingtuo},
  booktitle={Proceedings of the IEEE/CVF International Conference on Computer Vision},
  pages={13847--13858},
  year={2025}
}

@article{skyreelsv3,
  title={SkyReels-V3 Technique Report},
  author={Li, Debang and Fei, Zhengcong and Li, Tuanhui and Dou, Yikun and Chen, Zheng and Yang, Jiangping and Fan, Mingyuan and Xu, Jingtao and Wang, Jiahua and Gu, Baoxuan and others},
  journal={arXiv preprint arXiv:2601.17323},
  year={2026}
}

@misc{mochi,
      title={Mochi 1},
      author={Genmo Team},
      year={2024},
      publisher = {GitHub},
      journal = {GitHub repository},
      howpublished={\url{https://github.com/genmoai/models}}
}

@article{kandinsky5,
  title={Kandinsky 5.0: A Family of Foundation Models for Image and Video Generation},
  author={Arkhipkin, Vladimir and Korviakov, Vladimir and Gerasimenko, Nikolai and Parkhomenko, Denis and Vasilev, Viacheslav and Letunovskiy, Alexey and Vaulin, Nikolai and Kovaleva, Maria and Kirillov, Ivan and Novitskiy, Lev and others},
  journal={arXiv preprint arXiv:2511.14993},
  year={2025}
}

@article{magictime,
  title={Magictime: Time-lapse video generation models as metamorphic simulators},
  author={Yuan, Shenghai and Huang, Jinfa and Shi, Yujun and Xu, Yongqi and Zhu, Ruijie and Lin, Bin and Cheng, Xinhua and Yuan, Li and Luo, Jiebo},
  journal={IEEE Transactions on Pattern Analysis and Machine Intelligence},
  year={2025},
  publisher={IEEE}
}

@article{mineworldv2,
  title={Reinforcement Learning with Inverse Rewards for World Model Post-training},
  author={Ye, Yang and He, Tianyu and Yang, Shuo and Bian, Jiang},
  journal={arXiv preprint arXiv:2509.23958},
  year={2025}
}

@article{stepvideo,
  title={Step-video-t2v technical report: The practice, challenges, and future of video foundation model},
  author={Ma, Guoqing and Huang, Haoyang and Yan, Kun and Chen, Liangyu and Duan, Nan and Yin, Shengming and Wan, Changyi and Ming, Ranchen and Song, Xiaoniu and Chen, Xing and others},
  journal={arXiv preprint arXiv:2502.10248},
  year={2025}
}

@article{freenoise,
  title={Freenoise: Tuning-free longer video diffusion via noise rescheduling},
  author={Qiu, Haonan and Xia, Menghan and Zhang, Yong and He, Yingqing and Wang, Xintao and Shan, Ying and Liu, Ziwei},
  journal={arXiv preprint arXiv:2310.15169},
  year={2023}
}

@article{fifo-diffusion,
  title={Fifo-diffusion: Generating infinite videos from text without training},
  author={Kim, Jihwan and Kang, Junoh and Choi, Jinyoung and Han, Bohyung},
  journal={Advances in Neural Information Processing Systems},
  volume={37},
  pages={89834--89868},
  year={2024}
}

@article{diffusionforcing,
  title={Diffusion forcing: Next-token prediction meets full-sequence diffusion},
  author={Chen, Boyuan and Mart{\'\i} Mons{\'o}, Diego and Du, Yilun and Simchowitz, Max and Tedrake, Russ and Sitzmann, Vincent},
  journal={Advances in Neural Information Processing Systems},
  volume={37},
  pages={24081--24125},
  year={2024}
}

@article{rollingdiffusionmodel,
      title={Rolling Diffusion Models}, 
      author={David Ruhe and Jonathan Heek and Tim Salimans and Emiel Hoogeboom},
      year={2024},
      journal={arXiv preprint arXiv:2402.09470}, 
}

@article{skyreelsv2,
  title={Skyreels-v2: Infinite-length film generative model},
  author={Chen, Guibin and Lin, Dixuan and Yang, Jiangping and Lin, Chunze and Zhu, Junchen and Fan, Mingyuan and Zhang, Hao and Chen, Sheng and Chen, Zheng and Ma, Chengcheng and others},
  journal={arXiv preprint arXiv:2504.13074},
  year={2025}
}

@article{infinitystar,
  title={InfinityStar: Unified Spacetime AutoRegressive Modeling for Visual Generation},
  author={Liu, Jinlai and Han, Jian and Yan, Bin and Wu, Hui and Zhu, Fengda and Wang, Xing and Jiang, Yi and Peng, Bingyue and Yuan, Zehuan},
  journal={arXiv preprint arXiv:2511.04675},
  year={2025}
}

@inproceedings{ttt,
  title={One-minute video generation with test-time training},
  author={Dalal, Karan and Koceja, Daniel and Xu, Jiarui and Zhao, Yue and Han, Shihao and Cheung, Ka Chun and Kautz, Jan and Choi, Yejin and Sun, Yu and Wang, Xiaolong},
  booktitle={Proceedings of the Computer Vision and Pattern Recognition Conference},
  pages={17702--17711},
  year={2025}
}

@article{ttrl,
  title={Ttrl: Test-time reinforcement learning},
  author={Zuo, Yuxin and Zhang, Kaiyan and Sheng, Li and Qu, Shang and Cui, Ganqu and Zhu, Xuekai and Li, Haozhan and Zhang, Yuchen and Long, Xinwei and Hua, Ermo and others},
  journal={arXiv preprint arXiv:2504.16084},
  year={2025}
}

@inproceedings{ar_diffusion,
  title={Ar-diffusion: Asynchronous video generation with auto-regressive diffusion},
  author={Sun, Mingzhen and Wang, Weining and Li, Gen and Liu, Jiawei and Sun, Jiahui and Feng, Wanquan and Lao, Shanshan and Zhou, SiYu and He, Qian and Liu, Jing},
  booktitle={Proceedings of the Computer Vision and Pattern Recognition Conference},
  pages={7364--7373},
  year={2025}
}

@article{diffusionforcingv2,
  title={History-guided video diffusion},
  author={Song, Kiwhan and Chen, Boyuan and Simchowitz, Max and Du, Yilun and Tedrake, Russ and Sitzmann, Vincent},
  journal={arXiv preprint arXiv:2502.06764},
  year={2025}
}

@article{rcm,
  title={Large Scale Diffusion Distillation via Score-Regularized Continuous-Time Consistency},
  author={Zheng, Kaiwen and Wang, Yuji and Ma, Qianli and Chen, Huayu and Zhang, Jintao and Balaji, Yogesh and Chen, Jianfei and Liu, Ming-Yu and Zhu, Jun and Zhang, Qinsheng},
  journal={arXiv preprint arXiv:2510.08431},
  year={2025}
}

@article{concatid,
  title={Concat-ID: Towards Universal Identity-Preserving Video Synthesis},
  author={Zhong, Yong and Yang, Zhuoyi and Teng, Jiayan and Gu, Xiaotao and Li, Chongxuan},
  journal={arXiv preprint arXiv:2503.14151},
  year={2025}
}

@article{standin,
  title={Stand-in: A lightweight and plug-and-play identity control for video generation},
  author={Xue, Bowen and Duan, Zheng-Peng and Yan, Qixin and Wang, Wenjing and Liu, Hao and Guo, Chun-Le and Li, Chongyi and Li, Chen and Lyu, Jing},
  journal={arXiv preprint arXiv:2508.07901},
  year={2025}
}

@article{pyramidflow,
  title={Pyramidal Flow Matching for Efficient Video Generative Modeling},
  author={Jin, Yang and Sun, Zhicheng and Li, Ningyuan and Xu, Kun and Xu, Kun and Jiang, Hao and Zhuang, Nan and Huang, Quzhe and Song, Yang and Mu, Yadong and Lin, Zhouchen},
  jounal={arXiv preprint arXiv:2410.05954},
  year={2024}
}

@article{far,
  title={Long-context autoregressive video modeling with next-frame prediction},
  author={Gu, Yuchao and Mao, Weijia and Shou, Mike Zheng},
  journal={arXiv preprint arXiv:2503.19325},
  year={2025}
}

@article{hitvideo,
  title={Hitvideo: Hierarchical tokenizers for enhancing text-to-video generation with autoregressive large language models},
  author={Zhou, Ziqin and Yang, Yifan and Yang, Yuqing and He, Tianyu and Peng, Houwen and Qiu, Kai and Dai, Qi and Qiu, Lili and Luo, Chong and Liu, Lingqiao},
  journal={arXiv preprint arXiv:2503.11513},
  year={2025}
}

@article{var,
  title={Visual autoregressive modeling: Scalable image generation via next-scale prediction},
  author={Tian, Keyu and Jiang, Yi and Yuan, Zehuan and Peng, Bingyue and Wang, Liwei},
  journal={Advances in neural information processing systems},
  volume={37},
  pages={84839--84865},
  year={2024}
}

@article{progressivedistillation,
  title={Progressive distillation for fast sampling of diffusion models},
  author={Salimans, Tim and Ho, Jonathan},
  journal={arXiv preprint arXiv:2202.00512},
  year={2022}
}

@article{nova,
  title={Autoregressive video generation without vector quantization},
  author={Deng, Haoge and Pan, Ting and Diao, Haiwen and Luo, Zhengxiong and Cui, Yufeng and Lu, Huchuan and Shan, Shiguang and Qi, Yonggang and Wang, Xinlong},
  journal={arXiv preprint arXiv:2412.14169},
  year={2024}
}

@article{SVI,
  title={Stable Video Infinity: Infinite-Length Video Generation with Error Recycling},
  author={Li, Wuyang and Pan, Wentao and Luan, Po-Chien and Gao, Yang and Alahi, Alexandre},
  journal={arXiv preprint arXiv:2510.09212},
  year={2025}
}

@article{LCT,
  title={Long context tuning for video generation},
  author={Guo, Yuwei and Yang, Ceyuan and Yang, Ziyan and Ma, Zhibei and Lin, Zhijie and Yang, Zhenheng and Lin, Dahua and Jiang, Lu},
  journal={arXiv preprint arXiv:2503.10589},
  year={2025}
}

@article{moga,
  title={MoGA: Mixture-of-Groups Attention for End-to-End Long Video Generation},
  author={Jia, Weinan and Lu, Yuning and Huang, Mengqi and Wang, Hualiang and Huang, Binyuan and Chen, Nan and Liu, Mu and Jiang, Jidong and Mao, Zhendong},
  journal={arXiv preprint arXiv:2510.18692},
  year={2025}
}

@article{selfforcing,
  title={Self Forcing: Bridging the Train-Test Gap in Autoregressive Video Diffusion},
  author={Huang, Xun and Li, Zhengqi and He, Guande and Zhou, Mingyuan and Shechtman, Eli},
  journal={arXiv preprint arXiv:2506.08009},
  year={2025}
}

@article{selfforcing++,
  title={Self-Forcing++: Towards Minute-Scale High-Quality Video Generation},
  author={Cui, Justin and Wu, Jie and Li, Ming and Yang, Tao and Li, Xiaojie and Wang, Rui and Bai, Andrew and Ban, Yuanhao and Hsieh, Cho-Jui},
  journal={arXiv preprint arXiv:2510.02283},
  year={2025}
}

@inproceedings{causvid,
  title={From slow bidirectional to fast autoregressive video diffusion models},
  author={Yin, Tianwei and Zhang, Qiang and Zhang, Richard and Freeman, William T and Durand, Fredo and Shechtman, Eli and Huang, Xun},
  booktitle={Proceedings of the Computer Vision and Pattern Recognition Conference},
  pages={22963--22974},
  year={2025}
}

@article{framepack,
  title={Packing input frame context in next-frame prediction models for video generation},
  author={Zhang, Lvmin and Agrawala, Maneesh},
  journal={arXiv preprint arXiv:2504.12626},
  year={2025}
}

@article{rollingforcing,
  title={Rolling Forcing: Autoregressive Long Video Diffusion in Real Time},
  author={Liu, Kunhao and Hu, Wenbo and Xu, Jiale and Shan, Ying and Lu, Shijian},
  journal={arXiv preprint arXiv:2509.25161},
  year={2025}
}

@article{longlive,
  title={Longlive: Real-time interactive long video generation},
  author={Yang, Shuai and Huang, Wei and Chu, Ruihang and Xiao, Yicheng and Zhao, Yuyang and Wang, Xianbang and Li, Muyang and Xie, Enze and Chen, Yingcong and Lu, Yao and others},
  journal={arXiv preprint arXiv:2509.22622},
  year={2025}
}

@misc{krea_realtime_14b,
  title={Krea Realtime 14B: Real-time Video Generation},
  author={Erwann Millon},
  year={2025},
  url={https://github.com/krea-ai/realtime-video},
}

@misc{unsloth,
  author = {Daniel Han, Michael Han and Unsloth team},
  title = {Unsloth},
  url = {http://github.com/unslothai/unsloth},
  year = {2023}
}

@inproceedings{
    hsu2025ligerkernel,
    title={Liger-Kernel: Efficient Triton Kernels for {LLM} Training},
    author={Pin-Lun Hsu and Yun Dai and Vignesh Kothapalli and Qingquan Song and Shao Tang and Siyu Zhu and Steven Shimizu and Shivam Sahni and Haowen Ning and Yanning Chen and Zhipeng Wang},
    booktitle={Championing Open-source DEvelopment in ML Workshop @ ICML25},
    year={2025},
    url={https://openreview.net/forum?id=36SjAIT42G}
}

@article{vbench++,
  title={Vbench++: Comprehensive and versatile benchmark suite for video generative models},
  author={Huang, Ziqi and Zhang, Fan and Xu, Xiaojie and He, Yinan and Yu, Jiashuo and Dong, Ziyue and Ma, Qianli and Chanpaisit, Nattapol and Si, Chenyang and Jiang, Yuming and others},
  journal={IEEE Transactions on Pattern Analysis and Machine Intelligence},
  year={2025},
  publisher={IEEE}
}

@article{apt,
  title={Diffusion adversarial post-training for one-step video generation},
  author={Lin, Shanchuan and Xia, Xin and Ren, Yuxi and Yang, Ceyuan and Xiao, Xuefeng and Jiang, Lu},
  journal={arXiv preprint arXiv:2501.08316},
  year={2025}
}

@article{fastvideo,
  title={Fast video generation with sliding tile attention},
  author={Zhang, Peiyuan and Chen, Yongqi and Su, Runlong and Ding, Hangliang and Stoica, Ion and Liu, Zhengzhong and Zhang, Hao},
  journal={arXiv preprint arXiv:2502.04507},
  year={2025}
}

@article{alphaface,
  title={AlphaFace: High Fidelity and Real-time Face Swapper Robust to Facial Pose},
  author={Yu, Jongmin and Oh, Hyeontaek and Sun, Zhongtian and Aviles-Rivero, Angelica I and Jeon, Moongu and Yang, Jinhong},
  journal={arXiv preprint arXiv:2601.16429},
  year={2026}
}

@article{causalforcing,
      title={Causal Forcing: Autoregressive Diffusion Distillation Done Right for High-Quality Real-Time Interactive Video Generation}, 
      author={Hongzhou Zhu and Min Zhao and Guande He and Hang Su and Chongxuan Li and Jun Zhu},
      year={2026},
      journal={arXiv preprint arXiv:2602.02214}, 
}

@misc{infinite_forcing,
    Author = {Junyi Chen, Zhoujie Fu, Xianglong He},
    Year = {2025},
    Note = {https://github.com/SOTAMak1r/Infinite-Forcing},
    Title = {Infinite-Forcing: Towards Infinite-Long Video Generation}
}

@article{decoupleddmd,
  title={Decoupled DMD: CFG Augmentation as the Spear, Distribution Matching as the Shield},
  author={Dongyang Liu and Peng Gao and David Liu and Ruoyi Du and Zhen Li and Qilong Wu and Xin Jin and Sihan Cao and Shifeng Zhang and Hongsheng Li and Steven Hoi},
  journal={arXiv preprint arXiv:2511.22677},
  year={2025}
}

@inproceedings{peters2007reinforcement,
  title={Reinforcement learning by reward-weighted regression for operational space control},
  author={Peters, Jan and Schaal, Stefan},
  booktitle={Proceedings of the 24th international conference on Machine learning},
  pages={745--750},
  year={2007}
}

@inproceedings{flashattention,
  title={FlashAttention: Fast and Memory-Efficient Exact Attention with {IO}-Awareness},
  author={Dao, Tri and Fu, Daniel Y. and Ermon, Stefano and Rudra, Atri and R{\'e}, Christopher},
  booktitle={Advances in Neural Information Processing Systems (NeurIPS)},
  year={2022}
}

@inproceedings{flashattention2,
  title={FlashAttention-2: Faster Attention with Better Parallelism and Work Partitioning},
  author={Dao, Tri},
  booktitle={International Conference on Learning Representations (ICLR)},
  year={2024}
}

@misc{spark_wan,
  title={SparkWan: Few-Step Video Diffusion Transformer via Hybrid Adversarial Post-Training},
  author={Li, Zongjian and Yuan, Shenghai and Yuan, Li},
  year={2025},
  url={https://github.com/PKU-YuanGroup/Spark-Wan},
}

@article{self_resampling,
  title={End-to-end training for autoregressive video diffusion via self-resampling},
  author={Guo, Yuwei and Yang, Ceyuan and He, Hao and Zhao, Yang and Wei, Meng and Yang, Zhenheng and Huang, Weilin and Lin, Dahua},
  journal={arXiv preprint arXiv:2512.15702},
  year={2025}
}

@article{bagger,
  title={BAgger: Backwards Aggregation for Mitigating Drift in Autoregressive Video Diffusion Models},
  author={Po, Ryan and Chan, Eric Ryan and Chen, Changan and Wetzstein, Gordon},
  journal={arXiv preprint arXiv:2512.12080},
  year={2025}
}

@article{emu35,
  title={Emu3. 5: Native multimodal models are world learners},
  author={Cui, Yufeng and Chen, Honghao and Deng, Haoge and Huang, Xu and Li, Xinghang and Liu, Jirong and Liu, Yang and Luo, Zhuoyan and Wang, Jinsheng and Wang, Wenxuan and others},
  journal={arXiv preprint arXiv:2510.26583},
  year={2025}
}

@article{cai2025mixture,
  title={Mixture of contexts for long video generation},
  author={Cai, Shengqu and Yang, Ceyuan and Zhang, Lvmin and Guo, Yuwei and Xiao, Junfei and Yang, Ziyan and Xu, Yinghao and Yang, Zhenheng and Yuille, Alan and Guibas, Leonidas and others},
  journal={arXiv preprint arXiv:2508.21058},
  year={2025}
}

@article{storydiffusion,
  title={Storydiffusion: Consistent self-attention for long-range image and video generation},
  author={Zhou, Yupeng and Zhou, Daquan and Cheng, Ming-Ming and Feng, Jiashi and Hou, Qibin},
  journal={Advances in Neural Information Processing Systems},
  volume={37},
  pages={110315--110340},
  year={2024}
}

@article{ic_lora,
  title={In-context lora for diffusion transformers},
  author={Huang, Lianghua and Wang, Wei and Wu, Zhi-Fan and Shi, Yupeng and Dou, Huanzhang and Liang, Chen and Feng, Yutong and Liu, Yu and Zhou, Jingren},
  journal={arXiv preprint arXiv:2410.23775},
  year={2024}
}

@article{fdm,
  title={f-dm: A multi-stage diffusion model via progressive signal transformation},
  author={Gu, Jiatao and Zhai, Shuangfei and Zhang, Yizhe and Bautista, Miguel Angel and Susskind, Josh},
  journal={arXiv preprint arXiv:2210.04955},
  year={2022}
}

@article{bottleneck_sampling,
  title={Training-free diffusion acceleration with bottleneck sampling},
  author={Tian, Ye and Xia, Xin and Ren, Yuxi and Lin, Shanchuan and Wang, Xing and Xiao, Xuefeng and Tong, Yunhai and Yang, Ling and Cui, Bin},
  journal={arXiv preprint arXiv:2503.18940},
  year={2025}
}

@inproceedings{DMD,
    title={Improved Distribution Matching Distillation for Fast Image Synthesis},
    author={Yin, Tianwei and Gharbi, Micha{\"e}l and Park, Taesung and Zhang, Richard and Shechtman, Eli and Durand, Fredo and Freeman, William T},
    booktitle={NeurIPS},
    year={2024}
}

@article{videoalign,
  title={Improving video generation with human feedback},
  author={Liu, Jie and Liu, Gongye and Liang, Jiajun and Yuan, Ziyang and Liu, Xiaokun and Zheng, Mingwu and Wu, Xiele and Wang, Qiulin and Xia, Menghan and Wang, Xintao and others},
  journal={arXiv preprint arXiv:2501.13918},
  year={2025}
}

@inproceedings{DMD2,
    title={One-step Diffusion with Distribution Matching Distillation},
    author={Yin, Tianwei and Gharbi, Micha{\"e}l and Zhang, Richard and Shechtman, Eli and Durand, Fr{\'e}do and Freeman, William T and Park, Taesung},
    booktitle={CVPR},
    year={2024}
}

@article{sdxl-lightning,
  title={Sdxl-lightning: Progressive adversarial diffusion distillation},
  author={Lin, Shanchuan and Wang, Anran and Yang, Xiao},
  journal={arXiv preprint arXiv:2402.13929},
  year={2024}
}

@article{LCM,
  title={Latent consistency models: Synthesizing high-resolution images with few-step inference},
  author={Luo, Simian and Tan, Yiqin and Huang, Longbo and Li, Jian and Zhao, Hang},
  journal={arXiv preprint arXiv:2310.04378},
  year={2023}
}

@article{LCMlora,
  title={Lcm-lora: A universal stable-diffusion acceleration module},
  author={Luo, Simian and Tan, Yiqin and Patil, Suraj and Gu, Daniel and von Platen, Patrick and Passos, Apolin{\'a}rio and Huang, Longbo and Li, Jian and Zhao, Hang},
  journal={arXiv preprint arXiv:2311.05556},
  year={2023}
}

@article{unipc,
  title={Unipc: A unified predictor-corrector framework for fast sampling of diffusion models},
  author={Zhao, Wenliang and Bai, Lujia and Rao, Yongming and Zhou, Jie and Lu, Jiwen},
  journal={Advances in Neural Information Processing Systems},
  volume={36},
  pages={49842--49869},
  year={2023}
}

@article{aesthecit_predictor,
  title={Laion-5b: An open large-scale dataset for training next generation image-text models},
  author={Schuhmann, Christoph and Beaumont, Romain and Vencu, Richard and Gordon, Cade and Wightman, Ross and Cherti, Mehdi and Coombes, Theo and Katta, Aarush and Mullis, Clayton and Wortsman, Mitchell and others},
  journal={Advances in neural information processing systems},
  volume={35},
  pages={25278--25294},
  year={2022}
}

@inproceedings{dynamic,
  title={Real-time intermediate flow estimation for video frame interpolation},
  author={Huang, Zhewei and Zhang, Tianyuan and Heng, Wen and Shi, Boxin and Zhou, Shuchang},
  booktitle={European Conference on Computer Vision},
  pages={624--642},
  year={2022},
  organization={Springer}
}

@inproceedings{raft,
  title={Raft: Recurrent all-pairs field transforms for optical flow},
  author={Teed, Zachary and Deng, Jia},
  booktitle={European conference on computer vision},
  pages={402--419},
  year={2020},
  organization={Springer}
}

@article{viclip,
  title={Internvid: A large-scale video-text dataset for multimodal understanding and generation},
  author={Wang, Yi and He, Yinan and Li, Yizhuo and Li, Kunchang and Yu, Jiashuo and Ma, Xin and Li, Xinhao and Chen, Guo and Chen, Xinyuan and Wang, Yaohui and others},
  journal={arXiv preprint arXiv:2307.06942},
  year={2023}
}

@article{opens2v,
  title={Opens2v-nexus: A detailed benchmark and million-scale dataset for subject-to-video generation},
  author={Yuan, Shenghai and He, Xianyi and Deng, Yufan and Ye, Yang and Huang, Jinfa and Lin, Bin and Luo, Jiebo and Yuan, Li},
  journal={arXiv preprint arXiv:2505.20292},
  year={2025}
}

@article{dancegrpo,
  title={DanceGRPO: Unleashing GRPO on Visual Generation},
  author={Xue, Zeyue and Wu, Jie and Gao, Yu and Kong, Fangyuan and Zhu, Lingting and Chen, Mengzhao and Liu, Zhiheng and Liu, Wei and Guo, Qiushan and Huang, Weilin and others},
  journal={arXiv preprint arXiv:2505.07818},
  year={2025}
}

@article{vidprom,
  title={Vidprom: A million-scale real prompt-gallery dataset for text-to-video diffusion models},
  author={Wang, Wenhao and Yang, Yi},
  journal={Advances in Neural Information Processing Systems},
  volume={37},
  pages={65618--65642},
  year={2024}
}

@article{flowgrpo,
  title={Flow-grpo: Training flow matching models via online rl},
  author={Liu, Jie and Liu, Gongye and Liang, Jiajun and Li, Yangguang and Liu, Jiaheng and Wang, Xintao and Wan, Pengfei and Zhang, Di and Ouyang, Wanli},
  journal={arXiv preprint arXiv:2505.05470},
  year={2025}
}

@misc{cfgzerostar,
      title={CFG-Zero*: Improved Classifier-Free Guidance for Flow Matching Models}, 
      author={Weichen Fan and Amber Yijia Zheng and Raymond A. Yeh and Ziwei Liu},
      year={2025},
      eprint={2503.18886},
      archivePrefix={arXiv},
      primaryClass={cs.CV},
      url={https://arxiv.org/abs/2503.18886}, 
}

@article{fsvideo,
  title={FSVideo: Fast Speed Video Diffusion Model in a Highly-Compressed Latent Space},
  author={Team, FSVideo and Chen, Qingyu and Fang, Zhiyuan and Huang, Haibin and Huang, Xinwei and Jin, Tong and Lin, Minxuan and Liu, Bo and Liu, Celong and Ma, Chongyang and others},
  journal={arXiv preprint arXiv:2602.02092},
  year={2026}
}

\clearpage

\beginappendix
% \clearpage
\appendix
% \startcontents[chapters]
% {
%   \centering
%   \textbf{\Large Helios}

%   \vspace{5pt}
%   {\Large Appendix}
%   \par
% }
% \printcontents[chapters]{}{1}{}

\section{Calculation of \ours-Bench}\label{sec:calculation_of_bench}
To provide a robust and human-aligned evaluation of generated videos, \ours-Bench employs a multi-stage scoring pipeline that normalizes, discretizes, and aggregates raw automated metrics to mitigate inherent noise and scale variations. Specifically, let $s_k$ denote the raw score of a generated video evaluated on metric $k$. We first bound and normalize each raw score to a standard range of $[0, 1]$ to compute the normalized score $\bar{s}_k = \max(0, \min(1, \frac{s_k - s_{min, k}}{s_{max, k} - s_{min, k}}))$, where $s_{min, k}$ and $s_{max, k}$ are the predefined bounds for metric $k$. Because raw continuous scores often correlate poorly with human perception, we then map each normalized score $\bar{s}_k$ to a discrete 10-point rating $r_k \in \{1, 2, \dots, 10\}$ using a step function parameterized by empirically derived thresholds $\mathcal{T}_k = [\tau_0, \tau_1, \dots, \tau_8]$, as shown in Table \ref{tab:metric_thresholds}. For higher-is-better spatial and temporal quality metrics (i.e., Aesthetic, Motion Amplitude, Motion Smoothness, Semantic, and Naturalness), the rating is assigned by finding the first threshold the score meets or exceeds, formulated as $r_k = 10 - i$ such that $i = \min \{ j \mid \bar{s}_k \ge \tau_j \}$, with a minimum rating of $1$ if $\bar{s}_k < \tau_8$. Conversely, for lower-is-better metrics (specifically the drifting variants used in long-video evaluation), the rating is computed as $r_k = 10 - i$ such that $i = \min \{ j \mid \bar{s}_k \le \tau_j \}$. Finally, the overall performance of a model is calculated as the duration-aware weighted sum of its discretized ratings, $S = \sum_{k} w_k r_k$, where $w_k$ denotes the metric-specific weight. For short videos, the evaluation heavily prioritizes Semantic alignment and Naturalness (each weighted at $0.35$), with Aesthetic, Motion Amplitude, and Motion Smoothness each weighted at $0.10$. For long videos, to adequately penalize spatial and semantic degradation over time, drifting metrics (Drifting Semantic, Drifting Naturalness, Drifting Aesthetic, and Drifting Motion Smoothness) are introduced with a weight of $0.099$ each, while the base weights for Semantic and Naturalness are reduced to $0.255$, and Aesthetic, Motion Amplitude, and Motion Smoothness are reduced to $0.03$.

\begin{table*}[h]
    \centering
    \caption{Empirical Thresholds ($\mathcal{T}_k$) for Metric Discretization. For the Aesthetic metric, the final two thresholds are marked with an asterisk (*) to indicate a potential anomaly in the empirical data or source configuration.}
    \label{tab:metric_thresholds}
    \resizebox{\textwidth}{!}{
    \begin{tabular}{l l c c c c c c c c c}
        \toprule
        \textbf{Metric ($k$)} & \textbf{Type} & $\bm{\tau_0}$ & $\bm{\tau_1}$ & $\bm{\tau_2}$ & $\bm{\tau_3}$ & $\bm{\tau_4}$ & $\bm{\tau_5}$ & $\bm{\tau_6}$ & $\bm{\tau_7}$ & $\bm{\tau_8}$ \\
        \midrule
        Aesthetic & Higher Better & 0.70 & 0.65 & 0.60 & 0.55 & 0.50 & 0.45 & 0.40 & 0.35 & 0.30 \\
        Motion Amplitude & Higher Better & 0.45 & 0.40 & 0.35 & 0.30 & 0.25 & 0.20 & 0.15 & 0.10 & 0.05 \\
        Motion Smoothness & Higher Better & 0.99 & 0.98 & 0.97 & 0.96 & 0.95 & 0.94 & 0.93 & 0.92 & 0.91 \\
        Semantic & Higher Better & 0.30 & 0.29 & 0.28 & 0.27 & 0.26 & 0.25 & 0.24 & 0.23 & 0.22 \\
        Naturalness & Higher Better & 0.65 & 0.60 & 0.55 & 0.50 & 0.45 & 0.40 & 0.30 & 0.25 & 0.20 \\
        \midrule
        Drifting Aesthetic & Lower Better & 0.01 & 0.02 & 0.03 & 0.04 & 0.05 & 0.06 & 0.07 & 0.08 & 0.09 \\
        Drifting Motion Smoothness & Lower Better & 0.01 & 0.02 & 0.03 & 0.04 & 0.05 & 0.06 & 0.07 & 0.08 & 0.09 \\
        Drifting Semantic & Lower Better & 0.01 & 0.02 & 0.03 & 0.04 & 0.05 & 0.06 & 0.07 & 0.08 & 0.09 \\
        Drifting Naturalness & Lower Better & 0.06 & 0.08 & 0.10 & 0.12 & 0.14 & 0.16 & 0.18 & 0.20 & 0.22 \\
        \bottomrule
    \end{tabular}
    }
\end{table*}

\section{Latent Re-encode.}
Since the ablation of self-forcing employs Wan-2.1-T2V \cite{wan} as the real/fake-score estimator—which requires 21 latent frames for optimal performance—\ours generates videos section-by-section in the latent space to satisfy this constraint. However, concatenating multiple sections introduces a distributional mismatch: within each section, the first frame exhibits statistics that differ substantially from those of subsequent frames. Consequently, the concatenated sequence contains multiple occurrences of a ``first-frame'' distribution, biasing the score estimators. To eliminate this artifact, we decode each latent section of $x_0^K$ via the VAE decoder and re-encode the resulting video as a single, continuous latent sequence, which contains only a single first-frame distribution. The standard re-noising operation is then applied. This re-encoding step improves distributional alignment with both $p_{\text{real}}$ and $p_{\text{fake}}$, thereby reducing score-estimation bias and enhancing temporal coherence.

% 由于我们选择Wan-2.1-T2V作为fake-score estimator，which需要输入21个latent帧才能获取最优的效果。因此\textit{\ours}需要generates videos section by section in latent space以满足这个要求。

% 由于 \textit{\ours} 采用逐 latent-section 的生成方式，$x_0^K$ 既可以仅包含单个 latent-section，也可以由多个连续的 latent-section 组成。对于仅包含单个 latent-section 的情况，可以直接对其进行 re-noise 并输入到 $p_{real}$ 和 $p_{fake}$ 中。
% 然而，当 $x_0^K$ 由多个 latent-section 拼接而成时，会面临分布不一致的问题：每个 latent-section 的首帧在统计分布上与其后续帧存在显著差异，若直接将多个 latent-section 拼接在一起，会在序列中引入多个“首帧分布”，从而导致 score estimator 出现偏差，如图 XXX 所示。为解决这一问题，我们首先使用 VAE 对 $x_0^K$ 进行逐 latent-section 的 decode，随后再将解码后的结果重新 encode 为一个统一的连续序列，使其仅包含单一的首帧分布，最后再进行 re-noise 操作。通过该过程，$x_0^K$ 的分布能够与 $p_{real}$ 和 $p_{fake}$ 更好地对齐，从而有效缓解由多首帧分布带来的估计误差，并使最终生成的视频在时序上更加连贯和自然。

% \begin{figure*}[!t]
%   \centering
%   \includegraphics[width=0.7\linewidth]{figures/Latent_Re-encode.pdf}
%   \caption{\textbf{Workflow of Latent Re-encoding.} To better align the distribution of $x_0^K$ with $p_{\text{real}}$ and $p_{\text{fake}}$, we use VAE to decode each section independently and then re-encode them into a unified sequence.}
%   \label{figure_latent_re-encode}
% \end{figure*}

\end{document}